\documentclass[10pt,fullpage,letterpaper]{article}

\usepackage{natbib}

\setcitestyle{authoryear,round,citesep={;},aysep={,},yysep={;}}
\usepackage[english]{babel}

\usepackage[margin=1in]{geometry}
\usepackage{fancyhdr}
\usepackage[utf8]{inputenc} 
\usepackage[T1]{fontenc}    
\usepackage{titletoc}
\usepackage[toc, page, header]{appendix}
\usepackage{url}        
\usepackage{booktabs}       
\usepackage{amsfonts}       
\usepackage{nicefrac}      
\usepackage{microtype}     
\usepackage{xcolor}       

\usepackage{parskip}

\title{Introducing Feedback Delays to Non-Delayed Adversarial MDPs: \\Optimal $\sqrt{S}$ and $\sqrt{K}$ Dependencies in Tabular and Linear AMDPs}
\author{%
    Aarush Kulkarni\,\thanks{Independent Researcher. Email: \texttt{aarushkulkarni26@gmail.com}} \and Khang Nguyen\,\thanks{Department of Mathematics, University of Southern California. Email: \texttt{slnguyen@usc.edu}} \and William Chang\,\thanks{Department of Mathematics, University of California Los Angeles. Email: \texttt{chang314@g.ucla.edu}} \and Yan Dai\,\thanks{Corresponding author. Operations Research Center, Massachusetts Institute of Technology. Email: \texttt{yandai20@mit.edu}}%
}
\date{}

\usepackage{multirow}
\usepackage{arydshln}
\usepackage{footnote}
\usepackage{enumitem}
\setitemize{leftmargin=1.5em,itemsep=0.1em,topsep=0.2em}
\setenumerate{leftmargin=1.5em,itemsep=0.1em,topsep=0.2em}
\usepackage{tabularx}
\usepackage{colortbl}

\usepackage{amsmath}
\usepackage{amsthm}
\usepackage{amssymb}
\usepackage{mathtools}
\usepackage{thmtools}
\usepackage{comment}
\usepackage{xspace}
\usepackage{bbm}
\usepackage{bm}
\allowdisplaybreaks

\let\originalmiddle=\middle
\def\middle#1{\mathrel{}\originalmiddle#1\mathrel{}}

\usepackage{algorithm}
\usepackage[noEnd,commentColor=blue!70!white]{algpseudocodex}
\renewcommand{\algorithmicrequire}{\textbf{Input:}}

\makeatletter
\newcommand{\algeq}[2][]{
  \par
  \noindent
  \makebox[\columnwidth]{
    \hfill #2
    \if\relax\detokenize{#1}\relax
      \hfill
    \else
      \hfill (\refstepcounter{equation}\theequation)\label{#1}%
    \fi
  }
  \par
}
\makeatother
\usepackage[colorlinks=true, linkcolor=blue!60!black, citecolor=blue!60!black,
urlcolor=black]{hyperref}
\usepackage{cleveref}
\AddToHook{cmd/appendix/before}{
    \crefalias{section}{appendix}
    \crefalias{subsection}{appendix}
}

\Crefname{ALG@line}{Line}{Lines}
\Crefname{assumption}{Assumption}{Assumptions}
\Crefname{protocol}{Protocol}{Protocols}
\Crefformat{equation}{Eq. #2(#1)#3}
\Crefrangeformat{equation}{Eqs. #3(#1)#4 to #5(#2)#6}
\Crefmultiformat{equation}{Eqs. #2(#1)#3}{ and #2(#1)#3}{, #2(#1)#3}{ and #2(#1)#3}
\Crefrangemultiformat{equation}{Eqs. #3(#1)#4 to #5(#2)#6}{ and #3(#1)#4 to #5(#2)#6}{, #3(#1)#4 to #5(#2)#6}{ and #3(#1)#4 to #5(#2)#6}
\newtheorem{theorem}{Theorem}[section]
\newtheorem{lemma}[theorem]{Lemma}

\theoremstyle{definition}
\newtheorem{definition}[theorem]{Definition}

\makeatletter
\newcommand{\citepname}[2]{\nocite{#1}(\hyper@linkstart{cite}{cite.#1}#2\hyper@linkend)}
\makeatother
\algrenewcommand\algorithmicrequire{\textbf{Input:}}
\newcommand{\Otil}{\widetilde{\mathcal O}}
\renewcommand{\O}{{\mathcal O}}
\newcommand{\Thetatil}{\widetilde{\Theta}}
\newcommand{\Omegatil}{\widetilde{\Omega}}
\DeclareMathOperator{\poly}{poly}
\newcommand{\argmin}{\operatornamewithlimits{argmin}}
\newcommand{\mR}{\mathcal{R}}
\newcommand{\cS}{\mathcal{S}}
\newcommand{\cA}{\mathcal{A}}
\newcommand{\cP}{\mathcal{P}}
\newcommand{\cF}{\mathcal{F}}

\newcommand{\cH}{\mathcal{H}}
\newcommand{\cC}{\mathcal{C}}
\newcommand{\cI}{\mathcal{I}}
\newcommand{\cT}{\mathcal{T}}
\newcommand{\R}{\mathbb{R}}
\newcommand{\N}{\mathbb{N}}
\newcommand{\E}{\operatornamewithlimits{\mathbb{E}}}
\newcommand{\PP}{\mathbb{P}} % true transition
\newcommand{\ind}{\mathbbm{1}}
\newcommand{\qlow}{\underline{q}}
\newcommand{\Qhat}{\widehat{Q}}
\newcommand{\Qtil}{\widetilde{Q}}
\newcommand{\what}{\widehat{w}}
\newcommand{\Phat}{\widehat{P}}
\newcommand{\Ptil}{\widetilde{P}}
\newcommand{\Sigmahat}{\widehat{\Sigma}}
\newcommand{\phat}{\widehat{p}}
\newcommand{\fd}{\mathfrak{d}} % feedback delays: \fd_k is the delay of episode k
\newcommand{\dbar}{\bar{\fd}}
\newcommand{\mD}{\mathfrak{D}} % total feedback delay
\newcommand{\BiasOne}{\textsc{Bias-1}\xspace}
\newcommand{\BiasTwo}{\textsc{Bias-2}\xspace}
\newcommand{\RegTerm}{\textsc{Reg-Term}\xspace}
\newcommand{\Bonus}{\textsc{Bonus}\xspace}

\AtBeginDocument{%
}

\renewcommand{\paragraph}[1]{\vspace{2pt}\noindent\textbf{#1}}

\begin{document}
\maketitle

\begin{abstract}
Characterizing the optimal regret for online learning with partial information, and whether it can be achieved in polynomial time, is a fundamental question in learning theory. We study this question for regret minimization in episodic Adversarial Markov Decision Processes (AMDPs) with unknown transitions and bandit feedback. In tabular AMDPs with $S$ states, $A$ actions, and horizon $H$, the best known $K$-episode regret is $\widetilde{\mathcal O}(HS\sqrt{AK})$, whose $\sqrt S$ gap to the $\Omega(H\sqrt{SAK})$ lower bound was posed as an open problem \citepname{jin2020learning}{Jin et al., ICML'20}. In linear AMDPs with $d$-dimensional features, computationally efficient algorithms without a simulator or exploratory assumptions only attain $\widetilde{\mathcal O}(K^{3/4})$ regret, whereas $\widetilde{\mathcal O}(\sqrt K)$ regret is only attained inefficiently \citepname{liu2024towards}{Liu et al., ICLR'24}.

A common challenge in both settings is to estimate the Q-function of a changing policy: past data comes from earlier policies that differ from (and are correlated with) the current one, while freezing it for future episodes worsens the dependency on $K$. We instead propose a general technique called \textit{introduced feedback delays}, where we estimate the Q-function of the policy using \emph{future} episodes while still performing policy updates, as if the feedback were \emph{delayed}. While feedback delays are usually viewed as an obstacle imposed by the environment, here they are introduced voluntarily and, perhaps surprisingly, come at no extra cost. We develop a \textit{Geometric Resampling with Rejection} (GRR) procedure to account for policy changes during the delay, and a threshold on visitation probabilities to avoid long delays on rarely visited state-action pairs. We obtain policy optimization algorithms with the following high-probability bounds:
\begin{itemize}
  \item $\widetilde{\mathcal O}(\mathrm{poly}(H)\sqrt{SAK})$ regret in tabular AMDPs, improving on the six-year-old $\widetilde{\mathcal O}(HS\sqrt{AK})$ bound and answering the above open problem in the affirmative.
  \item $\widetilde{\mathcal O}(\mathrm{poly}(d,H)\sqrt K)$ regret in linear AMDPs, giving the first efficient algorithm that closes the gap in $K$, after a five-year line of improvements starting from $K^{14/15}$ \citepname{luo2021policy}{Luo et al., NeurIPS'21}.
  \item $\widetilde{\mathcal O}(\mathrm{poly}(H)\sqrt {\mathfrak D})$ and $\widetilde{\mathcal O}(\mathrm{poly}(d,H)\sqrt {\mathfrak D})$ overheads in delayed AMDPs with total delay $\mathfrak D$, the first bounds optimal in both $S$ and $\mathfrak D$ (tabular AMDPs) and in both $K$ and $\mathfrak D$ (linear AMDPs).
\end{itemize}
\end{abstract}

\section{Introduction}
\label{sec:intro}

Markov Decision Processes (MDPs) are widely used to model reinforcement learning problems, where an agent makes sequential decisions under uncertainty. In an episodic MDP model, the agent interacts with the environment for $K$ episodes of $H$ steps each, and every episode starts from the same initial \emph{state}. At each step, she observes the current state, takes an \emph{action}, suffers a \emph{loss}, and moves to a next state drawn from a fixed distribution (the \emph{transition}) that depends on the current state and action. In each episode, she executes a \emph{policy}, that is, a randomized mapping from states to actions. There has been increasing interest in the Adversarial MDP (AMDP) \citep{even2009online,yu2009markov} model, where an oblivious adversary chooses the loss function of every episode in advance.

We study AMDPs with unknown transitions and \emph{bandit feedback}, where the agent only observes the losses of her actually visited state-action pairs, and bound the \emph{regret} against the optimal policy in hindsight. We consider \emph{tabular} AMDPs with $S<\infty$ states and $A$ actions, and \emph{linear} AMDPs, where the state space is infinite but the transition and losses are linear in known $d$-dimensional features.

For tabular AMDPs, the best known regret is $\Otil(HS\sqrt{AK})$ \citep{jin2020learning},%
\footnote{As is customary in the AMDP literature, all regret bounds outside the formal theorems omit additive terms that grow with neither the number of episodes $K$ nor the total delay $\mD$ (defined in \Cref{sec:setup-delayed}), since the MDP (hence $S,A,H,d$) is fixed when $K$ and $\mD$ grow.\label{footnote:lower-order} Some existing regret bounds are stated in terms of the number of interactions $T=HK$ or with a different definition of $S$. Throughout this paper, we have translated all existing results into our notation for a fair comparison.\label{footnote:notations}} while the lower bound is $\Omega(H\sqrt{SAK})$ \citep{jin2018q}, and this $\sqrt S$ gap has been open for six years \citep{jin2020learning}. The gap is already closed in two easier settings, namely \emph{known} transitions with bandit feedback \citep{zimin2013online} and unknown transitions with \emph{full-information} feedback \citep{pmlr-v258-tiapkin25a}, but it remains open when both difficulties are present. Linear AMDPs exhibit a similar gap, but in $K$ rather than $S$: despite a sequence of improvements \citep{luo2021policy,dai2023refined,sherman2023improved,kong2024improved}, the best computationally efficient algorithm only attains $\Otil(K^{3/4})$ regret, while the optimal $\Otil(\sqrt K)$ rate is only known to be attainable by running a linear bandit algorithm over exponentially many policies (hence inefficient) \citep{liu2024towards}.

Both gaps arise from combining bandit feedback with unknown transitions: to evaluate her current policy, the agent must estimate its expected loss starting from each state-action pair (known as the \emph{Q-function}), which requires knowing how likely the policy is to visit each pair under the unknown transition. Existing algorithms determine these visitation probabilities in one of two ways. \textit{i)} In tabular AMDPs, they infer them from \emph{past} episodes \citep{jin2020learning,luo2021policy}, which were generated by earlier policies that may rarely visit the states visited by the current policy; this mismatch induces massive uncertainty, which leads to the extra $\sqrt S$ factor. \textit{ii)} In linear AMDPs, they \emph{freeze} the policy for many future episodes to collect i.i.d.\ samples \citep{dai2023refined,liu2024towards}, a process known as \emph{epoching}; this avoids the mismatch, but the slower policy updates lead to $K^{3/4}$ regret. We thus ask:
\begin{center}
\textit{Can we evaluate a policy without \emph{i)} relying on the past or \emph{ii)} freezing the future,\\thus attaining the optimal $\bm{\sqrt S}$ in tabular AMDPs and $\bm{\sqrt K}$ in linear AMDPs?}
\end{center}

\subsection{Key Insight: Introduced Feedback Delays}
In this paper, we answer this question in the affirmative by proposing a general technique called \textit{introduced feedback delays}. Our key insight is that a policy can be evaluated \emph{after} it is executed, without freezing it in the meantime. Specifically, the agent estimates the Q-function of the policy $\pi_k$ she used in episode $k$ via the trajectories of the \emph{future} episodes $k+1,k+2,\ldots$, while still updating her policy in every episode. This idea is enabled by the following two key procedures:
\begin{enumerate}
  \item Although a later episode $\kappa>k$ executes a different policy $\pi_\kappa$, by rejection sampling, its trajectory can be treated as a fresh trajectory of $\pi_k$. This idea \emph{cannot} be applied to algorithms that rely on the past, since $\pi_k$ is computed from (and dependent on) past trajectories.\footnote{Instead, prior works on tabular AMDPs handle the mismatch between previous and current policies via optimism, that is, by considering the most favorable transition in a confidence set, whose uncertainty brings a linear $S$ factor.\label{footnote:optimism}} The likelihood ratio between $\pi_k$ and $\pi_\kappa$ on a trajectory only involves the actions taken, so it can be computed without the transition. It further stays bounded for many episodes if we control how much the policy changes in every episode. Combining rejection sampling with Geometric Resampling \citep{neu2016importance}, we can estimate $1/q^{\pi_k}(s,a)$, the inverse of the probability that $\pi_k$ visits a state-action pair $(s,a)$, without freezing the future. We call this procedure \textit{\textbf{G}eometric \textbf{R}esampling with \textbf{R}ejection} (GRR).
  \item GRR takes about $1/q^{\pi_k}(s,a)$ episodes to finish, during which we still need to update the policies. We attain this by treating $\pi_k$'s feedback as \emph{delayed}, allowing us to borrow techniques from the delayed AMDP literature \citep{lancewicki2020learning,jin2022near}. Deviating from the conventional view that delays are challenges imposed by the environment, we instead \emph{voluntarily} introduce delays, which, perhaps surprisingly, come at no extra cost -- the overhead due to delays is of the same order as that due to estimation errors. A caveat is that GRR cannot handle the pairs $(s,a)$ that are visited too rarely, as their estimates, taking $1/q^{\pi_k}(s,a)$ episodes, cannot finish while the policies remain close. But precisely because they are visited rarely, we can simply leave them unestimated and cancel the resulting errors with a large exploration bonus; this thresholding procedure is detailed in \Cref{sec:technique}.
\end{enumerate}

\begin{table}[!t]
\begin{minipage}{\textwidth}
\begin{savenotes}
\setcounter{mpfootnote}{1}
\caption[Overview of our results and comparisons with the most related ones.]{Overview of our results and comparisons with the most related ones.\textsuperscript{\itshape a}}
\footnotetext[1]{All bounds are translated into our notation (see \Cref{footnote:notations}), where $\mD=\sum_k\fd_k$ is the total feedback delay.}
\label{tab:results}
\centering
\footnotesize
\renewcommand{\arraystretch}{1.4}
\setlength{\aboverulesep}{0pt}
\setlength{\belowrulesep}{0pt}
\newcolumntype{Y}{>{\hsize=0.94\hsize\linewidth=\hsize\centering\arraybackslash}X}
\newcolumntype{W}{>{\hsize=1.06\hsize\linewidth=\hsize\centering\arraybackslash}X}
\newcommand{\gap}[1]{{\color{red!70!black}\bm{#1}}}
\newcommand{\opt}[1]{{\color{green!50!black}\bm{#1}}}
\newlength{\slotunit}
\setlength{\slotunit}{3.6\baselineskip}
\newcommand{\prior}[2]{\par\vskip0pt\nointerlineskip\noindent\parbox[c][\dimexpr6\slotunit/#1\relax][c]{\linewidth}{\centering #2}}
\newcommand{\dashsep}{\par\nointerlineskip\vbox to 0pt{\vss\hbox to\linewidth{\kern-\tabcolsep\hbox to\dimexpr\linewidth+2\tabcolsep\relax{\leaders\hbox to 3pt{\hss\vrule width 1.5pt height .2pt depth .2pt\hss}\hfill}\hss}\vss}\nointerlineskip}
\newcommand{\stack}[1]{\parbox{\linewidth}{\centering #1}}
\newcommand{\tcite}[1]{\mbox{\scriptsize\citep{#1}}}
\begin{tabularx}{\textwidth}{!{\vrule width\heavyrulewidth}Y|W!{\vrule width\heavyrulewidth}Y|W!{\vrule width\heavyrulewidth}}
\toprule
\multicolumn{2}{!{\vrule width\heavyrulewidth}c!{\vrule width\heavyrulewidth}}{Tabular AMDP} & \multicolumn{2}{c!{\vrule width\heavyrulewidth}}{Linear AMDP} \\
\hline
Bandit & Delayed Bandit & Bandit & Delayed Bandit \\
\midrule[\heavyrulewidth]
\vskip-10pt
\prior{6}{$\Otil(H^{3/2}\gap{SA^{1/4}K^{3/4}})$\\{\scriptsize(\citealp{rosenberg2019onlineb})}}\dashsep
\prior{6}{$\Otil(H\gap{S}\opt{\sqrt{AK}})$\\\tcite{jin2020learning}}\dashsep
\prior{6}{$\Otil(H^2\gap{S}\opt{\sqrt{AK}})$\\\tcite{luo2021policy}}\dashsep
\prior{6}{$\Otil(H\gap{S}\opt{\sqrt{AK}})$\\\tcite{dai2022follow}}\dashsep
\prior{6}{$\Otil(H^2\gap{S}\opt{\sqrt{AK}})$\\\tcite{dann2023best}}\dashsep
\prior{6}{$\Otil(H\gap{S}\opt{\sqrt{AK}})$\\\tcite{li2026closing}}
&
\vskip-10pt
\prior{5}{$\Otil(\gap{S}\opt{\sqrt A}\gap{K^{2/3}}+{}$\\$H^2\gap{\mD^{2/3}})$\\\tcite{lancewicki2020learning}}\dashsep
\prior{5}{$\Otil(H\gap{S}\opt{\sqrt{AK}}+{}$\\$H\gap{\sqrt[4]{SA}}\opt{\sqrt{\mD}})$\\\tcite{jin2022near}}\dashsep
\prior{5}{$\Otil(H\gap{S}\opt{\sqrt{AK}}+{}$\\$H\gap{\sqrt{SA}}\opt{\sqrt{\mD}})$\\\tcite{dai2022follow}}\dashsep
\prior{5}{$\Otil(H^2\gap{S}\opt{\sqrt{AK}}+{}$\\$H^3\opt{\sqrt{\mD}})$\\\tcite{lancewicki2023delay}}\dashsep
\prior{5}{$\Otil(H\gap{S}\opt{\sqrt{AK}}+{}$\\$H\opt{\sqrt{\mD}})$\\\tcite{zierahn2025unified}\footnote{Their bound also exhibits a linear dependency on the maximum feedback delay $\max_k\fd_k$.}}
&
\vskip-10pt
\prior{6}{$\Otil(d^2H^4\gap{K^{14/15}})$\\\tcite{luo2021policy}}\dashsep
\prior{6}{$\Otil(d^{2/3}H^{20/9}\gap{A^{1/9}K^{8/9}})$\\\tcite{dai2023refined}}\dashsep
\prior{6}{$\Otil(dH^2\gap{K^{6/7}}+{}$\\$d^{3/2}H^4\gap{K^{5/7}})$\\\tcite{sherman2023improved}}\dashsep
\prior{6}{$\Otil(d^{7/5}H^{12/5}\gap{K^{4/5}})$\\\textbf{\color{red!70!black}(inefficient)}\\\tcite{kong2024improved}\footnote{Their bound additionally requires an exploratory assumption (see \Cref{footnote:exploratory}).}}\dashsep
\prior{6}{$\Otil(d^{3/2}H^3\gap{K^{3/4}})$\\\tcite{liu2024towards}}\dashsep
\prior{6}{$\Otil(d^{7/2}H^{7/2}\opt{\sqrt{K}})$\\\textbf{\color{red!70!black}(inefficient)}\\\tcite{liu2024towards}}
&
\vskip-10pt
\prior{1}{$\Otil(d^{5/4}H^3\gap{K^{3/4}}+{}$\\$H^2\gap{\mD^{3/4}})$\\\tcite{lancewicki2023delay}\footnote{Their bound additionally assumes a simulator, which draws trajectories from any state-action pair for free.}} \\
\midrule
\rowcolor{gray!15}
\stack{$\Otil(H^{5/2}\opt{\sqrt{SAK}})$\\\textbf{(\Cref{thm:tabular})}}
& \stack{$\Otil(H^{5/2}\opt{\sqrt{SAK}}+{}$\\$H^3\opt{\sqrt{\mD}})$\\\textbf{(\Cref{thm:delayed-tabular})}}
& \stack{$\Otil(d^3H^4\opt{\sqrt{K}})$\\\textbf{(\Cref{thm:linear})}}
& \stack{$\Otil(d^3H^4\opt{\sqrt{K}}+{}$\\$d^{3/2}H^3\opt{\sqrt{\mD}})$\\\textbf{(\Cref{thm:delayed-linear})}} \\
\hline
\rowcolor{gray!15}
\stack{$\Omega(H\opt{\sqrt{SAK}})$\\\tcite{jin2018q}}
& \stack{$\Omega(H\opt{\sqrt{SAK}}+{}$\\$H\opt{\sqrt{\mD}})$\\\tcite{lancewicki2020learning}}
& \stack{$\Omega(dH^{3/2}\opt{\sqrt{K}})$\\\tcite{zhou2021nearly}}
& \stack{$\Omega(dH^{3/2}\opt{\sqrt{K}}+{}$\\$H\opt{\sqrt{\mD}})$\\\mbox{\scriptsize(\citealp{zhou2021nearly};}\\\mbox{\scriptsize\citealp{lancewicki2020learning})}} \\
\bottomrule
\end{tabularx}
\end{savenotes}
\end{minipage}
\end{table}
Our introduced feedback delays are the key to all of our improvements, and we believe that this technique is of independent interest for other AMDP problems where a changing policy needs to be evaluated without freezing the future. Building upon introduced feedback delays, we present computationally efficient AMDP algorithms with the following results (summarized in \Cref{tab:results}):
\begin{enumerate}
  \item We start with tabular AMDPs, where existing methods \citep{luo2021policy} compute the visitation probability of the current policy (that is, its \emph{occupancy measure}) from a confidence set of transitions built from past trajectories, whose uncertainty sums to an $S$ factor (see \Cref{footnote:optimism} for details). We instead directly estimate the occupancy measure from future episodes via GRR, which avoids this uncertainty and attains $\Otil(H^{5/2}\sqrt{SAK})$ regret (\Cref{thm:tabular}); the confidence sets only enter a $K$-independent term. This is the \emph{first} bound to match the $\Omega(H\sqrt{SAK})$ lower bound \citep{jin2018q} in $S$, $A$, and $K$, answering the open problem posed by \citet{jin2020learning}.

  \item We next find that the same idea also removes the need for freezing the future in linear AMDPs. The best efficient algorithm freezes each policy for about $\sqrt K$ episodes to estimate its covariance matrix, which leads to $\Otil(d^{3/2}H^3K^{3/4})$ regret \citep{liu2024towards}. We instead estimate the inverse covariance matrix from future episodes while still updating policies, via \textit{\textbf{M}atrix \textbf{G}eometric \textbf{R}esampling with \textbf{R}ejection} (MGRR), which adds rejection sampling to Matrix Geometric Resampling \citep{neu2020efficient}, separately on $\O(\log K)$ feature \textit{regions} that keep the estimates bounded. This leads to $\Otil(d^3H^4\sqrt K)$ regret (\Cref{thm:linear}), the \emph{first} $\sqrt K$-order regret of an efficient algorithm without simulators or exploratory assumptions, matching the $\Omega(\sqrt K)$ lower bound \citep{zhou2021nearly} in $K$.

  \item While our motivation has no feedback delays, our analysis naturally generalizes to delayed AMDPs, where the losses of episode $k$ arrive at the end of episode $k+\fd_k$ for an adversarial and unknown $\fd_k$. We obtain $\Otil(H^{5/2}\sqrt{SAK}+H^3\sqrt\mD)$ and $\Otil(d^3H^4\sqrt K+d^{3/2}H^3\sqrt\mD)$ regret for tabular and linear AMDPs, respectively, where $\mD=\sum_k\fd_k$ (\Cref{thm:delayed-tabular,thm:delayed-linear}). These are the \emph{first} bounds optimal in $S$, $A$, $K$, and $\mD$ for tabular AMDPs, where all previous leading terms are linear in $S$ \citep{lancewicki2020learning,jin2022near,zierahn2025unified}, and in $K$ and $\mD$ for linear AMDPs, where the only known bound requires a simulator \citep{lancewicki2023delay}.
\end{enumerate}

\subsection{Related Work}
\label{sec:related}

\paragraph{Learning in Tabular AMDPs.}
Learning in AMDPs was initiated by \citet{even2009online} and \citet{yu2009markov}. With known transitions, \citet{zimin2013online} attain the optimal $\Otil(H\sqrt K)$ and $\Otil(\sqrt{HSAK})$ regret under full-information and bandit feedback, respectively. With unknown transitions and full-information feedback, \citet{rosenberg2019online} attain $\Otil(HS\sqrt{AK})$ regret, which \citet{pmlr-v258-tiapkin25a} improve to $\Otil(H^{7/2}\sqrt{SAK})$. With unknown transitions and bandit feedback, the best upper and lower bounds are $\Otil(HS\sqrt{AK})$ \citep{jin2020learning} and $\Omega(H\sqrt{SAK})$ \citep{jin2018q}, respectively. These methods (except for \citealt{pmlr-v258-tiapkin25a}) all use the occupancy measure framework: in every episode, they solve a convex program over a polytope whose size scales with the number of state-action pairs, which might be costly.

Motivated by this, the policy optimization framework (see also \Cref{sec:po}) was proposed, which replaces this global optimization with local updates at every state. With unknown transitions and bandit feedback, \citet{shani2020optimistic} give the first policy optimization algorithm with sublinear ($K^{2/3}$) regret for tabular AMDPs, which \citet{luo2021policy} improve to $\Otil(H^2S\sqrt{AK})$. Concurrent to our work, \citet{li2026closing} attain $\Otil(HS\sqrt{AK})$ regret with policy optimization, improving the $\Otil(H^2S\sqrt{AK})$ bound by a factor of $H$.

Another line of research learns unknown-transition AMDPs via the Follow-the-Perturbed-Leader (FTPL) framework, where the best known rate is $\Otil(HS\sqrt{AK})$ under both full-information feedback \citep{neu2012adversarial,wang2020refined} and bandit feedback \citep{dai2022follow}. FTPL-based algorithms are computationally cheaper, but require more specialized analyses and are less flexible in algorithm design.

\paragraph{Learning in Linear AMDPs.}
For linear AMDPs with bandit feedback, \citet{neu2021online} give the first sublinear-regret algorithm, attaining $\sqrt K$ regret with known transitions and under a strong exploratory assumption. With unknown transitions, \citet{luo2021policy} drop both assumptions and attain $K^{2/3}$ regret with a \emph{simulator}, allowing the agent to sample trajectories from any state-action pair, and $K^{14/15}$ regret without one. \citet{dai2023refined} improve these to $\sqrt K$ and $K^{8/9}$, respectively. Without a simulator, \citet{sherman2023improved} attain $K^{6/7}$, and \citet{kong2024improved} attain $K^{4/5}$ under a weaker exploratory assumption. The currently best bounds are due to \citet{liu2024towards}: $\Otil(d^{3/2}H^3K^{3/4})$ with an efficient policy optimization algorithm, and $\Otil(d^{7/2}H^{7/2}\sqrt K)$ with an inefficient algorithm viewing the exponentially many policies as arms of a linear bandit. Hence, it has long remained open whether $\sqrt K$ regret is attainable by an efficient algorithm (without exploratory assumptions or simulators), which we answer in the affirmative.\footnote{The answer is known to be affirmative under full-information feedback \citep{pmlr-v235-sherman24a,cassel2024warm}, but has long been open under bandit feedback. When preparing this draft, we were made aware of a concurrent work by \citet{liu2026computationally}, which also efficiently attains $\sqrt K$ regret in linear AMDPs. Their algorithm builds upon the occupancy measure framework and calls a linear optimization oracle $\Omega(K^2)$ times, which is costly, and their regret bound only holds in expectation. In contrast, we use policy optimization with closed-form updates and no oracle calls, our regret bound holds with high probability, and our generic technique also applies to tabular AMDPs and delayed AMDPs.}

Another related model is the linear mixture AMDP, where the transition is a linear combination of several known base models. Under full-information feedback, \citet{cai2020provably} attain $\Otil(dH^2\sqrt K)$, and \citet{he2022near} attain $\Otil(dH^{3/2}\sqrt K)$ with a matching lower bound. Under bandit feedback \citep{zhao2023learning,li2024improved,cheng2026policy,li2026closing}, the best known upper bound is $\Otil(d\sqrt{HS^3K}+\sqrt{HSAK})$ \citep{li2024improved}, and \citet{zhao2023learning} prove an $\Omega(dH\sqrt K+\sqrt{HSAK})$ lower bound. Although both models use linear function approximation, linear mixture AMDPs are incomparable to linear AMDPs, so none of these implies a $\sqrt K$-regret algorithm for linear AMDPs. Nevertheless, the $\Omega(dH^{3/2}\sqrt K)$ lower bound of \citet{zhou2021nearly} for linear mixture MDPs also applies to linear AMDPs due to its special construction.

\paragraph{Learning in AMDPs with Feedback Delays.}
For delayed tabular AMDPs with bandit feedback, the lower bound is $\Omega(H\sqrt{SAK}+H\sqrt\mD)$ \citep{lancewicki2020learning}, but all upper bounds are linear in $S$: $\Otil(HS\sqrt{AK}+H(SA)^{1/4}\sqrt\mD)$ \citep{jin2022near}, $\Otil(H^2S\sqrt{AK}+H^3\sqrt\mD)$ \citep{lancewicki2023delay}, and $\Otil(HS\sqrt{AK}+H\sqrt\mD)$ plus a term linear in the maximum delay \citep{zierahn2025unified}. For delayed linear AMDPs, the only known result is an $\Otil(d^{5/4}H^3K^{3/4}+H^2\mD^{3/4})$ bound \citep{lancewicki2023delay} that requires a simulator (although it applies to the more general linear-Q MDPs). While our main model has no feedback delays, our analysis naturally accommodates them, giving $\Otil(H^{5/2}\sqrt{SAK}+H^3\sqrt\mD)$ regret for delayed tabular AMDPs, the \emph{first} to match the lower bound in $S$, $A$, $K$, and $\mD$ simultaneously; and $\Otil(d^3H^4\sqrt K+d^{3/2}H^3\sqrt\mD)$ regret for delayed linear AMDPs, the \emph{first} that is optimal in both $K$ and $\mD$.

\section{Preliminaries}\label{sec:prelim}
\paragraph{Notation.}
For $N\in\N$, we use $[N]$ to denote the set $\{1,2,\ldots,N\}$. For a (possibly infinite) set $\mathcal X$, we use $\Delta(\mathcal X)$ to denote the probability simplex over $\mathcal X$. For a random event $\mathcal E$, we use $\ind[\mathcal E]$ to denote its indicator. For $x\in\R$, let $[x]_+\triangleq\max\{x,0\}$. For a positive semi-definite matrix $\Sigma$, let $\Sigma^\dagger$ be its Moore--Penrose pseudo-inverse, $\Sigma^{-1/2}$ be the square root of $\Sigma^\dagger$, and $\lVert x\rVert_\Sigma\triangleq\sqrt{x^\top\Sigma x}$ for $x\in\R^d$. We use $\Otil(\cdot)$, $\Omegatil(\cdot)$, and $\Thetatil(\cdot)$ to hide all factors that are polylogarithmic in $S$, $A$, $H$, $K$, $d$, and $1/\delta$, where $\delta$ is the failure probability.

\subsection{Episodic Adversarial MDPs}\label{sec:setup-episodic}
An episodic Adversarial MDP (AMDP) is defined by a tuple $\mathcal M=(\cS,\cA,\PP,\{\ell_k\}_{k=1}^K)$, where $\cS$ is the (possibly infinite) state space, $\cA$ is the finite action space with $A=\lvert\cA\rvert$, $\PP\colon\cS\times\cA\to\Delta(\cS\cup\cS_H)$ is the transition where $\cS_H$ is the set of all terminal states, and $\ell_k\colon\cS\times\cA\to[0,1]$ is the loss function of episode $k$, which is chosen arbitrarily by an oblivious adversary before the game starts. Without loss of generality, we assume that the state space is layered as $\cS=\cS_0\cup\cS_1\cup\cdots\cup\cS_{H-1}$ \citep{jin2020learning}, where $\cS_h\cap\cS_{h'}=\varnothing$ for any $h\ne h'$, the first layer $\cS_0=\{s_0\}$ only contains the initial state $s_0$, and transitions only happen from one layer to the next one, that is, $\PP(\cdot\mid s,a)\in\Delta(\cS_{h+1})$ for all $s\in\cS_h$ and $a\in\cA$. In tabular AMDPs with finite state spaces, we use $S=\sum_{h=0}^{H-1} \lvert \cS_h\rvert<\infty$ to denote the number of non-terminal states.

The agent interacts with the environment for $K$ episodes, each consisting of $H$ steps. In episode $k\in[K]$, the agent starts from the initial state $s_{k,0}=s_0$. At every layer $h=0,1,\ldots,H-1$, she observes the current state $s_{k,h}\in\cS_h$, chooses an action $a_{k,h}\in\cA$, suffers and observes the loss $\ell_k(s_{k,h},a_{k,h})$, and then transitions to a new state $s_{k,h+1}$ that is independently sampled from $\PP(\cdot\mid s_{k,h},a_{k,h})$. That is, she only observes the losses along her own trajectory (i.e., bandit feedback). The transition $\PP$ is unknown, and the agent has to learn it from her own trajectories. We use
\begin{equation*}
  L_{k,h}\triangleq\sum_{h'=h}^{H-1}\ell_k(s_{k,h'},a_{k,h'})
\end{equation*}
to denote the loss suffered from layer $h$ to the end of episode $k$. Let $\{\cF_k\}_{k=0}^K$ be the natural filtration such that $\cF_k$ contains the history of episodes $1,2,\ldots,k$, including the agent's internal randomness, and we write $\E_k[\cdot]\triangleq\E[\cdot\mid\cF_k]$.

\paragraph{Policies, Q-Functions, and Occupancy Measures.}
A (Markov) policy $\pi$ of the agent is a mapping from $\cS$ to $\Delta(\cA)$, and we use $\Pi$ to denote the set of all such policies. For a policy $\pi$ and a transition $P'$ over the layered state space, a \emph{trajectory} of $\pi$ under $P'$ is the random sequence $(s_0,a_0,s_1,a_1,\ldots,s_{H-1},a_{H-1})$ generated by $a_h\sim\pi(\cdot\mid s_h)$ and $s_{h+1}\sim P'(\cdot\mid s_h,a_h)$ for all $0\le h<H$. For $0\le h<H$, its prefix $(s_0,a_0,\ldots,s_h,a_h)$ is its \emph{partial trajectory} up to layer $h$. For an episode $k\in[K]$ and a policy $\pi\in\Pi$, the \emph{state-value function} (or V-function for short) $V_k^\pi(s)$ is the expected loss suffered by following $\pi$ from state $s$ until the end of the episode, and the \emph{state-action-value function} (or Q-function for short) $Q_k^\pi(s,a)$ is the same quantity when starting from the state-action pair $(s,a)$. Formally, we let $V_k^\pi(s)\triangleq0$ for all terminal states $s\in\cS_H$, and define the Q- and V-functions backward from the last layer via the Bellman equations:
\begin{equation}\label{eq:Q-function}
  Q_k^\pi(s,a)\triangleq\ell_k(s,a)+\E_{s'\sim \PP(\cdot \mid s,a)}\bigl[V_k^\pi(s')\bigr],\;V_k^\pi(s)\triangleq\E_{a\sim \pi(\cdot \mid s)}\bigl[Q_k^\pi(s,a)\bigr],\;\forall s\in\cS,a\in\cA.
\end{equation}
Another important concept is the occupancy measure \citep{altman1999constrained,neu2012adversarial}, the probability that a policy $\pi$ visits a state-action pair under a transition $P'$. Formally, for a layer-$h$ state $s\in \cS_h$ and action $a\in \cA$,
\begin{equation}\label{eq:occupancy measure}
  q^{P',\pi}(s,a)\triangleq\Pr_{a_{h'}\sim \pi(\cdot\mid s_{h'}),\,s_{h'+1}\sim P'(\cdot \mid s_{h'},a_{h'})}\left\{s_h=s,a_h=a\right\},\;q^{P',\pi}(s)\triangleq\sum_{a\in \cA}q^{P',\pi}(s,a).
\end{equation}
Since the action at $s$ is drawn from $\pi(\cdot\mid s)$, we have $q^{P',\pi}(s,a)=q^{P',\pi}(s)\,\pi(a\mid s)$. We abbreviate the occupancy measure under the true transition $\PP$ as $q^\pi\triangleq q^{\PP,\pi}$. As every layer is visited exactly once in each episode, the V-function of $\pi$ is linear in its occupancy measure: $V_k^\pi(s_0)=\sum_{s,a}q^\pi(s,a)\ell_k(s,a)$. All of $V_k^\pi$, $Q_k^\pi$, and $q^\pi$ depend on the unknown $\PP$, so the agent cannot compute them.

\paragraph{Regret.}
The agent can decide her policies adaptively and randomly: for each episode $k\in [K]$, the agent deploys a policy $\pi_k\in \Pi$ that is $\cF_{k-1}$-measurable up to any amount of internal randomness. Her expected loss in episode $k$, conditional on $\pi_k$, is therefore $V_k^{\pi_k}(s_0)$ according to \Cref{eq:Q-function}.

The goal of the agent is to minimize the regret, the difference between her cumulative loss and that of the optimal policy in hindsight. Specifically, let $\pi^\star\in\argmin_{\pi\in\Pi}\sum_{k=1}^KV_k^\pi(s_0)$, which is \emph{not} randomized since all loss functions are determined by an oblivious adversary. The regret is then:
\begin{equation}\label{def:regret}
  \mR_K\triangleq\sum_{k=1}^KV_k^{\pi_k}(s_0)-\sum_{k=1}^KV_k^{\pi^\star}(s_0),
\end{equation}
which is a random variable since $\pi_k$ depends on the random observations and the internal randomness of the agent. In this paper, we provide high-probability bounds on $\mR_K$ that hold with probability at least $1-\delta$ for any given $\delta\in(0,1)$. Such bounds are strictly stronger than those on the expected regret, since $\mR_K\le HK$ almost surely, so $\delta=1/(2K)$ turns them into bounds on $\E[\mR_K]$.

\subsection{Linear AMDPs}\label{sec:setup-linear}
Linear MDPs were proposed by \citet{yang2019sample} and \citet{jin2020provably} for reinforcement learning with large state spaces and fixed rewards, and were later studied with adversarial losses and bandit feedback \citep{neu2021online,luo2021policy}. In a linear AMDP, the state space $\cS$ can be infinite, but the transition and the losses are linear in some known $d$-dimensional features.

\begin{definition}[Linear AMDP]\label{def:linear}
  In a linear AMDP, every state-action pair $(s,a)$ has a known feature $\phi(s,a)\in\R^d$ with $\lVert\phi(s,a)\rVert_2\le1$. There is an unknown mapping $\psi\colon\cS\to\R^d$ for the transitions, and $K\times H$ vectors $\zeta_{k,h}\in\R^d$ for the losses, chosen by the oblivious adversary in advance, such that
  \begin{alignat*}{2}
    \PP(s'\mid s,a)&=\langle\phi(s,a),\psi(s')\rangle,&&\forall s\in\cS_h,a\in\cA,s'\in\cS_{h+1},0\le h<H-1,\\
    \ell_k(s,a)&=\langle\phi(s,a),\zeta_{k,h}\rangle,&&\forall s\in\cS_h,a\in\cA,k\in[K],0\le h<H.
  \end{alignat*}
  As is standard in the literature, $\lVert\sum_{s'\in\cS_{h+1}}\lvert\psi(s')\rvert\rVert_2\le\sqrt d$ (with $\lvert\cdot\rvert$ element-wise) and $\lVert\zeta_{k,h}\rVert_2\le\sqrt d$.
\end{definition}

When $\cS$ is uncountable, $\PP(\cdot\mid s,a)$ is a probability measure rather than a probability for every state. Then $\psi=(\psi_1,\ldots,\psi_d)$ consists of $d$ signed measures over $\cS_{h+1}$, i.e., measures that may take negative values, such that $\PP(\mathcal B\mid s,a)=\langle\phi(s,a),\psi(\mathcal B)\rangle$ for every measurable $\mathcal B\subseteq\cS_{h+1}$ \citep{jin2020provably}. Sums over states then become integrals, and $\sum_{s'}\lvert\psi(s')\rvert$ becomes the vector of total variations $(\lvert\psi_i\rvert(\cS_{h+1}))_{i=1}^d$. For simplicity, we write sums throughout. Tabular AMDPs are the special case with $d=SA$ and one-hot features $\phi(s,a)=\mathbf e_{(s,a)}$, where $\psi(s')=(\PP(s'\mid s,a))_{(s,a)}$ and $\zeta_{k,h}=(\ell_k(s,a))_{(s,a)}$. As $\phi$ is known, the agent can compute $\phi(s,a)$ for every visited state $s$ and every action $a\in\cA$, and she still only observes the losses along her own trajectories.

\paragraph{Linear Q-Functions and Covariance Matrices.}
By \Cref{def:linear} and \Cref{eq:Q-function}, the Q-function of every policy is also linear in $\phi$ \citep{jin2020provably}. Specifically, for every $k\in[K]$, $\pi\in\Pi$, and $0\le h<H$, we have $Q_k^\pi(s,a)=\langle\phi(s,a),w_{k,h}^\pi\rangle$ for all $s\in\cS_h$ and $a\in\cA$, where $w_{k,h}^\pi\triangleq\zeta_{k,h}+\sum_{s'\in\cS_{h+1}}\psi(s')V_k^\pi(s')$ (without the sum if $h=H-1$) satisfies $\lVert w_{k,h}^\pi\rVert_2\le H\sqrt d$ as $0\le V_k^\pi(s')\le H-1$. Hence, every linear AMDP is a linear-Q MDP \citep[Assumption~1]{luo2021policy}. For a policy $\pi$ and a layer $h$, the \emph{covariance matrix} of $\pi$ at layer $h$ is
\begin{equation}\label{eq:covariance}
  \Sigma_h^\pi\triangleq\sum_{s\in\cS_h,a\in\cA}q^\pi(s,a)\phi(s,a)\phi(s,a)^\top=\E\bigl[\phi(s_h,a_h)\phi(s_h,a_h)^\top\bigr],
\end{equation}
where $(s_h,a_h)$ is the layer-$h$ state-action pair of a trajectory of $\pi$ under $\PP$ \citep{luo2021policy,dai2023refined}. As $\lVert\phi(s,a)\rVert_2\le1$, we have $0\preceq\Sigma_h^\pi\preceq I$, and in tabular AMDPs, $\Sigma_h^\pi$ is diagonal with entries $q^\pi(s,a)$.

\subsection{Delayed AMDPs}\label{sec:setup-delayed}
Learning with delayed feedback has been studied in online learning \citep{mesterharm2005line,joulani2013online} and adversarial bandits \citep{cesa2016delay,thune2019nonstochastic,bistritz2019online}, and was extended to AMDPs by \citet{lancewicki2020learning} and \citet{jin2022near}. We follow the model of the latter, where only the losses are delayed.

\begin{definition}[Delayed AMDP]\label{def:delayed}
  A delayed AMDP is an AMDP with integer delays $\fd_1,\ldots,\fd_K\ge0$, which the oblivious adversary chooses together with the losses before the game starts. In episode $k\in[K]$, the agent still observes her trajectory $(s_{k,h},a_{k,h})_{h=0}^{H-1}$ during the episode, but the losses $\{\ell_k(s_{k,h},a_{k,h})\}_{h=0}^{H-1}$ along it are only revealed at the end of episode $k+\fd_k$, and never if $k+\fd_k>K$. Hence, $\pi_k$ may only depend on the trajectories of episodes $1,\ldots,k-1$, the losses of episodes $j<k$ with $j+\fd_j\le k-1$, and the agent's internal randomness. We write $\mD\triangleq\sum_{k=1}^K\fd_k$ for the total delay.
\end{definition}

We write $\fd_k$ instead of the more common $d_k$ to avoid confusion with the feature dimension $d$. As in \citet{thune2019nonstochastic}, the delays can be arbitrarily large, and our algorithm only depends on their total $\mD$ rather than a maximum. The regret is still defined as in \Cref{def:regret}, and taking $\mD=0$ recovers \Cref{sec:setup-episodic}. For simplicity, we assume that $\mD$ is known; the standard doubling trick removes this assumption \citep[see][]{thune2019nonstochastic,bistritz2019online}. \citet{lancewicki2020learning} also study delayed trajectories, which we leave for future work.

\subsection{Policy Optimization and Exploration Bonuses}\label{sec:po}

We use the policy optimization framework to minimize $\mR_K$ \citep[see, e.g.,][]{luo2021policy}. Its foundation is the performance difference lemma \citep{kakade2002approximately}, which decomposes $\mR_K$ into a weighted sum over states:
\begin{equation}\label{eq:pdl}
  \mR_K=\sum_{s\in\cS}q^\star(s)\sum_{k=1}^K\sum_{a\in\cA}\bigl(\pi_k(a\mid s)-\pi^\star(a\mid s)\bigr)Q_k^{\pi_k}(s,a),
\end{equation}
where $q^\star\triangleq q^{\pi^\star}$ is shorthand for the occupancy measure of $\pi^\star$.

For every state $s\in \cS$, the inner sum is exactly the regret of a \emph{multi-armed bandit} problem with $A$ actions over the $K$ episodes, where each action $a\in \cA$ suffers the \emph{bandit loss} $Q_k^{\pi_k}(s,a)$ (not to be confused with the MDP loss function $\ell_k(s,a)$) in episode $k$. Policy optimization algorithms then run a bandit algorithm at every state -- for example, exponential weights (Hedge) \citep{freund1997decision} -- which sets
\begin{equation*}
  \pi_k(a\mid s)\propto\exp\left(-\eta\sum_{\kappa<k}g_\kappa(s,a)\right),\;\forall s\in\cS,a\in\cA,
\end{equation*}
for some learning rate $\eta>0$. Here, at each state $s$, $g_\kappa(s,a)$ is the estimated bandit loss of action $a$ in episode $\kappa$, whose design is the key ingredient of policy optimization algorithms.

\paragraph{Bonuses.}
As $Q_k^{\pi_k}$ is unknown, we must construct an estimate $\Qhat_k$ to feed into the bandit algorithm, whose estimation error causes extra regret. For example, the importance-weighted estimate divides the observed loss $L_{k,h}$ from layer $h$ onward by $q^{\pi_k}(s,a)$:
\begin{equation}\label{eq:iw}
  \Qhat_k(s,a)\triangleq\frac{\ind[(s_{k,h},a_{k,h})=(s,a)]\,L_{k,h}}{q^{\pi_k}(s,a)},\;\forall s\in\cS_h,a\in\cA,0\le h<H.
\end{equation}
The episode-$k$ regret of exponential weights at state $s$ depends on the second moment of $\Qhat_k$ as
\begin{equation*}
  \eta\sum_{a\in\cA}\pi_k(a\mid s)\E_{k-1}\bigl[\Qhat_k(s,a)^2\bigr]\le\eta\sum_{a\in\cA}\pi_k(a\mid s)\frac{H^2}{q^{\pi_k}(s,a)}=\frac{\eta H^2A}{q^{\pi_k}(s)}\triangleq b_k(s),
\end{equation*}
where the last equality uses $q^{\pi_k}(s,a)=q^{\pi_k}(s)\pi_k(a\mid s)$. Such functions $b_k\colon\cS\to\R_{\ge0}$ are commonly known as the (exploration) \emph{bonuses}, whose name will be explained shortly. As \Cref{eq:pdl} weights the regret of exponential weights at state $s$ by $q^\star(s)$, these terms add up to
\begin{equation*}
  \sum_{k=1}^K \sum_{s\in \cS} q^\star(s)\left (\eta\sum_{a\in\cA}\pi_k(a\mid s)\E_{k-1}\bigl[\Qhat_k(s,a)^2\bigr]\right )\le \sum_{k=1}^K\sum_{s\in\cS}q^\star(s)\,b_k(s).
\end{equation*}
When using \Cref{eq:iw}, this equals $\eta H^2A\sum_k\sum_sq^\star(s)/q^{\pi_k}(s)$, which is arbitrarily large when $\pi^\star$ visits a state that $\pi_k$ rarely visits. In contrast, if the bonuses were weighted by $q^{\pi_k}$ instead of $q^\star$, we would only suffer $\sum_sq^{\pi_k}(s)b_k(s)\le\eta H^2SA$ per episode. To handle the mismatch between $q^\star$ and $q^{\pi_k}$, let $B_k$ be the Q-function of $\pi_k$ for the loss $b_k$, that is, $B_k=0$ on $\cS_H$ and
\begin{equation}\label{eq:bonus-q}
  B_k(s,a)=b_k(s)+\sum_{s'\in\cS_{h+1}}\PP(s'\mid s,a)\sum_{a'\in\cA}\pi_k(a'\mid s')B_k(s',a'),\;\forall s\in\cS_h,a\in\cA,0\le h<H.
\end{equation}
Then, we feed $g_k\triangleq\Qhat_k-B_k$ into the bandit algorithm at every state. This $-B_k$ makes rarely visited states (whose $b_k$ is larger) more attractive and thus encourages exploration (hence the name). The following now-standard lemma shows that this resolves the mismatch between $q^\star$ and $q^{\pi_k}$.
\begin{lemma}[Exploration Bonuses]\label{lem:bonus}
  If $B_k$ satisfies \Cref{eq:bonus-q} for all $k\in[K]$, then
  \begin{equation*}
    \mR_K\le\sum_{s\in\cS}q^\star(s)\Biggl(\sum_{k=1}^K\sum_a\bigl(\pi_k(a\mid s)-\pi^\star(a\mid s)\bigr)\bigl(Q_k^{\pi_k}-B_k\bigr)(s,a)-\sum_{k=1}^Kb_k(s)\Biggr)+\sum_{k=1}^K\sum_{s\in\cS}q^{\pi_k}(s)b_k(s),
  \end{equation*}
  where the first sum in the parentheses is the bandit regret with respect to $Q_k^{\pi_k}-B_k$ at state $s$. If $B_k$ only satisfies \Cref{eq:bonus-q} with ``$=$'' replaced by ``$\ge$'', then the same bound holds with the last term replaced by $\sum_{k=1}^K\sum_a\pi_k(a\mid s_0)B_k(s_0,a)$.
\end{lemma}

Hence, the main challenge is estimating the Q-function. The most natural importance-weighted estimate in \Cref{eq:iw} depends on the unknown transition $\PP$ and cannot be directly used. While existing algorithms handle this by relying on past episodes \citep{jin2020learning,luo2021policy} or freezing the future \citep{liu2024towards}, we instead directly estimate the unknown $1/q^{\pi_k}(s,a)$ in \Cref{eq:iw} using \emph{future} episodes. Since such an estimate only becomes available episodes later, we introduce feedback delays \emph{on purpose}. This generic idea works in tabular (\Cref{sec:tech-tabular}), linear (\Cref{sec:tech-linear}), and delayed (\Cref{sec:tech-delay}) AMDPs.

\section{Technical Overview: Introduced Feedback Delays}\label{sec:overview}

In this section, we explain introduced feedback delays, first in tabular AMDPs (\Cref{sec:tech-tabular}), then in linear AMDPs (\Cref{sec:tech-linear}), and finally with actual feedback delays (\Cref{sec:tech-delay}). The formal algorithms and guarantees are in \Cref{sec:tabular,sec:linear,sec:delay}.

\subsection{Introduced Feedback Delays in Tabular AMDPs: GRR and Thresholding}\label{sec:technique}\label{sec:tech-tabular}

Fix an episode $k$ and a state-action pair $(s,a)\in\cS_h\times\cA$. The importance-weighted estimate $\Qhat_k(s,a)$ in \Cref{eq:iw} uses $1/q^{\pi_k}(s,a)$, which depends on the unknown $\PP$. Instead of using an upper confidence bound as in previous tabular AMDP works \citep{jin2020learning,luo2021policy}, we estimate $1/q^{\pi_k}(s,a)$ directly.

\citet{neu2016importance} propose Geometric Resampling (GR), which applies to any sequence of independent \textit{attempts}, each of which \textit{succeeds} with the same probability. For example, if the policy $\pi_k$ remains unchanged for several future episodes (i.e., freezing the future), then each attempt would be the trajectory of a subsequent episode $\kappa>k$, and success would mean visiting $(s,a)$ again (so the success probability is $q^{\pi_k}(s,a)$). The number of attempts until the first success is then geometric with mean $1/q^{\pi_k}(s,a)$ and second moment at most $2/q^{\pi_k}(s,a)^2$. However, we instead keep updating the policies, so attempts no longer share the same success probability $q^{\pi_k}(s,a)$, making GR inapplicable.

Hence, we propose \textbf{G}eometric \textbf{R}esampling with \textbf{R}ejection (GRR), which uses future episodes $\kappa>k$ to generate attempts. Since $\pi_\kappa$ differs from $\pi_k$ (so their trajectories are not identically distributed), we use \emph{rejection sampling}: we accept each future episode with a probability proportional to the likelihood ratio between $\pi_k$ and $\pi_\kappa$, such that accepted trajectories are distributed as those of $\pi_k$.

\begin{lemma}[GRR Guarantee]\label{lem:rejection}
  Let $k<\kappa$ and $0\le h<H$, and suppose that $\prod_{h'=0}^h\frac{\pi_k(a_{h'}\mid s_{h'})}{\pi_\kappa(a_{h'}\mid s_{h'})}\le2$ for every partial trajectory $(s_0,a_0,\ldots,s_h,a_h)$. Let episode $\kappa$ be accepted with probability
  \begin{equation}\label{eq:rejection accept probability}
    \alpha_{k,\kappa}\triangleq\frac12\prod_{h'=0}^h\frac{\pi_k(a_{\kappa,h'}\mid s_{\kappa,h'})}{\pi_\kappa(a_{\kappa,h'}\mid s_{\kappa,h'})},
  \end{equation}
  using independent randomness, which lies in $[0,1]$ by this assumption (we omit its dependence on $h$ from the notation). Given $\cF_{\kappa-1}$, episode $\kappa$ is accepted with probability $\E_{\kappa-1}[\alpha_{k,\kappa}]=\frac12$. Further, given $\cF_{\kappa-1}$ and that episode $\kappa$ is accepted, its partial trajectory up to layer $h$ is distributed as that of a trajectory of $\pi_k$ under $\PP$. Finally, given $\cF_{\kappa-1}$, the probability that episode $\kappa$ is accepted and also visits $(s,a)\in\cS_h\times\cA$ (a success) is $q^{\pi_k}(s,a)/2$, that is,
  \begin{equation*}
    \E_{\kappa-1}\bigl[\ind[(s_{\kappa,h},a_{\kappa,h})=(s,a)]\,\alpha_{k,\kappa}\bigr]=\frac{q^{\pi_k}(s,a)}2.
  \end{equation*}
\end{lemma}
Note that GRR requires no knowledge of $\PP$, as the transitions cancel out in the likelihood ratio between $\pi_k$ and $\pi_\kappa$, namely $\Pr_{\pi_k}(s_0,a_0,\ldots,s_h,a_h)/\Pr_{\pi_\kappa}(s_0,a_0,\ldots,s_h,a_h)=\prod_{h'=0}^h\frac{\pi_k(a_{h'}\mid s_{h'})}{\pi_\kappa(a_{h'}\mid s_{h'})}=2\alpha_{k,\kappa}$.

To summarize, for each $(s,a)$ visited by $\pi_k$ in episode $k$, we start a GRR estimate for $(s,a)$. It estimates $1/q^{\pi_k}(s,a)$, which in turn gives an estimate of $Q_k^{\pi_k}(s,a)$. Every episode $\kappa>k$ constitutes an attempt, which is accepted with probability $\alpha_{k,\kappa}$ in \Cref{eq:rejection accept probability}, using independent randomness. A GRR estimate succeeds when an accepted attempt visits $(s,a)$.

Let $\what_k(s,a)$ be the number of attempts (accepted or rejected) until the estimate succeeds, with $\what_k(s,a)=0$ if $(s,a)$ is not visited in episode $k$ (in which case we do not start a GRR estimate). By \Cref{lem:rejection}, every attempt succeeds with probability $q^{\pi_k}(s,a)/2$, regardless of the history. Therefore, replacing $1/q^{\pi_k}(s,a)$ in \Cref{eq:iw} by $\what_k(s,a)/2$, we obtain the following GRR estimate:
\begin{align}\label{eq:grr}
  \Qhat_k(s,a)&\triangleq\ind[(s_{k,h},a_{k,h})=(s,a)]\,L_{k,h}\frac{\what_k(s,a)}{2},
\end{align}
which is unbiased ($\E_{k-1}\bigl[\Qhat_k(s,a)\bigr]=Q_k^{\pi_k}(s,a)$) with a second moment $\E_{k-1}\bigl[\Qhat_k(s,a)^2\bigr]\le\frac{2H^2}{q^{\pi_k}(s,a)}$.

\paragraph{Splitting Estimates and Thresholding.}
\Cref{lem:rejection} requires $\pi_\kappa$ to be close to $\pi_k$ during all the $\approx2/q^{\pi_k}(s,a)$ attempts (namely, the likelihood ratio $2\alpha_{k,\kappa}$ is at most $2$). Hence, the loss fed to exponential weights in every episode must be small, so that the policy only changes slightly. However, $\Qhat_k(s,a)$ is typically of order $H/q^{\pi_k}(s,a)$, so feeding it at once would change $\pi_k(a\mid s)$ a lot.

We handle this by dividing $\Qhat_k(s,a)$ into smaller \textit{pieces}. Every attempt, accepted or not, adds $L_{k,h}/2$ to the \textit{received loss} $g_\kappa(s,a)$ of its episode $\kappa$ (that is, one piece per attempt counted by $\what_k(s,a)$ in \Cref{eq:grr}). Exponential weights runs on the received losses as $\pi_k(a\mid s)\propto\exp(-\eta\sum_{\kappa<k}g_\kappa(s,a))$. Because a GRR estimate for $(s,a)$ is started with probability $q^{\pi_k}(s,a)$ and lasts about $2/q^{\pi_k}(s,a)$ episodes, in expectation, every $(s,a)$ pair only has $\Otil(1)$ unfinished estimates in any episode. Therefore, $g_\kappa(s,a)$ is of order $\Otil(H^2)$, and consequently $\log\pi_\kappa(a\mid s)$ changes by only $\Otil(\eta H^2)$. Over a partial trajectory of at most $H$ layers, the likelihood ratio therefore stays below $2$ for $\Thetatil(1/(\eta H^3))$ episodes.

Because the likelihood ratio only stays below $2$ for $\Thetatil(1/(\eta H^3))$ episodes, GRR can only handle the pairs $(s,a)$ with $q^{\pi_k}(s,a)\gtrsim\eta H^3$. Fortunately, the other pairs, rarely visited by $\pi_k$, only cause small overheads even if their bonuses are large, as the bonuses are weighted by $q^{\pi_k}(s)$ in \Cref{lem:bonus}. We thus only need GRR estimates for the pairs $(s,a)$ with $q^{\pi_k}(s,a)\ge\eta H^3$, up to a logarithmic factor, and set $\Qhat_k(s,a)=0$ for the others. This underestimates $Q_k^{\pi_k}(s,a)\in[0,H]$ by at most $H$, so we construct the following \textit{thresholding bonus}:
\begin{equation}\label{eq:bonus informal}
  b_k(s)=\sum_{a\in \cA} \pi_k(a\mid s)\left (\frac{\eta H^3}{q^{\pi_k}(s,a)} \ind[q^{\pi_k}(s,a)\ge\eta H^3]+H\ind[q^{\pi_k}(s,a)<\eta H^3]\right ),
\end{equation}
where the first term is a standard bonus, explained below, and the second term handles the rarely visited state-action pairs. After summing $b_k(s)$ over $s\in \cS$ (weighted by $q^{\pi_k}(s)$ using \Cref{lem:bonus}), all second terms together only contribute an $\Otil(\eta H^4 SA K)$ regret overhead.

Of course, since $q^{\pi_k}$ depends on the unknown transition $\PP$, the agent cannot directly use \Cref{eq:bonus informal}. We handle this with the confidence set of transitions from previous works \citep{jin2020learning,luo2021policy}; however, unlike in these works, the confidence set does not enter $\Qhat_k(s,a)$, but only decides \emph{which} pairs to estimate. The \textit{bonus Q-function} $B_k$ in \Cref{eq:bonus-q} also depends on $\PP$. As in these works, we compute it by dynamic programming under the most optimistic transition in the confidence set, which satisfies \Cref{eq:bonus-q} with ``$\ge$'' and thus suffices for \Cref{lem:bonus} (see \Cref{sec:tabular} for more details).

\paragraph{Updating as If the Feedback Were Delayed.}
As $\Qhat_k(s,a)$ is received piece by piece over $\what_k(s,a)$ episodes, exponential weights faces delayed feedback, where the loss of an episode is only revealed several episodes later. Borrowing ideas from the delayed AMDP literature \citep{jin2022near,lancewicki2023delay}, we split the exponential weights regret at a state $s\in \cS$ into two terms:
\begin{equation}\label{eq:delay-split}
  \begin{aligned}
  \sum_{\kappa=1}^K\sum_{a\in \cA}\bigl(\pi_\kappa-\pi^\star\bigr)(a\mid s)\,g_\kappa(s,a)+\sum_{k=1}^K\sum_{a\in \cA}\frac{L_{k,h}}2\sum_{n=1}^{\what_k(s,a)}\bigl(\pi_k-\pi_{k+n}\bigr)(a\mid s).
  \end{aligned}
\end{equation}
The first term is the regret of exponential weights on the received losses. Using the standard analysis of exponential weights and plugging in the second moment of $\Qhat_k(s,a)$ in \Cref{eq:grr}, we have
\begin{equation}\label{eq:exponential weights regret in tabular}
  \begin{aligned}
  \E\Biggl[\sum_{\kappa=1}^K\sum_{a\in \cA}\bigl(\pi_\kappa-\pi^\star\bigr)(a\mid s)\,g_\kappa(s,a)\Biggr]&\le\frac{\log A}\eta+\Otil(\eta H^2)\sum_{k=1}^K\E\Biggl[\sum_{a\in \cA}\frac{\pi_k(a\mid s)}{q^{\pi_k}(s,a)}\Biggr],
  \end{aligned}
\end{equation}
where the first term is the penalty term and the second is the stability term.

The second term in \Cref{eq:delay-split} is the \textit{delay cost}, as the $n$-th piece of $\Qhat_k(s,a)$ is used by $\pi_{k+n}$ instead of $\pi_k$. Since every episode changes $\log\pi(a\mid s)$ by $\Otil(\eta H^2)$, $\pi_k(a\mid s)-\pi_{k+n}(a\mid s)\le\Otil(\eta H^2n)\pi_k(a\mid s)$. Therefore, the total delay cost of $\Qhat_k(s,a)$ is at most $\Otil(\eta H^3)\pi_k(a\mid s)\what_k(s,a)^2$. Because episode $k$ visits $(s,a)$ with probability $q^{\pi_k}(s,a)$, after which $\what_k(s,a)$ has second moment at most $8/q^{\pi_k}(s,a)^2$, the expected delay cost is at most $\Otil(\eta H^3)\pi_k(a\mid s)/q^{\pi_k}(s,a)$ -- the same order in $q^{\pi_k}(s,a)$ as the stability term, and only an $H$ factor larger.

Therefore, for the pairs $(s,a)$ with a GRR estimate, the first term in the bonus $b_k(s)$ (defined in \Cref{eq:bonus informal}) cancels both the stability term and the delay cost. Following the standard analysis of policy optimization (\Cref{lem:bonus}), the bonus then only leaves an overhead of $\sum_{k=1}^K\sum_{s\in\cS}q^{\pi_k}(s)b_k(s)$. As $q^{\pi_k}(s,a)=q^{\pi_k}(s)\pi_k(a\mid s)$ cancels the denominator in \Cref{eq:bonus informal}, this overhead is only $\Otil(\eta H^4SAK)$. Since the bonuses no longer involve the width of the confidence set, we avoid the extra $\sqrt S$ factor of \citet{luo2021policy}. This gives the following result, formally stated in \Cref{sec:tabular}.

\begin{theorem}[Informal Version of \Cref{thm:tabular}]\label{thm:informal-tabular}
  In tabular AMDPs, policy optimization with GRR and a threshold of order $\Thetatil(\eta H^3)$ attains $\mR_K=\Otil(H^{5/2}\sqrt{SAK})$ with high probability.
\end{theorem}

\subsection{Introduced Feedback Delays in Linear AMDPs: Matrix GRR and Regions}\label{sec:tech-linear}

In linear AMDPs, formally defined in \Cref{sec:setup-linear}, the state space can be infinite, but all transitions, losses, and Q-functions are linear in some known $d$-dimensional features $\phi\colon \cS\times \cA\to \R^d$. Specifically, recall the covariance matrix $\Sigma_h^\pi=\E_\pi[\phi_h\phi_h^\top]$ of a policy $\pi$ at layer $h$ in \Cref{eq:covariance}, where $\phi_h$ is the layer-$h$ feature of a trajectory of $\pi$. As a linear analogue of the importance-weighted estimate in \Cref{eq:iw}, it turns out that the Q-function $Q_k^{\pi_k}(s,a)$ can be estimated by $\langle \phi(s,a),(\Sigma_h^{\pi_k})^\dagger\phi(s_{k,h},a_{k,h})L_{k,h}\rangle$ \citep{neu2021online}. It remains to estimate the inverse covariance $(\Sigma_h^{\pi_k})^\dagger$.

\paragraph{Estimating Inverse Covariances via MGRR.}
Similar to the occupancy measure $q^{\pi_k}$, the covariance $\Sigma_h^{\pi_k}$ also depends on the unknown $\PP$. This is typically handled by freezing the policy within epochs \citep{liu2024towards}. As longer epochs reduce the estimation error but slow down the policy updates, balancing the two gives their $\Otil(K^{3/4})$ regret. Instead, as with GRR, we estimate $(\Sigma_h^{\pi_k})^\dagger$ with future episodes using \textbf{M}atrix \textbf{G}eometric \textbf{R}esampling \citep{neu2020efficient} with \textbf{R}ejection (MGRR).

Instead of GRR initiated for state-action pairs, MGRR is initiated for layers. An MGRR estimate for episode $k$ and layer $h$ makes an attempt in every later episode $\kappa>k$. If episode $\kappa$ is accepted (with probability $\alpha_{k,\kappa}$ in \Cref{eq:rejection accept probability}), we take its layer-$h$ feature $\phi_{\kappa,h}\triangleq\phi(s_{\kappa,h},a_{\kappa,h})$. Using \Cref{lem:rejection}, we have $\E_{\kappa-1}[\ind[\kappa\text{ accepted}]\phi_{\kappa,h}\phi_{\kappa,h}^\top]=\Sigma_h^{\pi_k}/2$. Using the identity $\Sigma^\dagger=\frac12\sum_{n=1}^\infty (I-\Sigma/2)^{n-1}$ for $0\prec\Sigma\preceq I$, we can therefore estimate the inverse covariance matrix $(\Sigma_h^{\pi_k})^\dagger$ as an infinite series:
\begin{equation}\label{eq:MGRR raw}
  (\Sigma_h^{\pi_k})^\dagger\approx \frac12\sum_{n\ge 1}\prod_{\kappa=k+1}^{k+n-1} \left (I-\ind[\kappa\text{ accepted}]\phi_{\kappa,h}\phi_{\kappa,h}^\top\right ).
\end{equation}

But there are two issues. First, MGRR never ``succeeds.'' In GRR, we essentially wrote $1/q^{\pi_k}(s,a)=\frac12\sum_{n=1}^\infty (1-q^{\pi_k}(s,a)/2)^{n-1}\approx \frac 12\sum_{n=1}^\infty \prod_{\kappa=k+1}^{k+n-1}(1-\ind[\kappa\text{ accepted}]\ind[\pi_\kappa\text{ visits }(s,a)])$. Once both indicators are $1$, all subsequent terms are zero, and GRR succeeds. This property is absent from \Cref{eq:MGRR raw}, and we must instead truncate the infinite sum after some $M$ attempts. To make the truncation error $(I-\Sigma_h^{\pi_k}/2)^M$ small, $M$ must be at least $1/\lambda_{\min}(\Sigma_h^{\pi_k})$ (where $\lambda_{\min}(\cdot)$ is the smallest eigenvalue of a matrix). This can be arbitrarily large without additional assumptions.%
\footnote{For example, one may assume knowledge of a policy $\pi_0$ with $\lambda_{\min}(\Sigma_h^{\pi_0})\ge\lambda_0$ for all $h$ and a constant $\lambda_0>0$. This is known as an exploratory assumption, which we avoid since even the existence of $\pi_0$ is unclear (unlike in linear bandits, where such a $\pi_0$ not only exists but can also be found efficiently via John's ellipsoid \citep{john1948extremum}).\label{footnote:exploratory}} Second, an MGRR estimate changes the received losses of \emph{all} pairs in $\cS_h\times\cA$, as their Q-functions all rely on $(\Sigma_h^{\pi_k})^\dagger$. If each estimate takes so long to finish, the received losses become much larger, and consequently policies no longer stay close for a long time (which is required by \Cref{lem:rejection}).

\paragraph{Thresholding on Feature Regions.}
We fix both issues again using thresholding. In the tabular case, which can be viewed as having one-hot features, a rarely visited state-action pair corresponds to a small eigenvalue of $\Sigma_h^{\pi_k}$ (namely, $q^{\pi_k}(s,a)$). In linear AMDPs, a small eigenvalue is a rarely visited direction. We thus group the features into regions, within which no direction is rarely visited.

This idea is formalized in \Cref{alg:regions}. For each episode $k$ and layer $h$, it constructs regions $\cC_1,\cC_2,\ldots\subseteq\R^d$, one at a time, from the layer-$h$ features of the last $\sqrt K$ episodes. To construct $\cC_j$, it starts from the $m_j$ features that are not in any region yet, and repeatedly removes a feature $\phi$ with $\lVert\phi\rVert_{\Lambda_j^{-1}}^2>2d\log(1+1/\lambda)=\Thetatil(d)$, where $\Lambda_j$ is $\lambda I$ (for a tiny $\lambda>0$) plus the sum of $\phi\phi^\top/m_j$ over the remaining features. Similarly to the elliptical potential lemma \citep[Lemma~11]{abbasi2011improved}, at least $m_j/2$ features remain, and they induce an ellipsoid region $\cC_j\triangleq\{\phi:\lVert\phi\rVert_{\Lambda_j^{-1}}^2\le2d\log(1+1/\lambda)\}$. Hence, there are $\O(\log K)$ regions. Similarly to the pairs with $q^{\pi_k}(s,a)<\eta H^3$ in \Cref{eq:bonus informal}, we stop once only a small fraction of the $\sqrt K$ features is left and leave them to the bonuses, so for every region $\cC_j$, the fraction $\phat_j$ of the $\sqrt K$ features that are contained in $\cC_j$ is at least half of this small fraction.

For every region $\cC_j$, let $\Sigmahat_j$ be the empirical covariance matrix of the features in $\cC_j$. For a policy $\pi$, let $\Sigma_{h,j}^\pi\triangleq\E_\pi\bigl[\ind[\phi_h\in\cC_j]\phi_h\phi_h^\top\bigr]$ be its covariance matrix in $\cC_j$, where $\phi_h$ is the layer-$h$ feature of a trajectory of $\pi$. We now run MGRR separately in every region: whenever $\pi_k$ visits a feature $\phi_{k,h}=\phi(s_{k,h},a_{k,h})$ that lies in a region $\cC_j$, we start an MGRR estimate that \textit{i)} accepts every later episode $\kappa>k$ with probability $\alpha_{k,\kappa}$ (\Cref{eq:rejection accept probability}); \textit{ii)} feeds the zero vector instead of $\phi_{\kappa,h}$ into \Cref{eq:MGRR raw} if $\phi_{\kappa,h}\notin\cC_j$, just as a GRR attempt has no effect if it fails to visit $(s,a)$; and \textit{iii)} rescales the features in $\cC_j$ as $\phi\mapsto\Sigmahat_j^{-1/2}\phi$, which we explain below. Formally, the MGRR in each region $\cC_j$ estimates
\begin{equation}\label{eq:MGRR}
  (\Sigmahat_j^{-1/2}\Sigma_{h,j}^{\pi_k}\Sigmahat_j^{-1/2})^\dagger\approx \frac12\sum_{n=1}^{M_j}\prod_{\kappa=k+1}^{k+n-1}\left(I-\ind[\kappa\text{ accepted}]\ind[\phi_{\kappa,h}\in\cC_j]\bigl(\Sigmahat_j^{-1/2}\phi_{\kappa,h}\bigr)\bigl(\Sigmahat_j^{-1/2}\phi_{\kappa,h}\bigr)^\top\right),
\end{equation}
where $M_j$ is a truncation level that cuts off the infinite series. Unlike a GRR estimate, this estimate gives a piece to every $(s,a)$ whose $\phi(s,a)\in \cC_j$, as their Q-functions are linear in the same vector.

We claim this thresholding resolves both issues. Indeed, as long as the policies of the last $\sqrt K$ episodes stay close to $\pi_k$, a standard uniform convergence bound \citep[Theorem~5.1]{boucheron2005theory} gives
\begin{equation}\label{eq:uniform convergence bound}
  \frac{\phat_j}{4} \Sigmahat_j\preceq \Sigma_{h,j}^{\pi_k}\preceq 4\phat_j \Sigmahat_j,\;\text{w.h.p. }\forall j.
\end{equation}
Hence, after the rescaling by $\Sigmahat_j^{-1/2}$, the matrix to invert has a $\lambda_{\min}$ at least $\phat_j/4$, and thus setting $M_j=\Thetatil(d/\phat_j)\le\sqrt K$ suffices to control the truncation error. Furthermore, as an estimate in $\cC_j$ is started with probability about $\phat_j$ and lasts $M_j$ episodes, every region has $\Thetatil(d)$ unfinished estimates at a time, just as every state-action pair only has $\Otil(1)$ in tabular AMDPs. Following \Cref{sec:tech-tabular}, the policies thus stay close for $\Thetatil(1/(\eta dH^3))$ episodes, which is at least $\sqrt K$ for $\eta=\Thetatil(1/(dH^3\sqrt K))$.

Our bonus function for linear AMDPs is similar to \Cref{eq:bonus informal}, but with $q^{\pi_k}(s,a)$ replaced by the probability of visiting the region that contains $\phi(s,a)$ (i.e., the $\phat$ above). The second term still handles those rarely visited state-action pairs (i.e., those whose features lie outside all the regions). Unlike in tabular AMDPs, we do not keep a confidence set of transitions, so we cannot compute the bonus Q-function $B_k$ (recall \Cref{eq:bonus-q}) using optimism. Instead, we estimate $Q_k^{\pi_k}-B_k$ directly. As $\PP$ is linear in $\phi(s,a)$, the expected future bonus $B_k(s,a)-b_k(s)=\sum_{s'}\PP(s'\mid s,a)\sum_{a'}\pi_k(a'\mid s')B_k(s',a')$ is also linear in $\phi(s,a)$, just like $Q_k^{\pi_k}$. Hence, if we combine the \textit{cumulative loss-minus-bonus} function
\begin{equation*}
  Y_{k,h}\triangleq\sum_{h'=h}^{H-1}\ell_k(s_{k,h'},a_{k,h'})-\sum_{h'=h+1}^{H-1}b_k(s_{k,h'})
\end{equation*}
with the MGRR estimate derived in \Cref{eq:MGRR}, we can directly estimate $Q_k^{\pi_k}-B_k$ at every $(s,a)$ with $\phi(s,a),\phi_{k,h}\in\cC_j$ as (we defer the corner case where some $(s,a)$'s do not have estimates to \Cref{sec:linear})
\begin{equation*}
  Q_k^{\pi_k}(s,a)-B_k(s,a)\approx \left \langle \Sigmahat_j^{-1/2}\phi(s,a), (\Sigmahat_j^{-1/2}\Sigma_{h,j}^{\pi_k}\Sigmahat_j^{-1/2})^\dagger \Sigmahat_j^{-1/2}\phi_{k,h} Y_{k,h}\right \rangle -b_k(s).
\end{equation*}

\paragraph{High-Probability Bounds.}
Our algorithm also allows high-probability bounds, whereas all existing efficient algorithms for linear AMDPs only bound the expected regret \citep{luo2021policy,dai2023refined,sherman2023improved,liu2024towards}.%
\footnote{The only high-probability bound for linear AMDPs \citep{kong2024improved} runs exponential weights over $K^{\O(dH^2)}$ policies (and is thus computationally inefficient), requires an exploratory assumption (see \Cref{footnote:exploratory}), and only attains $K^{4/5}$ regret. Apart from this one, we exclude from this discussion prior works that use exploratory assumptions or simulators (see \Cref{sec:related}): although most such regret bounds are stated in expectation, we believe some of them can be made high-probability.} Using the bias correction idea of linear bandits \citep{bartlett2008high,zimmert2022return}, the exponential moment of an estimate is bounded if \textit{i)} its magnitude is at most $1/\eta$, and \textit{ii)} given the history, it does not overestimate its target in expectation. Previous estimators, essentially MGRR without regions and without rejections, fail to meet condition \textit{i)}. They all suffer from the aforementioned first issue: in rarely visited directions, the estimated inverse covariance matrix is large. Hence, previous works add a regularization $\lambda I$ before taking the inverse (i.e., estimate $(\Sigma_h^{\pi_k}+\lambda I)^\dagger$ instead), which introduces a bias. Satisfying condition \textit{i)} thus requires $\eta\lesssim\lambda/H$, which rules out a small $\lambda$ and causes a large bias \citep{luo2021policy}. Subsequent works develop various techniques that allow $\lambda\ll \eta H$ to reduce the bias, and therefore fail to satisfy condition \textit{i)} \citep{dai2023refined,liu2024towards}. In contrast, the regions and the rejections make our estimates satisfy both conditions, which gives a high-probability bound (see \Cref{sec:linear}).

Altogether, omitting $d$, $H$, and logarithmic factors, the regret is $\O(1/\eta+K/\sqrt K)=\O(\sqrt K)$, where $1/\eta$ is the exponential weights regret as in \Cref{eq:exponential weights regret in tabular}, and $K/\sqrt K$ comes from the features outside all regions. This gives the following result, formally stated in \Cref{sec:linear}.

\begin{theorem}[Informal Version of \Cref{thm:linear}]\label{thm:informal-linear}
  In linear AMDPs, policy optimization with MGRR on feature regions attains $\mR_K=\Otil(d^3H^4\sqrt K)$ with high probability.
\end{theorem}

\subsection{Extending to Actual Feedback Delays}\label{sec:tech-delay}

In delayed AMDPs, formally defined in \Cref{sec:setup-delayed}, the agent still observes her trajectory during every episode, but the losses of episode $k$ are only revealed at the end of episode $k+\fd_k$ (or never, if $k+\fd_k>K$). Here, the delays $\fd_k\ge0$ are arbitrarily chosen by an oblivious adversary in advance. Recall that $\mD=\sum_{k=1}^K\fd_k$ is the total delay, and taking $\mD=0$ recovers our non-delayed models.

Our analysis in \Cref{eq:delay-split} already accommodates feedback delays, so the only required change is to postpone the start of a (GRR or MGRR) estimate until its feedback arrives, as both GRR in \Cref{eq:grr} and MGRR in \Cref{eq:MGRR raw} rely on the cumulative losses $L_{k,h}$. Following the skipping technique \citep{thune2019nonstochastic}, we fix a cap $\dbar$ on the delays and \textit{process} every episode $k$ at the end of episode $k+\min\{\fd_k,\dbar\}$. If $\fd_k\le\dbar$ and $k+\fd_k\le K$, the losses of episode $k$ have arrived by then, so we start its estimates as usual. Otherwise, we call episode $k$ \textit{skipped} and start no estimate for it. In both cases, the trajectory of episode $k$ makes its attempts only when episode $k$ is processed, which keeps \Cref{lem:rejection} valid (see \Cref{sec:delay} for details). With $\dbar\approx\sqrt\mD/H$ in tabular AMDPs and $\dbar\approx\sqrt{\mD/(dH^2)}$ in linear AMDPs, the delays only add $\Otil(H^3\sqrt\mD)$ and $\Otil(d^{3/2}H^3\sqrt\mD)$ to the regret, respectively.

\begin{theorem}[Informal Version of \Cref{thm:delayed-tabular,thm:delayed-linear}]\label{thm:informal-delay}
  In delayed AMDPs with total delay $\mD$, the algorithms above attain $\mR_K=\Otil(H^{5/2}\sqrt{SAK}+H^3\sqrt\mD)$ in tabular AMDPs and $\mR_K=\Otil(d^3H^4\sqrt K+d^{3/2}H^3\sqrt\mD)$ in linear AMDPs, both with high probability.
\end{theorem}

\section{Attaining \texorpdfstring{$\Otil(H^{5/2}\sqrt{SAK})$ Regret for Tabular AMDPs}{Regret for Tabular AMDPs}}\label{sec:tabular}

We now formalize the ideas of \Cref{sec:tech-tabular} into \Cref{alg:tabular} and present its regret guarantee. To handle the unknown transition, we mostly follow existing ideas \citep{jin2020learning,luo2021policy}. First, for each episode $k$, we maintain a confidence set $\cP_k$ of transitions via empirical Bernstein bounds \citep{maurer2009empirical}. Specifically, let $N_k(s,a)$ be the number of visits to $(s,a)$ before episode $k$, let $\Phat_n(\cdot\mid s,a)$ be the empirical distribution of the first $n$ next states observed after visiting $(s,a)$, and let $\beta\triangleq\log(16S^2AK/\delta)$. Then
\begin{equation}\label{eq:conf-set}
  \cP_k\triangleq\left\{P'\ \middle|\
  \begin{aligned}
  &\bigl\lvert P'(s'\mid s,a)-\Phat_n(s'\mid s,a)\bigr\rvert\le2\sqrt{\Phat_n(s'\mid s,a)\beta/n}+5\beta/n,\\
  &\forall\,0\le h<H,\ (s,a,s')\in\cS_h\times\cA\times\cS_{h+1},\ 1\le n\le N_k(s,a)
  \end{aligned}
  \right\},
\end{equation}
where, if no distribution in $\Delta(\cS_{h+1})$ satisfies all constraints of a pair $(s,a)$, we drop these constraints, so that $P'(\cdot\mid s,a)$ can be any distribution and $\cP_k$ is never empty. These confidence sets ensure that $\PP\in\cP_k$ for all $k\in[K]$ with probability at least $1-\delta/4$ (\Cref{lem:good-event}), so the lower confidence bound $\qlow_k(s,a)\triangleq\min_{P'\in\cP_k}q^{P',\pi_k}(s,a)$ satisfies $\qlow_k\le q^{\pi_k}$.

Although our bonus in \Cref{sec:tech-tabular} no longer has (the confidence width of) $q^{\pi_k}$ in the numerator due to our GRR technique, its vanilla form thresholds on $q^{\pi_k}$ to distinguish those rarely visited state-action pairs. We then replace it by the lower confidence bound $\qlow_k$, which gives $b_k$ in \Cref{eq:tabular-bonus} of \Cref{alg:tabular}. Note that this does not introduce any dependency on the widths of the confidence sets.

Similarly, the bonus Q-function $B_k$ in \Cref{eq:tabular-B} is calculated using optimism (i.e., maximizing over $\cP_k$) \citep{luo2021policy,lancewicki2023delay}, and we denote the maximizing transition by $\Ptil_k\in\cP_k$. When $\PP\in\cP_k$, $B_k$ thus satisfies \Cref{eq:bonus-q} with ``$\ge$'', and \Cref{lem:bonus} turns the bonus overhead into $\sum_{s\in\cS}q^{\Ptil_k,\pi_k}(s)b_k(s)$ per episode, which we later bound via the widths of $\cP_k$. Unlike \citet{luo2021policy}, we do not need to ``dilate'' our bonus $b_k$ in $B_k$, since it is already small: $0\le b_k\le H$ gives $0\le B_k\le H^2$, so $-B_k$ only adds $\O(\eta H^5)$ per episode to the stability term \citep[Lemma~C.6]{lancewicki2023delay}.

With these bonuses, \Cref{alg:tabular} runs the GRR in \Cref{sec:tech-tabular}, but with three slight modifications: \textit{i)} if an attempt does not succeed, the estimate is \textit{randomly stopped} with probability $\gamma$, which slightly underestimates $Q_k^{\pi_k}$ but allows a high-probability bound, similar to the implicit exploration technique \citep{neu2015explore}; \textit{ii)} an estimate is forced to finish after $M=\Thetatil(1/(\eta H^3))$ attempts, so that \Cref{lem:rejection} always applies; and \textit{iii)} if a state-action pair already has $\nu=\Otil(1)$ unfinished estimates, we no longer initiate new ones, so that every episode adds at most $\nu H/2$ to every received loss.

For each episode $\kappa$, the received loss $g_\kappa(s,a)$ is then $-B_\kappa(s,a)$ plus $L_{k,h}/2$ for every ongoing GRR estimate for $(s,a)$ (i.e., a piece). If $\what_k(s,a)$ denotes the final number of attempts of the estimate started for $(s,a)$ in episode $k$, these pieces add up to $\Qhat_k(s,a)=\ind[(s_{k,h},a_{k,h})=(s,a)]\,L_{k,h}\what_k(s,a)/2$, exactly as in \Cref{eq:grr}. The full procedure is given in \Cref{alg:tabular}.

\begin{algorithm}[t]
\caption{Policy Optimization with GRR for Tabular AMDPs}
\label{alg:tabular}
\begin{algorithmic}[1]
\Require Confidence parameter $\delta$; parameters $\eta,\gamma,\theta,M,\nu,c_1$ (specified in \Cref{thm:tabular}).
\For{$\kappa=1,2,\ldots,K$}
  \State For every state $s$, let $\pi_\kappa(a\mid s)\propto\exp\bigl(-\eta\sum_{\kappa'<\kappa}g_{\kappa'}(s,a)\bigr)$. \label{line:tab-policy}
  \State Initialize the received loss $g_\kappa\gets0$.
  \State Execute $\pi_\kappa$, and observe the trajectory $(s_{\kappa,h},a_{\kappa,h})_{h=0}^{H-1}$ and losses $(\ell_\kappa(s_{\kappa,h},a_{\kappa,h}))_{h=0}^{H-1}$.
  \State Compute $\cP_\kappa$ via \Cref{eq:conf-set} with the transitions of the episodes before $\kappa$.
  \State Compute $\qlow_\kappa(s,a)=\min_{P'\in\cP_\kappa}q^{P',\pi_\kappa}(s,a)$ and the bonus
  \algeq[eq:tabular-bonus]{$\displaystyle b_\kappa(s)\triangleq\sum_{a\in\cA}\pi_\kappa(a\mid s)\left(\frac{c_1\eta H^3}{\qlow_\kappa(s,a)}\ind[\qlow_\kappa(s,a)\ge\theta]+H\,\ind[\qlow_\kappa(s,a)<\theta]\right).$}
  \State Compute the bonus Q-function backward from the last layer, with $B_\kappa=0$ on $\cS_H$:
  \algeq[eq:tabular-B]{$\displaystyle B_\kappa(s,a)\triangleq b_\kappa(s)+\max_{P'\in\cP_\kappa}\sum_{s'\in\cS_{h+1}}P'(s'\mid s,a)\sum_{a'}\pi_\kappa(a'\mid s')B_\kappa(s',a'),\;\forall s\in\cS_h,a\in\cA.$}
  \State Update $g_\kappa\gets g_\kappa-B_\kappa$.
  \For{each unfinished estimate, started in an episode $k$ before $\kappa$ for a pair $(s,a)\in\cS_h\times\cA$}
    \State Set $\what_k(s,a)\gets\what_k(s,a)+1$ and $g_\kappa(s,a)\gets g_\kappa(s,a)+L_{k,h}/2$.
    \State Accept the attempt with probability $\min\{1,\alpha_{k,\kappa}\}$, where $\alpha_{k,\kappa}$ is defined in \Cref{eq:rejection accept probability}. \label{line:tab-accept}
    \State If the attempt is accepted and $(s_{\kappa,h},a_{\kappa,h})=(s,a)$, the estimate succeeds.
    \State If the estimate does not succeed, it is randomly stopped with probability $\gamma$.
    \State The estimate finishes if it succeeds, is randomly stopped, or has $\what_k(s,a)=M$.
  \EndFor
  \For{each $h$ such that $\qlow_\kappa(s_{\kappa,h},a_{\kappa,h})\ge\theta$}
  \If{$(s_{\kappa,h},a_{\kappa,h})$ had fewer than $\nu$ unfinished estimates before the attempts with episode $\kappa$}
    \State Start a GRR estimate for $(s_{\kappa,h},a_{\kappa,h})$ and initialize $\what_\kappa(s_{\kappa,h},a_{\kappa,h})\gets0$.
  \EndIf
  \EndFor
\EndFor
\end{algorithmic}
\end{algorithm}

Formally, we have the following regret guarantee:

\begin{theorem}\label{thm:tabular}
  For tabular AMDPs and any $\delta\in(0,1)$, pick $\nu=\Theta(\log^2(SAHK/\delta))$, $c_1=\Theta(\nu)$, $\gamma=\eta H^2$, $\theta=\Thetatil(\eta H^3)$, $M=\Thetatil(1/\theta)$, and $\eta=\Thetatil(\min\{1/H^3,1/(H^{3/2}\sqrt{SAK})\})$, with the constants in \Cref{eq:tab-params}. Then \Cref{alg:tabular} ensures the following with probability at least $1-\delta$:
  \begin{equation*}
    \mR_K=\Otil\bigl(H^{5/2}\sqrt{SAK}+H^4S^2A\bigr).
  \end{equation*}
\end{theorem}

Compared with the $\Otil(H^2S\sqrt{AK})$ bound of \citet{luo2021policy}, our bound is smaller by a factor of $\sqrt{S/H}$, and the estimation error of the transition only enters the $K$-independent term $\Otil(H^4S^2A)$. Moreover, \Cref{alg:tabular} is computationally efficient, as $\qlow_k$ and $B_k$ can be computed by dynamic programming \citep{jin2020learning,luo2021policy}, and every episode makes at most $\nu SA$ attempts.

\begin{proof}[Proof Sketch of \Cref{thm:tabular}]
  To bound $\mR_K$ by \Cref{lem:bonus}, we focus on a single state $s$ and bound $\sum_{k}\sum_a(\pi_k(a\mid s)-\pi^\star(a\mid s))(Q_k^{\pi_k}-B_k)(s,a)$. As exponential weights only controls the regret on the estimates, we add and subtract $\Qhat_k(s,a)$. Hence, after summing over $s$ with weights $q^\star(s)$, we get
  \begin{equation}\label{eq:decomp}
    \mR_K=\BiasOne+\BiasTwo+\RegTerm+\Bonus,
  \end{equation}
  where the four terms are defined as follows:
  \begin{itemize}
    \item $\BiasOne=\sum_{k,s}q^\star(s)\sum_a\pi_k(a\mid s)(Q_k^{\pi_k}-\Qhat_k)(s,a)$, measuring the underestimation of $\Qhat_k$ under $\pi_k$;
    \item $\BiasTwo=\sum_{k,s}q^\star(s)\sum_a\pi^\star(a\mid s)(\Qhat_k-Q_k^{\pi_k})(s,a)$, measuring the overestimation of $\Qhat_k$ under $\pi^\star$;
    \item $\RegTerm=\sum_{k,s}q^\star(s)\sum_a(\pi_k(a\mid s)-\pi^\star(a\mid s))(\Qhat_k-B_k)(s,a)$, the regret of exponential weights on the losses $\Qhat_k-B_k$;
    \item $\Bonus=\sum_{k,s}q^\star(s)\sum_a(\pi_k(a\mid s)-\pi^\star(a\mid s))B_k(s,a)$, the bonus term. If every $B_k$ satisfies \Cref{eq:bonus-q} with ``$\ge$'', then \Cref{lem:bonus} gives $\Bonus\le\sum_k\sum_a\pi_k(a\mid s_0)B_k(s_0,a)-\sum_{k,s}q^\star(s)b_k(s)$, where the first sum is the bonus collected under $\pi_k$, and the second sum cancels part of $\BiasOne$ and $\RegTerm$.
  \end{itemize}
  Our choices of $\theta$ and $M$ keep the policies close during the at most $M$ attempts of every estimate (\Cref{lem:stability}), so by \Cref{lem:rejection}, every attempt succeeds or randomly stops its estimate with probability exactly $\gamma+(1-\gamma)q^{\pi_k}(s,a)/2$, regardless of the history (\Cref{lem:accept}). Therefore, with high probability, neither the $M$ cap nor the $\nu$ cap binds (\Cref{lem:unfinished}), so the estimates behave as in \Cref{sec:tech-tabular} except for the random stops. Hence, $\BiasTwo=\Otil(\nu/\eta)$ by concentration (\Cref{prop:bias2}). Our $c_1$ allows the bonus to cover the stability term, the delay cost, and the bias of the random stops. Hence, using \Cref{prop:bias1,prop:reg},
  \begin{equation*}
    \BiasOne+\RegTerm\le\sum_{k=1}^K\sum_{s\in\cS}q^\star(s)b_k(s)+\Otil\Bigl(\frac H\eta\Bigr)+2\eta H^5K,
  \end{equation*}
  where the $2\eta H^5K$ term comes from the magnitude of $-B_k$. As $\PP\in\cP_k$, $B_k$ satisfies \Cref{eq:bonus-q} with ``$\ge$'', so the bound on $\Bonus$ above cancels the first sum in this display. It leaves $\sum_k\sum_a\pi_k(a\mid s_0)B_k(s_0,a)=\sum_{k,s}q^{\Ptil_k,\pi_k}(s)b_k(s)$, as $B_k$ is the Q-function of $\pi_k$ for the loss $b_k$ under $\Ptil_k$.

  It remains to show that the width of $\cP_k$ only adds a $K$-independent term. If a transition has been observed $\Omegatil(H^2)$ times (\Cref{eq:large}), every $P'\in\cP_k$ is within a factor $1\pm\frac1H$ of $\PP$ on it. Hence, along such transitions, $\Ptil_k$ visits every pair at most $e^2$ times as often as $\qlow_k$ predicts. Every other transition has probability $\O(H^2\beta/\max\{1,N_k(s,a)\})$ under every $P'\in\cP_k$. Hence, by \Cref{lem:comparison},
  \begin{align*}
    \sum_{s\in\cS}q^{\Ptil_k,\pi_k}(s)b_k(s)&\lesssim\sum_{s,a}\qlow_k(s,a)\Bigl(\frac{c_1\eta H^3}{\qlow_k(s,a)}\ind[\qlow_k(s,a)\ge\theta]+H\ind[\qlow_k(s,a)<\theta]\Bigr)\\
    &\quad+H^4S\beta\sum_{s,a}\frac{q^{\pi_k}(s,a)}{\max\{1,N_k(s,a)\}},
  \end{align*}
  where the first sum is $\Otil(\eta H^4SA)$ and the second sums to $\Otil(H^4S^2A)$ over all episodes (\Cref{lem:harmonic}). This gives $\mR_K=\Otil(H/\eta+\eta H^4SAK)+\Otil(H^4S^2A)$. Plugging in our $\eta$ finishes the proof.
\end{proof}

\section{Attaining \texorpdfstring{$\Otil(d^3H^4\sqrt K)$ Regret for Linear AMDPs}{Regret for Linear AMDPs}}\label{sec:linear}

We now formalize the ideas of \Cref{sec:tech-linear} into \Cref{alg:linear}, which attains $\Otil(d^3H^4\sqrt K)$ regret in linear AMDPs (\Cref{def:linear}). As in \Cref{alg:tabular}, it runs exponential weights on estimates that arrive piece by piece, and adds a thresholding bonus for the pairs that are not estimated. The only two differences are that \textit{i)} it thresholds on feature regions instead of $\qlow_k$, and \textit{ii)} it estimates $Q_k^{\pi_k}-B_k$ directly instead of computing $B_k$. For simplicity, we assume that $\sqrt K$ is an integer.

\paragraph{Regions and Bonuses.}
In every episode $k$ and layer $h$, \Cref{alg:regions} constructs the regions from the features $\phi_{\kappa,h}\triangleq\phi(s_{\kappa,h},a_{\kappa,h})$ of the last $\sqrt K$ episodes, as in \Cref{sec:tech-linear}, and stops once at most $\sqrt K\theta$ features are left. As in the proof of the elliptical potential lemma \citep[Lemma~11]{abbasi2011improved}, every region keeps more than half of the $m_j$ features it starts with. Hence, there are at most $\lceil\log_2(2\sqrt K)\rceil$ regions, each with \textit{empirical probability} $\phat_j>\theta/2$. \Cref{alg:regions} also returns the empirical covariance matrix $\Sigmahat_j$ of the features in $\cC_j$, with its tiny eigenvalues truncated to zero. Every $\phi\in\cC_j$ satisfies $\lVert\phi\rVert_{\Sigmahat_j^\dagger}^2\le4d\log(1+1/\lambda)$ (\Cref{lem:lin-regions-det}), and we write $\Sigma_{h,j}^\pi$ for the covariance matrix of a policy $\pi$ in $\cC_j$ (defined in \Cref{sec:tech-linear}).

\begin{algorithm}[t]
\caption{Region Construction}
\label{alg:regions}
\begin{algorithmic}[1]
\Require Features $\phi_1,\ldots,\phi_{\sqrt K}\in\R^d$; regularization $\lambda>0$; probability threshold $\theta$.
\State Let $R\gets\{1,\ldots,\sqrt K\}$ index the features not in any region yet, and $j\gets0$.
\While{$\lvert R\rvert>\sqrt K\theta$}
  \State Let $j\gets j+1$, $m_j\gets\lvert R\rvert$, $R_j\gets R$, and $\Lambda_j\gets\lambda I+\frac1{m_j}\sum_{i\in R_j}\phi_i\phi_i^\top$.
  \While{some $i\in R_j$ has $\lVert\phi_i\rVert_{\Lambda_j^{-1}}^2>2d\log(1+1/\lambda)$}
    \State Remove one such $i$ from $R_j$, and recompute $\Lambda_j\gets\lambda I+\frac1{m_j}\sum_{i\in R_j}\phi_i\phi_i^\top$ with $m_j$ fixed.
  \EndWhile
  \State Let $\cC_j\gets\{\phi\in\R^d:\lVert\phi\rVert_{\Lambda_j^{-1}}^2\le2d\log(1+1/\lambda)\}\setminus(\cC_1\cup\cdots\cup\cC_{j-1})$ be the $j$-th region.
  \State Let $\phat_j\gets\lvert R_j\rvert/\sqrt K$ be its empirical probability.
  \State Let $\Sigmahat_j\gets\frac1{\lvert R_j\rvert}\sum_{i\in R_j}\phi_i\phi_i^\top$ with all eigenvalues below $\lambda m_j/\lvert R_j\rvert$ truncated to zero.
  \State Let $R\gets R\setminus R_j$.
\EndWhile
\State \Return $(\cC_j,\phat_j,\Sigmahat_j)$ for all $j$.
\end{algorithmic}
\end{algorithm}

With the regions, \Cref{alg:linear} decides which pairs to estimate as \Cref{alg:tabular} does: episode $k$ estimates a pair $(s,a)\in\cS_h\times\cA$, written $u_k(s,a)=0$, if $\phi(s,a)$ lies in a region and fewer than $\nu$ estimates at layer $h$ are unfinished; otherwise, $u_k(s,a)=1$. For an estimated pair, let $\phat_k(s,a)\triangleq\phat_j$ for the region $\cC_j$ that contains $\phi(s,a)$. The bonus $b_k$ in \Cref{eq:linear-bonus} has the form of \Cref{eq:tabular-bonus} with $\qlow_k(s,a)$ replaced by $\phat_k(s,a)$, and $b_k(s)\le4H$ as $\phat_k(s,a)>\theta/2$. Its first term covers the second moments and the delay cost, and its second term covers the unestimated pairs.

To apply \Cref{lem:bonus}, we also need a bonus Q-function $B_k$ that satisfies \Cref{eq:bonus-q} with ``$\ge$''. We first define $B_k$ using the true transition $\PP$, and then introduce how \Cref{alg:linear} estimates it. At those unestimated pairs, $B_k$ replaces the future bonus by its upper bound $4H(H-h-1)$. Formally, $B_k=0$ on $\cS_H$ and, for all $s\in\cS_h$ and $a\in\cA$,
\begin{equation}\label{eq:linear-B}
  B_k(s,a)\triangleq b_k(s)+(1-u_k(s,a))\sum_{s'}\PP(s'\mid s,a)\sum_{a'}\pi_k(a'\mid s')B_k(s',a')+4H(H-h-1)\,u_k(s,a).
\end{equation}

\paragraph{Estimates via MGRR.}
By \Cref{eq:linear-B}, on the unestimated pairs with $u_k(s,a)=1$, we already know $B_k(s,a)=b_k(s)+4H(H-h-1)$. Otherwise, \Cref{alg:linear} estimates $(Q_k^{\pi_k}-B_k)(s,a)$ following the cumulative loss-minus-bonus in \Cref{sec:tech-linear}, except for taking care of the possibility of encountering an unestimated state-action pair:
\begin{equation}\label{eq:response}
  Y_{k,h}\triangleq L_{k,h}-\sum_{h'=h+1}^{H-1}\left(\prod_{h''=h+1}^{h'-1}\bigl(1-u_k(s_{k,h''},a_{k,h''})\bigr)\right)\Bigl(b_k(s_{k,h'})+4H(H-h'-1)\,u_k(s_{k,h'},a_{k,h'})\Bigr).
\end{equation}
In words, $Y_{k,h}$ equals $L_{k,h}$ minus the bonuses up to the first later layer $h'$ whose pair $(s_{k,h'},a_{k,h'})$ is not estimated, where it also subtracts $4H(H-h'-1)$ and stops. We still have $\lvert Y_{k,h}\rvert\le H+4H^2$. As $\PP$ and $\ell_k$ are both linear, the conditional mean of $Y_{k,h}$ is also linear (\Cref{lem:lin-response}):
\begin{align}
  &\quad \E\bigl[Y_{k,h}\bigm|\cF_{k-1},(s_{k,h},a_{k,h})=(s,a)\bigr]\notag\\
  &=Q_k^{\pi_k}(s,a)-\sum_{s'}\PP(s'\mid s,a)\sum_{a'}\pi_k(a'\mid s')B_k(s',a')\nonumber\\
  &=\langle\phi(s,a),\xi_{k,h}\rangle,\label{eq:lin-response-mean}
\end{align}
where $\xi_{k,h}\in\R^d$ is an unknown vector. By \Cref{eq:linear-B}, the first line in \Cref{eq:lin-response-mean} is just $(Q_k^{\pi_k}-B_k+b_k)(s,a)$ when $u_k(s,a)=0$, so it suffices to estimate $\langle\phi(s,a),\xi_{k,h}\rangle$ and subtract the known $b_k(s)$.

\begin{algorithm}[t]
\caption{Policy Optimization with MGRR for Linear AMDPs}
\label{alg:linear}
\begin{algorithmic}[1]
\Require Parameters $\lambda,m_0,\nu,\eta,\theta$ (specified in \Cref{thm:linear}).
\For{$\kappa=1,2,\ldots,K$}
  \State For every visited state $s$, let $\pi_\kappa(a\mid s)\propto\exp\bigl(-\eta\sum_{\kappa'<\kappa}g_{\kappa'}(s,a)\bigr)$.
  \State Execute $\pi_\kappa$, and observe the trajectory and the losses; let $\phi_{\kappa,h}\triangleq \phi(s_{\kappa,h},a_{\kappa,h})$.
  \If{$\sqrt K<\kappa\le K-\sqrt K$}
    \State For every $h$, construct regions $(\cC_j,\phat_j,\Sigmahat_j)_j$ via \Cref{alg:regions} with features $\{\phi_{k,h}\}_{k=\kappa-\sqrt K}^{\kappa-1}$.
    \State Let $u_\kappa(s,a)\gets0$ if $\phi(s,a)$ lies in a region and its layer has fewer than $\nu$ unfinished estimates, and $u_\kappa(s,a)\gets1$ otherwise; define the bonus as
    \algeq[eq:linear-bonus]{$\displaystyle b_\kappa(s)\triangleq H\theta\sum_{a:\,u_\kappa(s,a)=0}\frac{\pi_\kappa(a\mid s)}{\phat_\kappa(s,a)}+4H\sum_{a:\,u_\kappa(s,a)=1}\pi_\kappa(a\mid s),\;\forall s\in\cS.$}
    \State Initialize $\Qhat_\kappa(s,a)\gets-b_\kappa(s)-4H(H-h-1)u_\kappa(s,a)-K^{-3}$ for all $s\in\cS_h$ and $a\in\cA$.
  \EndIf
  \For{each unfinished estimate, started in an episode $k$ before $\kappa$ at layer $h$ in region $\cC_j$}
    \State Let $n$ be the number of attempts this estimate has made, including this one.
    \State Accept episode $\kappa$ with probability $\min\{1,\alpha_{k,\kappa}\}$ (\Cref{eq:rejection accept probability}), and update $Z_n$ via \Cref{eq:matrix-update}.
    \State For every $(s,a)\in\cS_h\times\cA$ with $\phi(s,a)\in\cC_j$, add to $\Qhat_k(s,a)$ the piece
    \algeq[eq:increment]{$\displaystyle v_n(s,a)\triangleq\frac{\phi(s,a)^\top\Sigmahat_j^{-1/2}Z_n\Sigmahat_j^{-1/2}\phi_{k,h}Y_{k,h}}{16d\log(1+1/\lambda)}.$}
    \State Finish the estimate if $n=M_j-1$.
  \EndFor
  \If{$\sqrt K<\kappa\le K-\sqrt K$}
    \For{each $h$ such that $u_\kappa(s_{\kappa,h},a_{\kappa,h})=0$}
      \State Let $\cC_j$ be the region of layer $h$ that contains $\phi_{\kappa,h}$.
      \State Compute the cumulative loss-minus-bonus $Y_{\kappa,h}$ via \Cref{eq:response}.
      \State Start an MGRR estimate in $\cC_j$ with $Z_0=I$ and $M_j=\lceil m_0/\phat_j\rceil$.
      \State Add $v_0(s,a)$, that is, \Cref{eq:increment} with $n=0$ and $k=\kappa$, to $\Qhat_\kappa(s,a)$ for every $(s,a)\in\cS_h\times\cA$ with $\phi(s,a)\in\cC_j$.
    \EndFor
  \EndIf
  \State Let $g_\kappa(s,a)$ be the total change of $\sum_k(\Qhat_k-\eta\Qhat_k^2)(s,a)$ during this episode.
\EndFor
\end{algorithmic}
\end{algorithm}

We estimate $\langle\phi(s,a),\xi_{k,h}\rangle$ via MGRR separately on every region. Specifically, if $u_k(s_{k,h},a_{k,h})=0$, episode $k$ starts an MGRR estimate in the region $\cC_j$ that contains $\phi_{k,h}$, with truncation level $M_j\triangleq\lceil m_0/\phat_j\rceil$. Each of the next $M_j-1$ episodes $\kappa$ makes one attempt, which accepts episode $\kappa$ with probability $\alpha_{k,\kappa}$ (\Cref{eq:rejection accept probability}). As in Matrix Geometric Resampling \citep{neu2020efficient}, the $n$-th attempt multiplies the product $Z_{n-1}$ of the previous factors ($Z_0=I$) by a factor with the features rescaled to $\Sigmahat_j^{-1/2}\phi$ and divided by $\Thetatil(d)$:
\begin{equation}\label{eq:matrix-update}
  Z_n=\left(I-\frac{\ind[\text{episode $\kappa$ is accepted and }\phi_{\kappa,h}\in\cC_j]}{8d\log(1+1/\lambda)}\Sigmahat_j^{-1/2}\phi_{\kappa,h}\phi_{\kappa,h}^\top\Sigmahat_j^{-1/2}\right)Z_{n-1}.
\end{equation}
It then gives every pair whose feature lies in $\cC_j$ a piece $v_n(s,a)$ (\Cref{eq:increment}), and episode $k$ itself gives the first piece $v_0(s,a)$. On $\cC_j$, the scaling keeps every factor between $\frac12I$ and $I$, so the products $Z_n$ and the pieces stay bounded (as every $\phi\in\cC_j$ has $\lVert\phi\rVert_{\Sigmahat_j^\dagger}^2\le4d\log(1+1/\lambda)$; formalized in \Cref{lem:lin-pieces}).

\paragraph{Running Estimates and Policies.}
For every pair $(s,a)$, the \textit{running estimate} $\Qhat_k(s,a)$ starts at the known part of $-B_k(s,a)$, and receives the pieces of the estimate started at $(k,h)$ if $\phi(s,a)$ lies in its region. By \Cref{eq:linear-B,eq:lin-response-mean}, its final value underestimates $(Q_k^{\pi_k}-B_k)(s,a)$ in expectation, negligibly if $u_k(s,a)=0$ and by at most about $H$ otherwise. The functions $u_k$, $b_k$, and $\Qhat_k$ are defined for all pairs, but they only depend on $\phi(s,a)$, the regions, and the stored estimates, so \Cref{alg:linear} only evaluates them where needed, such as at the visited states.

Unlike in \Cref{alg:tabular}, exponential weights does not run on $\Qhat_k$ directly. In \Cref{alg:tabular}, the GRR estimates are non-negative and the random stops make them underestimate, so their overestimation $\BiasTwo$ is controlled by implicit exploration \citep{neu2015explore}. In contrast, MGRR estimates can be negative, and the estimates of all pairs in a region share the same attempts, so this argument does not apply. We therefore run exponential weights on $\Qtil_k\triangleq\Qhat_k-\eta\Qhat_k^2$, an idea from bandits \citep{bartlett2008high,zimmert2022return}, so that the received loss $g_\kappa(s,a)$ is the change of $\sum_{k\le\kappa}\Qtil_k(s,a)$ during episode $\kappa$. For a random variable $X$ with $|X|\le R$ and $\eta R\le1/4$, the correction ensures $\E[\exp(\eta(X-\eta X^2-\E[X]))]\le1$ (\Cref{lem:lin-exp}), so the corrected estimates exceed their means by at most $\Otil(H\nu/\eta)$ in total with high probability, which bounds $\BiasTwo$. In exchange, the correction underestimates by $\eta\E[\Qhat_k(s,a)^2]$, which can be covered by the first term of the bonus like all other second-moment terms. The regions need $\sqrt K$ past episodes and every estimate needs at most $\sqrt K$ future ones, so estimates only start in episodes $\sqrt K<k\le K-\sqrt K$. If $\theta>1/2$ or $K<2\sqrt K$, the agent plays a fixed policy. The full procedure is given in \Cref{alg:linear}.

Formally, we have the following regret guarantee, where for simplicity we only state the parameters up to constants and logarithmic factors.

\begin{theorem}\label{thm:linear}
  For linear AMDPs and any $\delta\in(0,1/2)$, pick $\lambda=(KdH)^{-C}$ for a large enough universal constant $C$, $m_0,\nu=\Thetatil(d)$, $\eta=\Thetatil(1/(dH^3\sqrt K))$, and $\theta=\Thetatil(d^3/\sqrt K)$, with the constants in \Cref{eq:lin-params}. Then \Cref{alg:linear} ensures the following with probability at least $1-\delta$:
  \begin{equation*}
    \mR_K=\Otil\bigl(d^3H^4\sqrt K\bigr).
  \end{equation*}
  Moreover, \Cref{alg:linear} is computationally efficient.
\end{theorem}

\begin{proof}[Proof Sketch of \Cref{thm:linear}]
  At a high level, we still decompose the regret as in \Cref{eq:decomp}, with $\Qhat_k-\eta\Qhat_k^2+B_k$ in place of $\Qhat_k$, and handle $\Bonus$ by \Cref{lem:bonus}. In this sketch, we only explain why MGRR is accurate on every region, which is the key new ingredient. By our choice of $\eta$, any two policies at most $\sqrt K$ episodes apart assign every partial trajectory the same probability up to a factor of $e^{\pm1/40}$ (\Cref{lem:lin-stability}), so \Cref{lem:rejection} applies to every attempt, and with high probability, every region satisfies \Cref{eq:uniform convergence bound} (\Cref{lem:lin-good}). Now consider an estimate started in episode $k$ in a region $\cC_j$, and let $X\triangleq\Sigmahat_j^{-1/2}\Sigma_{h,j}^{\pi_k}\Sigmahat_j^{-1/2}/(16d\log(1+1/\lambda))$. Every factor in \Cref{eq:matrix-update} has conditional mean $I-X$, regardless of the history, so the pieces have total mean
  \begin{equation}\label{eq:lin-geometric}
    \begin{aligned}
    &\quad \frac{\phi(s,a)^\top\Sigmahat_j^{-1/2}}{16d\log(1+1/\lambda)}\sum_{n=0}^{M_j-1}(I-X)^n\Sigmahat_j^{-1/2}\phi_{k,h}Y_{k,h}\\
    &=\phi(s,a)^\top\Sigmahat_j^{-1/2}\bigl(16d\log(1+1/\lambda)X\bigr)^\dagger\bigl(I-(I-X)^{M_j}\bigr)\Sigmahat_j^{-1/2}\phi_{k,h}Y_{k,h}
    \end{aligned}
  \end{equation}
  conditional on $\cF_k$. By \Cref{eq:uniform convergence bound}, the smallest eigenvalue of $X$ on the range of $\Sigmahat_j$ is of order $\phat_j/d$, up to logarithmic factors, so the truncation error $(I-X)^{M_j}$ is negligible for $M_j=\lceil m_0/\phat_j\rceil$. Ignoring this error, \Cref{eq:lin-geometric} is $\phi(s,a)^\top(\Sigma_{h,j}^{\pi_k})^\dagger\phi_{k,h}Y_{k,h}$, the linear analogue of \Cref{eq:iw}, whose mean is $\langle\phi(s,a),\xi_{k,h}\rangle$ by \Cref{eq:lin-response-mean} (\Cref{lem:lin-mean}). Hence, the estimates are nearly unbiased at the estimated pairs, with second moment $\O(H^4m_0^2/\phat_k(s,a))$, where $\phat_k(s,a)$ plays the role of $q^{\pi_k}(s,a)$ in \Cref{thm:tabular}.

  The rest follows the proof of \Cref{thm:tabular} with $q^{\pi_k}(s,a)$ replaced by $\phat_k(s,a)$. \Cref{lem:lin-bias,lem:lin-reg} bound $\BiasOne$ and $\RegTerm$ by the underestimation, the second moments, and the delay cost, and \Cref{lem:lin-bonus} shows that the bonus indeed covers these three and gives a leftover $\Otil(H^3\theta)$ overhead per episode, because a region $\cC_j$ is visited with probability at most $4\phat_j$ and the features outside all regions with probability at most $4\theta$. With our choices of $m_0,\nu=\Otil(d)$, $1/\eta=\Otil(dH^3\sqrt K)$, and $\theta=\Otil(d^3/\sqrt K)$, we obtain
  \begin{equation*}
    \mR_K=\Otil\Bigl(\frac{H\nu}\eta+\eta H^5\nu^2K+H^3\theta K+H\sqrt K\Bigr)=\Otil\bigl(d^3H^4\sqrt K\bigr).
  \end{equation*}

  Moreover, \Cref{alg:linear} is computationally efficient, as \Cref{alg:regions} runs in polynomial time, every episode updates at most $\nu H$ estimates, and $\pi_k$ is only evaluated at the visited states. Every estimate keeps the vector $Z_n\Sigmahat_j^{-1/2}\phi_{k,h}Y_{k,h}$ instead of $Z_n$, so every attempt takes $\O(d^2)$ operations.
\end{proof}

\section{Extension to Delayed Feedback}\label{sec:delay}

In this section, we show how our algorithms extend to delayed AMDPs, where the losses of episode $k$ are only revealed at the end of episode $k+\fd_k$. As in \Cref{sec:tech-delay}, the only change is to postpone processing an episode until its losses arrive: episode $k$ is processed at the end of episode $k+\min\{\fd_k,\dbar\}$, and all episodes processed at the same time are processed in increasing order of their indices (as is standard in delayed AMDPs, to avoid correlation issues).

Following the skipping technique \citep{thune2019nonstochastic}, episode $k$ is skipped unless $\fd_k\le\dbar$ and $k+\fd_k\le K$; for the linear AMDP version (\Cref{alg:delayed-linear}), we additionally require that at least $\sqrt K$ episodes are processed before $k$ is processed and that $k+\fd_k\le K-\sqrt K-\dbar$, which replaces the condition $\sqrt K<k\le K-\sqrt K$ of \Cref{alg:linear}. A non-skipped episode is processed in the same way as in \Cref{alg:tabular,alg:linear}, except that its confidence set, regions, and $u_k$ are computed only from the data processed so far. A skipped episode still makes attempts for other unfinished estimates, but its losses are discarded.

\Cref{lem:rejection-general} still applies, because the algorithms never use the trajectory of an episode (for attempts, confidence sets, or regions) before processing it. Episode $\kappa$ may be processed after some later episodes, for example, when $\fd_{\kappa+1}+1<\fd_\kappa\le\dbar$. However, the policies of these later episodes, and hence their trajectories, do not depend on the trajectory of episode $\kappa$. Therefore, when episode $\kappa$ is processed and makes attempts for the unfinished estimates, its trajectory is still a fresh trajectory of $\pi_\kappa$ after conditioning on all data processed so far, which is exactly what \Cref{lem:rejection-general} requires (formalized in \Cref{lem:processing}).

Formally, we have the following regret guarantees:

\begin{theorem}[Regret of \Cref{alg:delayed-tabular}]\label{thm:delayed-tabular}
  For delayed tabular AMDPs and any $\delta\in(0,1)$, pick $\dbar=\lceil\sqrt\mD/H\rceil$, $\eta=\Thetatil(\min\{1/(H^3(1+\dbar)),1/(H^{3/2}\sqrt{SAK})\})$, and the other parameters as in \Cref{thm:tabular} (see \Cref{eq:tab-params}). Then \Cref{alg:delayed-tabular} ensures the following with probability at least $1-\delta$:
  \begin{equation*}
    \mR_K=\Otil\bigl(H^{5/2}\sqrt{SAK}+H^3\sqrt\mD+H^4S^2A\bigr).
  \end{equation*}
\end{theorem}

\begin{theorem}[Regret of \Cref{alg:delayed-linear}]\label{thm:delayed-linear}
  For delayed linear AMDPs and any $\delta\in(0,1/2)$, pick $\dbar=\lceil\sqrt{\mD/(dH^2)}\rceil$, $\eta=\Thetatil(1/(dH^3(\sqrt K+\dbar)))$, and $\lambda,m_0,\nu,\theta$ as in \Cref{thm:linear} (see \Cref{eq:lin-params}). Then \Cref{alg:delayed-linear} ensures the following with probability at least $1-\delta$:
  \begin{equation*}
    \mR_K=\Otil\bigl(d^3H^4\sqrt K+d^{3/2}H^3\sqrt\mD\bigr).
  \end{equation*}
\end{theorem}

Compared with the $\Omega(H\sqrt{SAK}+H\sqrt\mD)$ lower bound \citep{lancewicki2020learning}, our tabular bound is optimal in $S$, $A$, $K$, and $\mD$, up to $\poly(H)$ and logarithmic factors. Our linear bound is optimal in $K$ and $\mD$, up to $\poly(d,H)$ and logarithmic factors. Both bounds are the first of their kind.

\begin{proof}[Proof Sketch of \Cref{thm:delayed-tabular,thm:delayed-linear}]
  We only sketch the terms that depend on $\mD$ for \Cref{alg:delayed-tabular}; those for \Cref{alg:delayed-linear} are analogous. With delays, episode $k$ is processed at the end of episode $k+\min\{\fd_k,\dbar\}$, and only at that time does the algorithm add $-B_k$ to the received loss and use the trajectory of episode $k$ for the attempts of the unfinished estimates. Every processed episode still changes $\log\pi(a\mid s)$ by $\Otil(\eta H^2)$ (\Cref{lem:stability}). Between playing $\pi_k$ and processing episode $k$, the algorithm processes \textit{i)} at most $\fd_k$ later episodes, and \textit{ii)} some earlier episodes whose losses arrive late; summed over all $k$, the numbers of these episodes are at most $2\mD$ (\Cref{lem:counting}). Compared with \Cref{thm:tabular}, the delays then affect the analysis as follows:
  \begin{itemize}
    \item As several episodes can now be processed at the end of the same episode, the stability term of exponential weights contains the products of their contributions to the received loss. Each pair contributes $\Otil(H^4)$, and there are at most $2\mD$ such pairs (\Cref{lem:counting}), thus giving an $\Otil(\eta H^5\mD)$ overhead.
    \item The weight mismatch of $B_k$ is new. As in the non-delayed case, every $B_k$ should be weighted by $\pi_k$ (recall \Cref{lem:bonus}). But it is now only received after extra episodes and weighted by the policy at that time. This adds another $\Otil(\eta H^5\mD)$.
    \item As in standard analyses of delayed feedback, delays themselves also bring an extra term. Namely, every piece of $\Qhat_k(s,a)$ is received by a policy that has changed during some extra episodes, each adding $\Otil(\eta H^3)$ to the delay cost of every piece. The only non-standard part is that the number of pieces $\what_k(s,a)$ is random. Fortunately, every pair has at most two pieces in expectation, compared with at most one in a standard delayed AMDP. Hence, this term adds $\Otil(\eta H^4\mD)$, and the bonus does not need to change.
    \item At most $\mD/(\dbar+1)+\dbar$ episodes are skipped, each adding at most $H$, so they add $\O(H\mD/\dbar+H\dbar)$.
  \end{itemize}

  Moreover, since we want to keep the policies close over the $\O(\dbar)$ extra episodes, we must set $\eta\lesssim1/(H^3\dbar)$ (\Cref{lem:stability}). This increases the penalty $H\log A/\eta$ by $\Otil(H^4\dbar)$, but makes the terms of order $\eta H^5\mD$ at most $\Otil(H^2\mD/\dbar)$. Therefore, all the terms that depend on $\mD$ add up to
  \begin{equation*}
    \Otil\Bigl(H^4\dbar+\frac{H^2\mD}{\dbar}\Bigr)=\Otil\bigl(H^3\sqrt\mD\bigr),\quad \text{when }\dbar=\left \lceil\sqrt\mD/H\right \rceil.
  \end{equation*}

  For \Cref{alg:delayed-linear}, a similar calculation gives
  \begin{equation*}
    \Otil\Bigl(d^2H^4\dbar+\frac{dH^2\mD}{\dbar}\Bigr)=\Otil\bigl( d^{3/2}H^3\sqrt\mD \bigr),\quad \text{when }\dbar=\left \lceil\sqrt{\mD/(dH^2)}\right \rceil.\qedhere
  \end{equation*}
\end{proof}

\section{Conclusion and Future Directions}\label{sec:conclusion}

In this paper, we propose introduced feedback delays, a general technique for evaluating a changing policy in AMDPs with unknown transitions and bandit feedback, without relying on past episodes or freezing the future. It estimates the Q-function of the current policy with future episodes via (Matrix) Geometric Resampling with Rejection while the policy keeps being updated as if this feedback were delayed, and handles those rarely visited state-action pairs using a thresholding idea.

As applications, we obtain the first $\Otil(\poly(H)\sqrt{SAK})$ regret in tabular AMDPs, answering the open problem of \citet{jin2020learning} in $S$, $A$, and $K$, and the first $\Otil(\poly(d,H)\sqrt K)$ regret with a computationally efficient algorithm in linear AMDPs, both with high probability. Both algorithms extend to delayed feedback with $\Otil(H^3\sqrt \mD)$ and $\Otil(d^{3/2}H^3\sqrt \mD)$ overheads, respectively.

We conclude with several future directions:
\begin{itemize}
  \item \textit{Can we achieve optimal dependence on $H$?} Our tabular bound is $H^{3/2}$ larger than the $\Omega(H\sqrt{SAK})$ lower bound \citep{jin2018q}, and our $\mD$-dependent term $H^3\sqrt \mD$ is $H^2$ larger than the $\Omega(H\sqrt \mD)$ lower bound \citep{lancewicki2020learning}. While we made no effort to optimize $H$, the concurrent work of \citet{li2026closing} attains the optimal dependence on $H$ but not on $S$; hence, we wonder: can the two techniques be combined for $\Otil(H\sqrt{SAK})$ regret?
  \item \textit{Can we achieve optimal dependence on $d$?} Our linear bound is $d^2H^{5/2}$ larger than the $\Omega(dH^{3/2}\sqrt K)$ lower bound \citep{zhou2021nearly}. As with $H$, we made no effort to optimize $d$. The factor $d^3$ comes from learning the feature regions from the last $\sqrt K$ episodes (\Cref{sec:tech-linear}), so a more sample-efficient way to normalize the features could possibly improve it.
  \item \textit{Could our technique be applied to other scenarios?} Introduced feedback delays only require \textit{i)} the policies to change slowly, and \textit{ii)} the likelihood ratio between two of the agent's own policies on a trajectory to always be computable. Policy optimization in other settings, such as multi-agent reinforcement learning (MARL), faces the same problem of evaluating a changing policy, to which we believe our technique can also be applied.
\end{itemize}

\bibliography{references}

\onecolumn
\newpage
\appendix
\renewcommand{\appendixpagename}{\centering \LARGE Supplementary Materials}
\appendixpage

\startcontents[section]
\printcontents[section]{l}{1}{\setcounter{tocdepth}{2}}

\section{Proofs for (Delayed) Tabular AMDPs}\label{app:tabular}\label{app:tab-recall}
In this appendix, we prove \Cref{thm:delayed-tabular}, which contains \Cref{thm:tabular} as the special case without delays. We first present the delayed version of \Cref{alg:tabular} in \Cref{app:delay-alg}, which differs from \Cref{alg:tabular} only in when the episodes are processed and in skipping some episodes.

We restate several notations and formulae for the ease of reading. For an episode $\kappa$ and a layer $h$, let
\begin{equation}\label{eq:app-partial}
  \tau_\kappa^h\triangleq(s_{\kappa,0},a_{\kappa,0},\ldots,s_{\kappa,h},a_{\kappa,h})
\end{equation}
be its \emph{partial trajectory} up to layer $h$, and let $L_{\kappa,h}\triangleq\sum_{h'=h}^{H-1}\ell_\kappa(s_{\kappa,h'},a_{\kappa,h'})$ be its loss from layer $h$ onward. By the performance difference lemma \citep{kakade2002approximately}, the regret is
\begin{equation}\label{eq:app-pdl}
  \mR_K=\sum_{s\in\cS}q^\star(s)\sum_{k=1}^K\sum_{a\in\cA}\bigl(\pi_k(a\mid s)-\pi^\star(a\mid s)\bigr)Q_k^{\pi_k}(s,a).
\end{equation}
For a bonus $b_k\colon\cS\to\R_{\ge0}$, we call a function $B_k$ with $B_k=0$ on $\cS_H$ an \emph{optimistic Q-function} of $\pi_k$ for $b_k$ if it satisfies \Cref{eq:bonus-q} with ``$=$'' replaced by ``$\ge$'', that is,
\begin{equation}\label{eq:app-bonus-q}
  B_k(s,a)\ge b_k(s)+\sum_{s'\in\cS_{h+1}}\PP(s'\mid s,a)\sum_{a'\in\cA}\pi_k(a'\mid s')B_k(s',a'),\quad\forall s\in\cS_h,a\in\cA,0\le h<H.
\end{equation}
Finally, for an estimate started in episode $k$ for a pair at layer $h$, GRR accepts episode $\kappa$ with probability
\begin{equation}\label{eq:app-alpha}
  \alpha_{k,\kappa}\triangleq\frac12\prod_{h'=0}^h\frac{\pi_k(a_{\kappa,h'}\mid s_{\kappa,h'})}{\pi_\kappa(a_{\kappa,h'}\mid s_{\kappa,h'})},
\end{equation}
where we omit the dependence on $h$ from the notation. Given $\pi_k$ and $\pi_\kappa$, $\alpha_{k,\kappa}$ only depends on $\tau_\kappa^h$, so it can be computed without knowing $\PP$.

\subsection{Algorithm with Delayed Feedback}
\label{app:delay-alg}\label{app:delay-pseudo}

For a cap $\dbar\ge0$ on the delays to be chosen later, where $\dbar=0$ without delays, \Cref{alg:delayed-tabular} \emph{processes} episode $\kappa$ at the end of episode
\begin{equation*}
  t_\kappa\triangleq\kappa+\min\{\fd_\kappa,\dbar\},
\end{equation*}
or never if $t_\kappa>K$. The episodes with the same $t_\kappa$ are processed in increasing order of their indices. We call the order in which the episodes are processed the \emph{processing order}, in which episode $\kappa$ is processed before $\kappa'$ if $t_\kappa<t_{\kappa'}$, or if $t_\kappa=t_{\kappa'}$ and $\kappa<\kappa'$. Without delays, $t_\kappa=\kappa$, so the processing order is the order in which the episodes are played. In the rest of this appendix, ``processed before'' and ``processed after'' refer to the processing order.

Following the skipping technique \citep{thune2019nonstochastic}, \Cref{alg:delayed-tabular} \emph{skips} episode $\kappa$ if $\fd_\kappa>\dbar$ or $\kappa+\fd_\kappa>K$. A skipped episode starts no estimate and has $b_\kappa\triangleq0$ and $B_\kappa\triangleq0$, but it still makes attempts for the unfinished estimates when it is processed. Every non-skipped episode $\kappa$ is therefore processed at the end of episode $t_\kappa=\kappa+\fd_\kappa\le K$, right after its losses arrive.

\begin{algorithm}[htb]
\caption{Policy Optimization with GRR for Tabular AMDPs with Delayed Feedback}
\label{alg:delayed-tabular}
\begin{algorithmic}[1]
\Require Confidence parameter $\delta$; cap $\dbar$; parameters $\eta,\gamma,\theta,M,\nu,c_1$ (specified in \Cref{eq:tab-params}).
\For{$t=1,2,\ldots,K$}
  \State For every state $s$, let $\pi_t(a\mid s)\propto\exp\bigl(-\eta\sum_{t'<t}g_{t'}(s,a)\bigr)$.
  \State Initialize the received loss $g_t\gets0$.
  \State Execute $\pi_t$, and observe the trajectory $(s_{t,h},a_{t,h})_{h=0}^{H-1}$; its losses $(\ell_t(s_{t,h},a_{t,h}))_{h=0}^{H-1}$ arrive at the end of episode $t+\fd_t$.
  \For{each episode $\kappa$ with $\kappa+\min\{\fd_\kappa,\dbar\}=t$, in increasing order of $\kappa$}
  \If{$\fd_\kappa\le \dbar$ (i.e., $\kappa$ is not skipped)}
  \State Compute $\cP_\kappa$ via \Cref{eq:app-conf-set} with the transitions of the episodes processed before $\kappa$.
  \State Compute $\qlow_\kappa(s,a)=\min_{P'\in\cP_\kappa}q^{P',\pi_\kappa}(s,a)$ and the bonus
  \algeq[eq:app-tab-bonus]{$\displaystyle b_\kappa(s)\triangleq\sum_{a\in\cA}\pi_\kappa(a\mid s)\left(\frac{c_1\eta H^3}{\qlow_\kappa(s,a)}\ind[\qlow_\kappa(s,a)\ge\theta]+H\,\ind[\qlow_\kappa(s,a)<\theta]\right).$}
  \State Compute the bonus Q-function backward from the last layer, with $B_\kappa=0$ on $\cS_H$:
  \algeq[eq:app-tab-B]{$\displaystyle B_\kappa(s,a)\triangleq b_\kappa(s)+\max_{P'\in\cP_\kappa}\sum_{s'\in\cS_{h+1}}P'(s'\mid s,a)\sum_{a'}\pi_\kappa(a'\mid s')B_\kappa(s',a'),\;\forall s\in\cS_h,a\in\cA.$}
  \State Set $g_t\gets g_t-B_\kappa$.
  \EndIf
  \For{each unfinished estimate, started in an episode $k$ processed before $\kappa$ for $(s,a)\in\cS_h\times\cA$}
    \State Set $\what_k(s,a)\gets\what_k(s,a)+1$ and $g_t(s,a)\gets g_t(s,a)+L_{k,h}/2$.
    \State Accept the attempt with probability $\min\{1,\alpha_{k,\kappa}\}$, where $\alpha_{k,\kappa}$ is defined in \Cref{eq:app-alpha}.
    \State If the attempt is accepted and $(s_{\kappa,h},a_{\kappa,h})=(s,a)$, the estimate succeeds.
    \State If the estimate does not succeed, it is randomly stopped with probability $\gamma$.
    \State The estimate finishes if it succeeds, is randomly stopped, or has $\what_k(s,a)=M$.
  \EndFor
  \If{$\fd_\kappa\le \dbar$ (i.e., $\kappa$ is not skipped)}
  \For{each $h$ such that $\qlow_\kappa(s_{\kappa,h},a_{\kappa,h})\ge\theta$}
  \If{$(s_{\kappa,h},a_{\kappa,h})$ had fewer than $\nu$ unfinished estimates before processing episode $\kappa$}
    \State Start a GRR estimate for $(s_{\kappa,h},a_{\kappa,h})$ and initialize $\what_\kappa(s_{\kappa,h},a_{\kappa,h})\gets0$.
  \EndIf
  \EndFor
  \EndIf
  \EndFor
\EndFor
\end{algorithmic}
\end{algorithm}

In \Cref{alg:delayed-tabular}, the outer loop plays the episodes: $\pi_t$ is the policy of episode $t$, and $\pi_{t+1}$ is obtained from $\pi_t$ by exponential weights on the received loss $g_t$. At the end of episode $t$, the inner loop processes every episode $\kappa$ with $t_\kappa=t$ in increasing order, each adding $-B_\kappa$ and the pieces of its attempts to $g_t$. Without delays, the inner loop only processes $\kappa=t$, and \Cref{alg:delayed-tabular} is therefore exactly \Cref{alg:tabular}.

The confidence set $\cP_\kappa$ is built from the trajectories of the episodes processed before $\kappa$. Specifically, let $N_\kappa(s,a)$ be the number of visits to $(s,a)$ in these trajectories, let $\Phat_n(\cdot\mid s,a)$ be the empirical distribution of the next states of the first $n$ visits to $(s,a)$ in the processing order, and let $\beta\triangleq\log(16S^2AK/\delta)$. Then
\begin{equation}\label{eq:app-conf-set}
  \cP_\kappa\triangleq\left\{P'\ \middle|\
  \begin{aligned}
  &\bigl\lvert P'(s'\mid s,a)-\Phat_n(s'\mid s,a)\bigr\rvert\le2\sqrt{\Phat_n(s'\mid s,a)\beta/n}+5\beta/n,\\
  &\forall\,0\le h<H,\ (s,a,s')\in\cS_h\times\cA\times\cS_{h+1},\ 1\le n\le N_\kappa(s,a)
  \end{aligned}
  \right\},
\end{equation}
where the constraints of a pair $(s,a)\in\cS_h\times\cA$ are dropped if no distribution in $\Delta(\cS_{h+1})$ satisfies all of them. Without delays, this recovers \Cref{eq:conf-set}.

For every episode $\kappa$, let $\cH_{\kappa-1}$ be the $\sigma$-algebra generated by the trajectories of the episodes processed before $\kappa$, together with the internal randomness used when processing them (that is, the acceptance tests and the random stops).\footnote{While the episode processed right before $\kappa$ is usually not $\kappa-1$, for ease of reading, we define $\cH_{\kappa-1}$ analogously to $\cF_{\kappa-1}$.} Without delays, $\cH_{\kappa-1}=\cF_{\kappa-1}$. The next lemma shows that $\cH_{\kappa-1}$ plays the role of $\cF_{\kappa-1}$ in the delayed setting.

\begin{lemma}\label{lem:processing}
  For every episode $\kappa$, the following hold:
  \begin{enumerate}[label=\textup{(\roman*)}]
    \item\label{item:proc-increase} $\cH_{\kappa-1}\subseteq\cH_{\kappa'-1}$ for every episode $\kappa'$ processed after $\kappa$;
    \item\label{item:proc-measurable} the policy $\pi_\kappa$, the quantities $\cP_\kappa$, $\qlow_\kappa$, $b_\kappa$, and $B_\kappa$, and the set of estimates that are unfinished before processing $\kappa$ are all $\cH_{\kappa-1}$-measurable;
    \item\label{item:proc-fresh} conditioned on $\cH_{\kappa-1}$, the trajectory of episode $\kappa$ is distributed as a trajectory of $\pi_\kappa$ under $\PP$.
  \end{enumerate}
\end{lemma}

\begin{proof}
  Property \ref{item:proc-increase} holds as processing only adds trajectories and internal randomness.

  For \ref{item:proc-measurable}, $\pi_\kappa$ is computed at the start of episode $\kappa$ from $g_1,\ldots,g_{\kappa-1}$, which only depend on the episodes processed at the ends of episodes $1,\ldots,\kappa-1$. As $\kappa$ itself is processed at the end of episode $t_\kappa\ge\kappa$, all such episodes are processed before $\kappa$, so $\pi_\kappa$ is $\cH_{\kappa-1}$-measurable. The other quantities in \ref{item:proc-measurable} are computed from $\pi_\kappa$ and the data of the episodes processed before $\kappa$, so they are $\cH_{\kappa-1}$-measurable as well.

  For \ref{item:proc-fresh}, it suffices to show that the data generating $\cH_{\kappa-1}$ do not use the trajectory of episode $\kappa$. Every episode $j$ processed before $\kappa$ has $j\le t_j\le t_\kappa$, so its policy $\pi_j$ only uses the episodes processed at the ends of episodes $1,\ldots,j-1$, which do not include $\kappa$. Processing $j$ only uses the trajectory of $j$, the policies of earlier episodes, and new randomness for the acceptance tests and random stops, which is independent of everything else, so it does not use the trajectory of $\kappa$ either. Hence, conditioned on $\cH_{\kappa-1}$, the actions and transitions of episode $\kappa$ are still drawn from $\pi_\kappa$ and $\PP$.
\end{proof}

Two policies $\pi_{t_1}$ and $\pi_{t_2}$ with $t_1<t_2$ differ by the updates at the ends of episodes $t_1,\ldots,t_2-1$. Formally, let $\mathfrak n_t$ be the number of episodes processed at the end of episode $t$ (namely $\mathfrak n_t\triangleq \sum_\kappa \ind[t_\kappa=t]$), and let
\begin{equation*}
  N(t_1,t_2)\triangleq\sum_{t=\min\{t_1,t_2\}}^{\max\{t_1,t_2\}-1}\mathfrak n_t
\end{equation*}
be the number of episodes processed between deciding $\pi_{t_1}$ and $\pi_{t_2}$. As every processed episode only changes the received loss slightly, $\pi_{t_1}$ and $\pi_{t_2}$ are close if $N(t_1,t_2)$ is small. The next lemma bounds these numbers.

\begin{lemma}\label{lem:counting}
  The processing order of \Cref{alg:delayed-tabular} satisfies the following:
  \begin{enumerate}[label=\textup{(\alph*)}]
    \item\label{item:cnt-batch} every episode $t$ has $\mathfrak n_t\le\dbar+1$, and $\sum_t\mathfrak n_t(\mathfrak n_t-1)\le2\mD$;
    \item\label{item:cnt-between} if episode $\kappa'$ is the $n$-th episode processed after episode $\kappa$, then $N(\kappa,\kappa')\le n+2\dbar$;
    \item\label{item:cnt-delay} let $\sigma_k\triangleq|\{j<k:t_j\ge k\}|$ be the number of episodes played before $k$ but not yet processed when episode $k$ starts. If episode $k$ is not skipped and $\kappa$ is the $n$-th episode processed after $k$ (the $0$-th one being $k$ itself), then $N(k,t_\kappa)\le n+\fd_k+\sigma_k$. Moreover, $\sigma_k\le\dbar$ and $\sum_k\sigma_k\le\mD$.
  \end{enumerate}
\end{lemma}

Without delays, we have $\mathfrak n_t=1$ and $\sigma_k=0$, and the numbers in \ref{item:cnt-between} and \ref{item:cnt-delay} are simply $n$. With delays, \ref{item:cnt-between} keeps the policies close during the attempts of an estimate, and \ref{item:cnt-delay}, where $\sigma_k$ is the number of outstanding episodes of \citet{thune2019nonstochastic}, bounds the delay costs.

\begin{proof}
  For \ref{item:cnt-batch}, the episodes processed at the end of episode $t$ are $j=t-\min\{\fd_j,\dbar\}$, so they have distinct values of $\min\{\fd_j,\dbar\}\in\{0,\ldots,\dbar\}$. Hence, $\mathfrak n_t\le\dbar+1$, and the $i$-th smallest of these values is at least $i-1$, so
  \begin{equation*}
    \sum_t\frac{\mathfrak n_t(\mathfrak n_t-1)}2\le\sum_t\sum_{j:\,t_j=t}\min\{\fd_j,\dbar\}\le\mD.
  \end{equation*}

  For \ref{item:cnt-between}, we count the number of episodes $j$ such that $t_j\in[\min\{\kappa,\kappa'\},\max\{\kappa,\kappa'\}-1]$. If $\kappa<\kappa'$, every such $j$ is processed before $\kappa'$, as $t_j<\kappa'\le t_{\kappa'}$. Such a $j$ is either $\kappa$ itself, one of the $n-1$ episodes processed after $\kappa$ and before $\kappa'$, or an episode processed before $\kappa$ with $t_j\ge\kappa$. In the last case, $t_j\le t_\kappa\le\kappa+\dbar$, so $j\in[t_j-\dbar,t_j]\subseteq[\kappa-\dbar,\kappa+\dbar]\setminus\{\kappa\}$, which contains at most $2\dbar$ episodes. If $\kappa>\kappa'$, then $\kappa\le t_\kappa\le t_{\kappa'}\le\kappa'+\dbar$, and every such $j$ satisfies $j\le\kappa-1$ and $j\ge t_j-\dbar\ge\kappa'-\dbar\ge\kappa-2\dbar$, so there are at most $2\dbar$ of them.

  For \ref{item:cnt-delay}, we count the number of episodes $j$ such that $t_j\in[k,t_\kappa-1]$. If $j<k$, then $j$ is one of the $\sigma_k$ episodes by the definition of $\sigma_k$. Otherwise, $j\ge k$ and $j$ is processed before $\kappa$. Such a $j$ is either processed before $k$, in which case $k<j\le t_j\le t_k=k+\fd_k$, giving at most $\fd_k$ episodes; or $k$ itself, which is only counted if $n\ge1$; or one of the $n-1$ episodes processed after $k$ and before $\kappa$. In total, there are at most $n+\fd_k+\sigma_k$ of them. Moreover, every $j$ counted in $\sigma_k$ satisfies $k\le t_j\le j+\dbar$, so $\sigma_k\le\dbar$, and every episode $j$ is counted in $\sigma_k$ for at most $\min\{\fd_j,\dbar\}$ values of $k$, namely $j<k\le t_j$. Hence, $\sum_k\sigma_k\le\sum_j\min\{\fd_j,\dbar\}\le\mD$.
\end{proof}

\subsection{Estimates and Good Events}
\label{app:tab-est}

We now formally describe the GRR estimates. For a non-skipped episode $k$ and a pair $(s,a)$, let $u_k(s,a)\triangleq1$ if $\qlow_k(s,a)<\theta$ or $(s,a)$ has $\nu$ unfinished estimates before processing episode $k$, and $u_k(s,a)\triangleq0$ otherwise. When processing episode $k$, \Cref{alg:delayed-tabular} starts a GRR estimate for every visited pair $(s,a)=(s_{k,h},a_{k,h})$ with $u_k(s,a)=0$, which we call the estimate $(k,s,a)$. For $s\in\cS_h$, we write
\begin{equation*}
  \ind_k(s,a)\triangleq\bigl(1-u_k(s,a)\bigr)\ind[(s_{k,h},a_{k,h})=(s,a)]
\end{equation*}
for the indicator that the estimate $(k,s,a)$ starts. As $u_k$ is $\cH_{k-1}$-measurable and episode $k$ follows $\pi_k$ given $\cH_{k-1}$ (\Cref{lem:processing}), we have
\begin{equation}\label{eq:tab-start}
  \E\left[\ind_k(s,a)\middle|\cH_{k-1}\right]=\bigl(1-u_k(s,a)\bigr)q^{\pi_k}(s,a).
\end{equation}

Whenever a later episode $\kappa$ is processed, every unfinished estimate makes one \emph{attempt} with it. The attempt \emph{succeeds} if \textit{i)} it accepts episode $\kappa$, which happens with probability $\min\{1,\alpha_{k,\kappa}\}$; and \textit{ii)} episode $\kappa$ also visits $(s,a)$. For an attempt that does not succeed, we additionally \emph{stop} it with probability $\gamma$. An estimate \emph{finishes} once an attempt succeeds, it is randomly stopped, or it has made $M$ attempts.

Let $\what_k(s,a)\le M$ be the number of attempts that the estimate $(k,s,a)$ makes, with $\what_k(s,a)=0$ if it does not start. Its pieces add up to
\begin{equation}\label{eq:app-Qhat}
  \Qhat_k(s,a)=\frac{L_{k,h}\what_k(s,a)}2,\quad\forall s\in\cS_h,a\in\cA.
\end{equation}
Let $I_{k,\kappa}(s,a)\in\{0,1\}$ indicate that the estimate $(k,s,a)$, during the attempt with episode $\kappa$, either succeeds or is randomly stopped. Therefore, $\sum_\kappa I_{k,\kappa}(s,a)\le1$. \Cref{lem:accept} below shows that every attempt, regardless of the previous attempts, has $I_{k,\kappa}(s,a)=1$ with probability
\begin{equation}\label{eq:pk}
  p_k(s,a)\triangleq\gamma+(1-\gamma)\frac{q^{\pi_k}(s,a)}2\ge\frac{q^{\pi_k}(s,a)+\gamma}2.
\end{equation}

The analysis needs the confidence sets to contain $\PP$, so that $\qlow_k\le q^{\pi_k}$. For a non-skipped episode $k$, let $G_k$ be the event that $\PP$ satisfies all constraints of \Cref{eq:app-conf-set} that define $\cP_k$. On $G_k$, these constraints are feasible and hence not dropped, so $\PP\in\cP_k$. (The event $\PP\in\cP_k$ alone is weaker, as it also holds when the constraints of some pair are infeasible and thus dropped.) We have $G_k\subseteq G_{k'}$ for every $k'$ processed before $k$. Moreover, $G_k$ is $\cH_{k-1}$-measurable by \Cref{lem:processing}. Let the \emph{good event} $G$ be the intersection of all $G_k$, which holds with probability at least $1-\delta/4$ by \Cref{lem:good-event}.

\subsection{Proof of Main Theorem}
\label{app:tab-main}

This subsection proves the main theorem, using lemmas proved in the remaining subsections. Let $C$ be a large enough universal constant, $\iota\triangleq2\log(CSAHK/\delta)$, and $\dbar\le K$ the cap on the delays. We pick the following parameters:
\begin{equation}\label{eq:tab-params}
  \begin{aligned}
  &\nu=\lceil10\iota^2\rceil=\Thetatil(1),\; c_1=C\nu=\Thetatil(1), \;
  \gamma=\eta H^2,\theta=C^2\eta H^2(\nu+H)\iota=\Thetatil(\eta H^3), \;
  M=\left\lceil\frac{4\iota}\theta\right\rceil=\Thetatil\left(\frac1\theta\right), \\
  \eta&=\min\left\{\frac1{C^3H^2(\nu+H)(\iota+\dbar)},\sqrt{\frac{H\log A+\iota^3}{H^3(\nu+H),
  \iota SAK}}\right\}=\Thetatil\left(\min\left\{\frac1{H^3(1+\dbar)},\frac1{H^{3/2}\sqrt{SAK}}\right\}\right).
  \end{aligned}
\end{equation}
Here, $\nu$ caps the number of unfinished estimates of a pair, $c_1$ is the coefficient of the bonus in \Cref{eq:app-tab-bonus}, $\gamma$ is the probability of a random stop, $\theta$ is the threshold, and $M$ caps the number of attempts of an estimate. The orders follow from $\nu+H=\Thetatil(H)$ and $H\log A+\iota^3=\Thetatil(H)$. The proofs only use the below bounds (which follow trivially from \Cref{eq:tab-params}):
\begin{equation}\label{eq:tab-conditions}
  c_1\eta H^3\le H\theta,\; c_1\ge56\nu,\;(1+4\nu H)\eta H\le\frac\theta2,\;\left(M+2\dbar\right)\eta H^2(\nu+2H)\le\frac12,\;\log_2\frac2\theta\le\iota.
\end{equation}
Indeed, the first three follow from $c_1=C\nu\le C^2(\nu+H)\iota$, $C\ge56$, and $1+4\nu H\le5\nu H$. For the fourth one, $M\le4\iota/\theta+1$ and $1+2\dbar\le2(\iota+\dbar)$ give
\begin{equation*}
  \left(M+2\dbar\right)\eta H^2(\nu+2H)\le\left(\frac{4\iota}\theta+2(\iota+\dbar)\right)2\eta H^2(\nu+H)\le\frac8{C^2}+\frac4{C^3}\le\frac12.
\end{equation*}
For the last one, the definition of $\eta$ gives $1/\eta\le C^3H^2(\nu+H)(\iota+\dbar)+\sqrt{H^3(\nu+H)\iota SAK}$, so $2/\theta\le2C(1+\dbar)+2\sqrt{SAK}\le6CSAHK\le2^\iota$, as $\dbar\le K$ and $C$ is large.

For the ease of presentation, we split the bonus in \Cref{eq:app-tab-bonus} as $b_k=b_k^{(1)}+b_k^{(2)}$, where
\begin{equation}\label{eq:bonus-split}
  b_k^{(1)}(s)\triangleq c_1\eta H^3\sum_a\frac{\pi_k(a\mid s)}{\qlow_k(s,a)}\ind[\qlow_k(s,a)\ge\theta]\quad\text{and}\quad b_k^{(2)}(s)\triangleq H\sum_a\pi_k(a\mid s)\ind[\qlow_k(s,a)<\theta]
\end{equation}
cover the pairs above the threshold and the other pairs, respectively. On $G_k$, we have $q^{\pi_k}\ge\qlow_k$, and $u_k(s,a)=0$ implies $\qlow_k(s,a)\ge\theta$, so every state $s$ satisfies
\begin{equation}\label{eq:tab-Phi-bonus}
  \ind[G_k]\sum_{a:\,u_k(s,a)=0}\frac{\pi_k(a\mid s)}{q^{\pi_k}(s,a)}\le\frac{b_k^{(1)}(s)}{c_1\eta H^3}.
\end{equation}
The bias of the random stops, the stability term of exponential weights, and the delay cost of the estimates will all be bounded by $\Otil(\eta H^3)$ times the left-hand side of \Cref{eq:tab-Phi-bonus}, so $b_k^{(1)}$ covers them.

\begin{theorem}[Formal Version of \Cref{thm:tabular,thm:delayed-tabular}]\label{thm:tabular-full}
  With the parameters in \Cref{eq:tab-params} and $\dbar=\lceil\sqrt\mD/H\rceil$, \Cref{alg:delayed-tabular} ensures $\mR_K=\Otil(H^{5/2}\sqrt{SAK}+H^3\sqrt\mD+H^4S^2A)$ with probability at least $1-\delta$. In particular, without delays, \Cref{alg:tabular} ensures $\mR_K=\Otil(H^{5/2}\sqrt{SAK}+H^4S^2A)$ with probability at least $1-\delta$.
\end{theorem}

\begin{proof}[Proof of \Cref{thm:tabular-full}]
  If $\mD\ge K^2$, then $\mR_K\le HK\le H\sqrt\mD$ trivially. Hence, we assume $\mD<K^2$, so that $\dbar\le K$. Let $\mathcal K_{\mathrm{skip}}$ be the set of skipped episodes. As $\mD=\sum_k\fd_k$, at most $\mD/(\dbar+1)$ episodes have $\fd_k>\dbar$, and every other skipped episode has $k>K-\fd_k\ge K-\dbar$, so
  \begin{equation}\label{eq:tab-skip}
    |\mathcal K_{\mathrm{skip}}|\le\frac\mD{\dbar+1}+\dbar.
  \end{equation}
  Every skipped episode contributes at most $H$ to $\mR_K$. For every other episode, we write $Q_k^{\pi_k}=(Q_k^{\pi_k}-\Qhat_k)+(\Qhat_k-B_k)+B_k$ in \Cref{eq:app-pdl}, which gives
  \begin{align}
    \mR_K&\le\underbrace{\sum_{k,s,a}q^\star(s)\pi_k(a\mid s)\left(Q_k^{\pi_k}-\Qhat_k\right)(s,a)}_{\text{\BiasOne}}+\underbrace{\sum_{k,s,a}q^\star(s)\pi^\star(a\mid s)\left(\Qhat_k-Q_k^{\pi_k}\right)(s,a)}_{\text{\BiasTwo}}\notag\\
    &\quad+\underbrace{\sum_{k,s,a}q^\star(s)\left(\pi_k-\pi^\star\right)(a\mid s)\left(\Qhat_k-B_k\right)(s,a)}_{\text{\RegTerm}}+\underbrace{\sum_{k,s,a}q^\star(s)\left(\pi_k-\pi^\star\right)(a\mid s)B_k(s,a)}_{\text{\Bonus}}+H|\mathcal K_{\mathrm{skip}}|,\label{eq:tab-decomp}
  \end{align}
  where the sums over $k$ range over the non-skipped episodes. By a union bound, the good event $G$ and the events of \Cref{lem:unfinished,lem:harmonic,prop:bias1,prop:bias2,prop:reg} hold simultaneously with probability at least $1-\delta/4-\delta/8-\delta/32-3\delta/32-\delta/16-\delta/8\ge1-\delta$. In the rest of the proof, we assume that they hold. Then \Cref{prop:bias1,prop:reg} and $b_k=b_k^{(1)}+b_k^{(2)}$ give
  \begin{equation*}
    \BiasOne+\RegTerm\le\sum_{k=1}^K\sum_sq^\star(s)b_k(s)+\frac{H\log A}\eta+2\eta H^5K+\O\left(\frac\iota\eta+H+\eta H^3(\nu+H)^2\mD+\nu H\dbar\right),
  \end{equation*}
  \Cref{prop:bias2} gives $\BiasTwo=\O(\nu\iota/\eta)$, and \Cref{prop:bonus} gives
  \begin{equation*}
    \Bonus+\sum_{k=1}^K\sum_sq^\star(s)b_k(s)=\O\left(\eta H^3(\nu+H)\iota SAK+H^4S^2A\beta\iota\right).
  \end{equation*}
  Plugging these bounds and \Cref{eq:tab-skip} into \Cref{eq:tab-decomp}, we get
  \begin{align}
    \mR_K&\le\frac{H\log A+\O(\nu\iota)}\eta+2\eta H^5K+\O\left(\eta H^3(\nu+H)\iota SAK+H^4S^2A\beta\iota+H\right)\notag\\
    &\quad+\O\left(\eta H^3(\nu+H)^2\mD+\nu H\dbar\right)+H\left(\frac\mD{\dbar+1}+\dbar\right)\notag\\
    &\overset{(a)}=\frac{H\log A+\O(\iota^3)}\eta+\O\bigl(\eta H^3(\nu+H)\iota SAK+H^4S^2A\beta\iota\notag\\
    &\qquad+H^2(\nu+H)\sqrt\mD+(\nu+1)H\dbar\bigr)\label{eq:tab-final}\\
    &\overset{(b)}\le\O\Bigl(\sqrt{H^3(\nu+H)\iota SAK(H\log A+\iota^3)}+H^2(\nu+H)(\iota+\dbar)(H\log A+\iota^3)\notag\\
    &\qquad+H^4S^2A\beta\iota+H^2(\nu+H)\sqrt\mD\Bigr),\notag
  \end{align}
  where (a) used $\nu\iota\le11\iota^3$, $2\eta H^5K\le\eta H^3(\nu+H)\iota SAK$ by $S\ge H$ and $\iota\ge2$, and, if $\mD>0$,
  \begin{equation*}
    \eta H^3(\nu+H)^2\mD\le\frac{H(\nu+H)\mD}{C^3\dbar}\le H^2(\nu+H)\sqrt\mD\quad\text{and}\quad\frac{H\mD}{\dbar+1}\le H^2\sqrt\mD,
  \end{equation*}
  both by $\dbar\ge\sqrt\mD/H$ and $\eta\le1/(C^3H^2(\nu+H)\dbar)$, and (b) used
  \begin{equation*}
    \frac1\eta\le C^3H^2(\nu+H)(\iota+\dbar)+\sqrt{\frac{H^3(\nu+H)\iota SAK}{H\log A+\iota^3}}\quad\text{and}\quad\eta\le\sqrt{\frac{H\log A+\iota^3}{H^3(\nu+H)\iota SAK}}.
  \end{equation*}
  As $\nu\le11\iota^2$, $\beta\le\iota$, $H\log A+\iota^3=\Otil(H)$, and $\dbar\le\sqrt\mD/H+1$, this is $\Otil(H^{5/2}\sqrt{SAK}+H^4+H^3\sqrt\mD+H^4S^2A)=\Otil(H^{5/2}\sqrt{SAK}+H^3\sqrt\mD+H^4S^2A)$.
\end{proof}

\subsection{Properties of GRR}
\label{app:tab-key}\label{app:proof-rejection}

We first prove \Cref{lem:rejection} in a more general form that allows delays, which will also be used in \Cref{app:linear}. For a policy $\pi$, we write $\Pr_\pi$ and $\E_\pi$ for the law and the expectation of a trajectory of $\pi$ under $\PP$, drawn independently of everything else, and $\tau^h=(s_0,a_0,\ldots,s_h,a_h)$ for its partial trajectory up to layer $h$.

\begin{lemma}[Generalized Version of \Cref{lem:rejection}]\label{lem:rejection-general}
  Let episode $k$ be processed before episode $\kappa$, and fix a step $0\le h<H$. Suppose that, given $\cH_{\kappa-1}$ and the partial trajectory $\tau_\kappa^h$ of episode $\kappa$, episode $\kappa$ is accepted with probability $\alpha_{k,\kappa}\le1$, and let $A_\kappa\in\{0,1\}$ indicate this event. Then, for every bounded (and possibly matrix-valued) function $f$ of partial trajectories up to step $h$ that is fixed given $\cH_{\kappa-1}$, we have
  \begin{equation}\label{eq:rejection-general}
    \E\left[A_\kappa f(\tau_\kappa^h)\middle|\cH_{\kappa-1}\right]=\frac12\E_{\pi_k}\left[f(\tau^h)\right],
  \end{equation}
  where the expectation on the right-hand side is only over $\tau^h$.
\end{lemma}

\begin{proof}
  A partial trajectory $\tau=(s_0,a_0,\ldots,s_h,a_h)$ has probability
  \begin{equation*}
    \Pr_\pi(\tau^h=\tau)=\prod_{h'=0}^h\pi(a_{h'}\mid s_{h'})\prod_{h'=0}^{h-1}\PP(s_{h'+1}\mid s_{h'},a_{h'})
  \end{equation*}
  under a policy $\pi$. The transitions cancel in the ratio between $\pi_k$ and $\pi_\kappa$, so \Cref{eq:app-alpha} gives
  \begin{equation}\label{eq:rejection-change}
    2\alpha_{k,\kappa}\Pr_{\pi_\kappa}(\tau^h=\tau)=\Pr_{\pi_k}(\tau^h=\tau),
  \end{equation}
  where $\alpha_{k,\kappa}$ is evaluated at $\tau_\kappa^h=\tau$. As $k$ is processed before $\kappa$, \Cref{lem:processing} shows that $\pi_k$ and $\pi_\kappa$ are $\cH_{\kappa-1}$-measurable and that, conditioned on $\cH_{\kappa-1}$, $\tau_\kappa^h$ is distributed as $\tau^h$ under $\Pr_{\pi_\kappa}$. Hence, by the tower rule of expectation and \Cref{eq:rejection-change},
  \begin{equation*}
    \E\left[A_\kappa f(\tau_\kappa^h)\middle|\cH_{\kappa-1}\right]=\E\left[\alpha_{k,\kappa}f(\tau_\kappa^h)\middle|\cH_{\kappa-1}\right]=\sum_\tau\Pr_{\pi_\kappa}(\tau^h=\tau)\,\alpha_{k,\kappa}f(\tau)=\frac12\sum_\tau\Pr_{\pi_k}(\tau^h=\tau)f(\tau),
  \end{equation*}
  which is the right-hand side of \Cref{eq:rejection-general}.
\end{proof}

\begin{proof}[Proof of \Cref{lem:rejection}]
  Without delays, $\cH_{\kappa-1}=\cF_{\kappa-1}$, $k<\kappa$ means that $k$ is processed before $\kappa$, and the hypothesis of \Cref{lem:rejection} is exactly $\alpha_{k,\kappa}\le1$. Hence, \Cref{lem:rejection-general} applies, and the three claims follow by picking $f\equiv1$, $f$ the indicator of a set of partial trajectories, and $f(\tau^h)=\ind[(s_h,a_h)=(s,a)]$, respectively.
\end{proof}

GRR requires $\alpha_{k,\kappa}\le1$, which holds if the policy changes little between the episode $k$ that starts an estimate and the episodes $\kappa$ that it attempts with. The next lemma shows that, when at most $M+2\dbar$ episodes are processed between $\pi_k$ and $\pi_\kappa$, we indeed have $\alpha_{k,\kappa}\le 1$.

\begin{lemma}\label{lem:stability}
  For every episode $t$, state $s$, and action $a$, we have $\eta|g_t(s,a)|\le1$, and furthermore
  \begin{equation}\label{eq:tab-step}
    \left|\log\pi_{t+1}(a\mid s)-\log\pi_t(a\mid s)\right|\le\mathfrak n_t\,\frac{\eta H(\nu+2H)}2.
  \end{equation}
  Consequently, if $N(t_1,t_2)\le M+2\dbar$, then every partial trajectory $(s_0,a_0,\ldots,s_h,a_h)$ satisfies
  \begin{equation}\label{eq:ratio}
    e^{-1/4}\le\prod_{h'=0}^h\frac{\pi_{t_1}(a_{h'}\mid s_{h'})}{\pi_{t_2}(a_{h'}\mid s_{h'})}\le e^{1/4}.
  \end{equation}
\end{lemma}

\begin{proof}
  We first bound the contribution of a processed episode $\kappa$ to the received loss. By \Cref{eq:tab-conditions}, $c_1\eta H^3/\qlow_\kappa(s,a)\le c_1\eta H^3/\theta\le H$ whenever $\qlow_\kappa(s,a)\ge\theta$, so \Cref{eq:app-tab-bonus} gives $0\le b_\kappa\le H$. As \Cref{eq:app-tab-B} adds up $b_\kappa$ over at most $H$ layers, we then have $0\le B_\kappa\le H^2$. Moreover, as \Cref{alg:delayed-tabular} starts no estimate for a pair with $\nu$ unfinished estimates, every pair has at most $\nu$ unfinished estimates at any time. Hence, the contribution of $\kappa$ to $g_{t_\kappa}(s,a)$, namely $-B_\kappa(s,a)$ plus a piece $L_{k,h}/2\le H/2$ for every estimate of $(s,a)$ that attempts with $\kappa$, lies in $[-H^2,\nu H/2]$.

  As $g_t$ adds up the contributions of the $\mathfrak n_t$ episodes processed at the end of episode $t$, every $g_t(s,\cdot)$ takes values in an interval of length $\mathfrak n_t(\nu H/2+H^2)$, and \Cref{lem:hedge-step} gives \Cref{eq:tab-step}. Moreover, by \Cref{lem:counting}\ref{item:cnt-batch}, $\mathfrak n_t\le\dbar+1\le M+2\dbar$, so \Cref{eq:tab-conditions} gives
  \begin{equation*}
    \eta|g_t(s,a)|\le\mathfrak n_t\,\frac{\eta H(\nu+2H)}2\le\left(M+2\dbar\right)\eta H^2(\nu+2H)\le\frac12.
  \end{equation*}
  Finally, the logarithm of the ratio in \Cref{eq:ratio} is a sum of at most $H$ terms of the form $\log\pi_{t_1}(a\mid s)-\log\pi_{t_2}(a\mid s)$, each of which is at most $N(t_1,t_2)\,\eta H(\nu+2H)/2$ in absolute value by \Cref{eq:tab-step}. Hence, if $N(t_1,t_2)\le M+2\dbar$, then \Cref{eq:tab-conditions} gives
  \begin{equation*}
    \left|\log\prod_{h'=0}^h\frac{\pi_{t_1}(a_{h'}\mid s_{h'})}{\pi_{t_2}(a_{h'}\mid s_{h'})}\right|\le H\left(M+2\dbar\right)\frac{\eta H(\nu+2H)}2\le\frac14,
  \end{equation*}
  which finishes the proof.
\end{proof}

Using \Cref{lem:rejection-general,lem:stability}, we now show that every attempt finishes its estimate with probability $p_k(s,a)$, regardless of the previous attempts.

\begin{lemma}\label{lem:accept}
  Fix an estimate $(k,s,a)$ with $s\in\cS_h$ that attempts with episode $\kappa$, and let
  \begin{equation*}
    \bar\alpha_{k,\kappa}(s,a)\triangleq\ind[(s_{\kappa,h},a_{\kappa,h})=(s,a)]\,\alpha_{k,\kappa}
  \end{equation*}
  be the probability that this attempt succeeds, given $\cH_{\kappa-1}$ and the trajectory $\tau_\kappa$ of episode $\kappa$. Then $\alpha_{k,\kappa}\le e^{1/4}/2$ and $\E[\bar\alpha_{k,\kappa}(s,a)\mid\cH_{\kappa-1}]=q^{\pi_k}(s,a)/2$. Moreover, given $\cH_{\kappa-1}$ and $\tau_\kappa$, the indicators $I_{k,\kappa}(s,a)$ that the attempts with $\kappa$ finish their estimates are independent across estimates, and
  \begin{equation}\label{eq:attempt}
    \Pr\left(I_{k,\kappa}(s,a)=1\middle|\cH_{\kappa-1},\tau_\kappa\right)=\gamma+(1-\gamma)\bar\alpha_{k,\kappa}(s,a).
  \end{equation}
  In particular, taking expectations over $\tau_\kappa$ gives $\Pr(I_{k,\kappa}(s,a)=1\mid\cH_{\kappa-1})=p_k(s,a)$.
\end{lemma}

\begin{proof}
  As the estimate has made fewer than $M$ attempts before $\kappa$, episode $\kappa$ is the $n$-th episode processed after $k$ for some $n\le M$. Hence, $N(k,\kappa)\le M+2\dbar$ by \Cref{lem:counting}\ref{item:cnt-between}, and \Cref{lem:stability} gives $\alpha_{k,\kappa}\le e^{1/4}/2<1$ for every partial trajectory. Let $A_\kappa\in\{0,1\}$ indicate that the attempt accepts episode $\kappa$. Given $\cH_{\kappa-1}$ and $\tau_\kappa$, the attempt succeeds, that is, $A_\kappa\ind[(s_{\kappa,h},a_{\kappa,h})=(s,a)]=1$, with probability $\bar\alpha_{k,\kappa}(s,a)$, and otherwise the estimate is randomly stopped with probability $\gamma$, which gives \Cref{eq:attempt}. As every estimate draws its acceptance tests and random stops independently of the other estimates, these indicators are independent given $\cH_{\kappa-1}$ and $\tau_\kappa$. Finally, \Cref{lem:rejection-general} with $f(\tau^h)=\ind[(s_h,a_h)=(s,a)]$ gives
  \begin{equation*}
    \E\left[\bar\alpha_{k,\kappa}(s,a)\middle|\cH_{\kappa-1}\right]=\E\left[A_\kappa\ind[(s_{\kappa,h},a_{\kappa,h})=(s,a)]\middle|\cH_{\kappa-1}\right]=\frac{q^{\pi_k}(s,a)}2,
  \end{equation*}
  and taking expectations in \Cref{eq:attempt} gives the last claim.
\end{proof}

\Cref{alg:delayed-tabular} caps the number of attempts of every estimate at $M$, and the number of unfinished estimates of every pair at $\nu$. The next lemma shows that, on the good event, neither cap is reached with high probability. The reason is that, on $G_k$, every attempt finishes the estimate $(k,s,a)$ with probability $p_k(s,a)\ge\theta/2$ (\Cref{lem:accept}). Hence, the number of attempts of every estimate is at most a geometric random variable with mean $2/\theta$, well below $M$, and only few estimates of the same pair can be unfinished at the same time.

\begin{lemma}\label{lem:unfinished}
  With probability at least $1-\delta/8$, if the good event $G$ holds, then:
  \begin{enumerate}[label=\textup{(\roman*)}]
    \item\label{item:unf-attempts} every estimate $(k,s,a)$ with at least $M$ episodes processed after $k$ finishes, that is, $\sum_\kappa I_{k,\kappa}(s,a)=1$; and
    \item\label{item:unf-cap} before every non-skipped episode is processed, every pair has fewer than $\nu$ unfinished estimates.
  \end{enumerate}
\end{lemma}

\begin{proof}
  Both parts use that, on $G_k$, every attempt of an estimate $(k,s,a)$ finishes it with probability $p_k(s,a)\ge\theta/2$ given $\cH_{\kappa-1}$ (\Cref{lem:accept}).

  For \ref{item:unf-attempts}, fix an estimate $(k,s,a)$ started on $G_k$ with at least $M$ episodes processed after $k$. Its first $M$ attempts all fail to finish it with probability at most
  \begin{equation*}
    \Bigl(1-\frac\theta2\Bigr)^M\le e^{-\theta M/2}\le e^{-2\iota},
  \end{equation*}
  where we used $\theta M\ge4\iota$. As at most $HK$ estimates start, a union bound shows that \ref{item:unf-attempts} fails with probability at most $HKe^{-2\iota}\le\delta/32$.

  For \ref{item:unf-cap}, note that an estimate with a larger $q^{\pi_k}(s,a)$ finishes faster but also starts more often, so we group the estimates by the order of magnitude of $q^{\pi_k}(s,a)$. Fix a pair $(s,a)$ and an integer $j\ge0$, and consider the estimates $(k,s,a)$ started on $G_k$ with $q^{\pi_k}(s,a)\in(2^{-j-1},2^{-j}]$. As $q^{\pi_k}(s,a)\ge\theta$ on $G_k$ and $\log_2(2/\theta)\le\iota$ by \Cref{eq:tab-conditions}, there are at most $\iota$ such values of $j$. Every attempt of such an estimate finishes it with probability $p_k(s,a)\ge q^{\pi_k}(s,a)/2>2^{-j-2}$, so it makes $\lceil2^{j+2}\iota\rceil$ attempts without finishing with probability at most $(1-2^{-j-2})^{\lceil2^{j+2}\iota\rceil}\le e^{-\iota}$.

  Now fix a non-skipped episode $k$. If none of the above events happens, every such estimate that is unfinished before $k$ is processed was started by one of the $\lceil2^{j+2}\iota\rceil$ episodes processed right before $k$. For each of these episodes $\kappa$, let $X_\kappa\triangleq\ind[G_\kappa]\,\ind[\text{$\kappa$ starts an estimate for $(s,a)$ with }q^{\pi_\kappa}(s,a)\in(2^{-j-1},2^{-j}]]$. By \Cref{eq:tab-start}, $\E[X_\kappa\mid\cH_{\kappa-1}]\le\ind[q^{\pi_\kappa}(s,a)\le2^{-j}]\,q^{\pi_\kappa}(s,a)\le2^{-j}$, so \Cref{lem:freedman}\ref{lem:adapted-upper} with $R=1$ shows that, except with probability $e^{-\iota}$,
  \begin{equation*}
    \sum_\kappa X_\kappa\le(e-1)\lceil2^{j+2}\iota\rceil2^{-j}+\iota\le(e-1)(4\iota+1)+\iota\le8\iota+2.
  \end{equation*}
  Summing over the at most $\iota$ values of $j$, the number of unfinished estimates of $(s,a)$ before $k$ is processed is at most $\iota(8\iota+2)<10\iota^2\le\nu$. A union bound over the estimates, the pairs, the values of $j$, and the episodes $k$ shows that \ref{item:unf-cap} fails with probability at most $(HK+\iota SAK)e^{-\iota}\le3\delta/32$ for a large enough $C$.

  Combining the two parts gives the failure probability $\delta/32+3\delta/32=\delta/8$, as claimed.
\end{proof}

In particular, on $G$ and the event of \Cref{lem:unfinished}, we have $u_k(s,a)=\ind[\qlow_k(s,a)<\theta]$ for every non-skipped episode $k$ and pair $(s,a)$.

\subsection{Concentration of the Numbers of Attempts}
\label{app:tab-conc}

Several terms of the regret, such as $\BiasOne$, contain weighted sums $\sum_{k,s,a}c_k(s,a)\what_k(s,a)$ of the numbers of attempts, which we want to compare with their means $\sum_{k,s,a}\ind_k(s,a)c_k(s,a)/p_k(s,a)$. A standard martingale argument does not directly apply, because the estimates that attempt with the same episode $\kappa$ share its trajectory, and are thus correlated. We therefore prove a concentration bound in the following abstract setting, where every index $i$ stands for an estimate:
\begin{itemize}
  \item the episodes are processed one by one, and when episode $\kappa$ is processed, the set of indices $i$ that attempt with it is $\cH_{\kappa-1}$-measurable;
  \item every attempt of $i$ with $\kappa$ has a weight $c_{i,\kappa}\ge0$ that is $\cH_{\kappa-1}$-measurable, and an indicator $I_{i,\kappa}\in\{0,1\}$ that this attempt succeeds or randomly stops $i$;
  \item given $\cH_{\kappa-1}$ and the trajectory $\tau_\kappa$ of episode $\kappa$, the indicators $I_{i,\kappa}$ are independent across $i$, with means $\gamma+(1-\gamma)\bar\alpha_{i,\kappa}$ for some $\bar\alpha_{i,\kappa}\in[0,1]$ that only depend on $\cH_{\kappa-1}$ and $\tau_\kappa$;
  \item almost surely, at most $\nu H$ of the $\bar\alpha_{i,\kappa}$ are nonzero, and $\E[\gamma+(1-\gamma)\bar\alpha_{i,\kappa}\mid\cH_{\kappa-1}]=p_i>0$, where $p_i$ is $\cH_{k-1}$-measurable for the episode $k$ that starts $i$.
\end{itemize}
By \Cref{lem:accept}, the estimates of \Cref{alg:delayed-tabular} fit this setting with $i=(k,s,a)$ and $p_i=p_k(s,a)$. In particular, $\bar\alpha_{k,\kappa}(s,a)\ne0$ requires episode $\kappa$ to visit $(s,a)$, and each of the $H$ pairs it visits has at most $\nu$ unfinished estimates, so at most $\nu H$ of the $\bar\alpha_{i,\kappa}$ are nonzero.

\begin{lemma}\label{lem:attempts}
  In the setting above, let $\lambda>0$, $\delta'\in(0,1)$, and $\epsilon_i\in[0,1/2]$, where $\epsilon_i$ is $\cH_{\kappa-1}$-measurable for every episode $\kappa$ with which $i$ makes an attempt. In the following, the inner sums range over the indices $i$ that make an attempt with $\kappa$.
  \begin{enumerate}[label=\textup{(\roman*)}]
    \item\label{item:att-upper} If $(1+\nu H)\lambda c_{i,\kappa}\le\epsilon_ip_i$ for all $i$ and $\kappa$, then, with probability at least $1-\delta'$,
    \begin{equation*}
      \sum_\kappa\sum_ic_{i,\kappa}\left(1-\frac{I_{i,\kappa}}{(1-\epsilon_i)p_i}\right)\le\frac{\log(1/\delta')}\lambda.
    \end{equation*}
    \item\label{item:att-lower} If $(1+4\nu H)\lambda c_{i,\kappa}\le\epsilon_ip_i$ for all $i$ and $\kappa$, then, with probability at least $1-\delta'$,
    \begin{equation*}
      \sum_\kappa\sum_ic_{i,\kappa}\left(\frac{I_{i,\kappa}}{(1+\epsilon_i)p_i}-1\right)\le\frac{\log(1/\delta')}\lambda.
    \end{equation*}
  \end{enumerate}
\end{lemma}

\begin{proof}
  By the tower rule of expectation over the episodes in the processing order and Markov's inequality, it suffices to show that every episode $\kappa$ satisfies
  \begin{equation}\label{eq:att-step}
    \E\left[\exp\left(\lambda\sum_ic_{i,\kappa}\left(1-\frac{I_{i,\kappa}}{(1-\epsilon_i)p_i}\right)\right)\middle|\cH_{\kappa-1}\right]\le1
  \end{equation}
  for \ref{item:att-upper}, and the analogous bound for \ref{item:att-lower}. Fix $\kappa$. As the attempts with $\kappa$ share its trajectory $\tau_\kappa$, we first condition on $\tau_\kappa$, which makes them independent, and then average over $\tau_\kappa$.

  For \ref{item:att-upper}, let $x_i\triangleq\lambda c_{i,\kappa}/((1-\epsilon_i)p_i)$ and $y_i\triangleq1-e^{-x_i}$, so that $e^{-x_iI}=1-y_iI$ for $I\in\{0,1\}$ and $x_i-x_i^2/2\le y_i\le x_i$. Given $\tau_\kappa$, the independence and $1-z\le e^{-z}$ imply
  \begin{equation*}
    \E\left[e^{-\sum_ix_iI_{i,\kappa}}\middle|\cH_{\kappa-1},\tau_\kappa\right]=\prod_i\left(1-\left(\gamma+(1-\gamma)\bar\alpha_{i,\kappa}\right)y_i\right)\le\exp\left(-\gamma\sum_iy_i-Z\right),
  \end{equation*}
  where $Z\triangleq(1-\gamma)\sum_i\bar\alpha_{i,\kappa}y_i\ge0$ has at most $\nu H$ nonzero terms, so $Z^2\le\nu H\sum_i\bar\alpha_{i,\kappa}y_i^2$ by the Cauchy--Schwarz inequality. We then average over $\tau_\kappa$ using $e^{-Z}\le1-Z+Z^2/2$, $1+z\le e^z$, $\E[(1-\gamma)\bar\alpha_{i,\kappa}\mid\cH_{\kappa-1}]=p_i-\gamma$, and $\E[\bar\alpha_{i,\kappa}\mid\cH_{\kappa-1}]\le p_i$, which gives
  \begin{align*}
    \log\E\left[e^{-\sum_ix_iI_{i,\kappa}}\middle|\cH_{\kappa-1}\right]&\le-\gamma\sum_iy_i-\sum_i(p_i-\gamma)y_i+\frac{\nu H}2\sum_ip_iy_i^2\le\sum_ip_i\left(-x_i+\frac{1+\nu H}2x_i^2\right)\\
    &=\sum_i\frac{\lambda c_{i,\kappa}}{1-\epsilon_i}\left(-1+\frac{(1+\nu H)\lambda c_{i,\kappa}}{2(1-\epsilon_i)p_i}\right).
  \end{align*}
  Adding $\lambda\sum_ic_{i,\kappa}$ to both sides, and then using $(1+\nu H)\lambda c_{i,\kappa}\le\epsilon_ip_i$ and $\epsilon_i\le1/2$, the logarithm of the left-hand side of \Cref{eq:att-step} is at most
  \begin{equation*}
    \sum_i\frac{\lambda c_{i,\kappa}}{1-\epsilon_i}\left(-\epsilon_i+\frac{(1+\nu H)\lambda c_{i,\kappa}}{2(1-\epsilon_i)p_i}\right)\le\sum_i\frac{\lambda c_{i,\kappa}}{1-\epsilon_i}\left(-\epsilon_i+\frac{\epsilon_i}{2(1-\epsilon_i)}\right)\le0.
  \end{equation*}

  For \ref{item:att-lower}, let $x_i\triangleq\lambda c_{i,\kappa}/((1+\epsilon_i)p_i)\le1/(8\nu H)$ and $z_i\triangleq e^{x_i}-1$, so that $x_i\le z_i\le x_i+x_i^2\le2x_i$. The same two steps, now with $Z'\triangleq(1-\gamma)\sum_i\bar\alpha_{i,\kappa}z_i\in[0,1/4]$, $Z'^2\le4\nu H\sum_i\bar\alpha_{i,\kappa}x_i^2$, and $e^{Z'}\le1+Z'+Z'^2$, give
  \begin{equation*}
    \log\E\left[e^{\sum_ix_iI_{i,\kappa}}\middle|\cH_{\kappa-1}\right]\le\sum_ip_i\left(z_i+4\nu Hx_i^2\right)\le\sum_ip_i\left(x_i+(1+4\nu H)x_i^2\right).
  \end{equation*}
  Subtracting $\lambda\sum_ic_{i,\kappa}$, now with $(1+4\nu H)\lambda c_{i,\kappa}\le\epsilon_ip_i$, we get
  \begin{equation*}
    \sum_i\frac{\lambda c_{i,\kappa}}{1+\epsilon_i}\left(-\epsilon_i+\frac{(1+4\nu H)\lambda c_{i,\kappa}}{(1+\epsilon_i)p_i}\right)\le\sum_i\frac{\lambda c_{i,\kappa}}{1+\epsilon_i}\left(-\epsilon_i+\frac{\epsilon_i}{1+\epsilon_i}\right)\le0,
  \end{equation*}
  which finishes the proof.
\end{proof}

In particular, if every attempt of an estimate $i$ has the same weight $c_i$, then $\sum_\kappa c_{i,\kappa}=c_i\what_i$. Hence, under the conditions of \ref{item:att-upper}, we have
\begin{equation}\label{eq:att-use}
  \sum_ic_i\what_i\le\sum_i\frac{c_i\sum_\kappa I_{i,\kappa}}{(1-\epsilon_i)p_i}+\frac{\log(1/\delta')}\lambda,
\end{equation}
and, under the conditions of \ref{item:att-lower}, we have
\begin{equation}\label{eq:att-use-lower}
  \sum_ic_i\what_i\ge\sum_i\frac{c_i\sum_\kappa I_{i,\kappa}}{(1+\epsilon_i)p_i}-\frac{\log(1/\delta')}\lambda.
\end{equation}

\subsection{Bounding \texorpdfstring{\BiasOne}{Bias-1} and \texorpdfstring{\BiasTwo}{Bias-2}}
\label{app:tab-bias}\label{app:tab-bias1}\label{app:tab-bias2}

$\BiasOne$ measures how much the estimates underestimate $Q_k^{\pi_k}$ under $\pi_k$. The estimates are unbiased on the estimated pairs except for the random stops, while every other unestimated pair contributes at most $H$. We show that (half of) $b_k^{(1)}$ covers the random stops and the concentration terms (\Cref{prop:reg} below uses the other half), and $b_k^{(2)}$ covers the latter. $\BiasTwo$ measures how much the estimates overestimate $Q_k^{\pi_k}$ under $\pi^\star$, which we bound by an implicit exploration argument similar to \citet{neu2015explore}. In both proofs, we replace the numbers of attempts by their means via \Cref{lem:attempts} (and more specifically, \Cref{eq:att-use,eq:att-use-lower}).

\begin{lemma}[Bounding \BiasOne]\label{prop:bias1}
  With probability at least $1-3\delta/32$, if the good event $G$ and the event of \Cref{lem:unfinished} hold, then
  \begin{equation*}
    \BiasOne\le\sum_{k=1}^K\sum_sq^\star(s)\left(b_k^{(2)}(s)+\frac12b_k^{(1)}(s)\right)+\O\left(\frac\iota\eta\right).
  \end{equation*}
\end{lemma}

\begin{proof}
  Under $G$ and the event of \Cref{lem:unfinished}, $\qlow_k\le q^{\pi_k}$, $u_k(s,a)=\ind[\qlow_k(s,a)<\theta]$, and every estimate $(k,s,a)$ with at least $M$ episodes processed after $k$ has $\sum_\kappa I_{k,\kappa}(s,a)=1$. Let $\mathcal K_M$ be the set of non-skipped episodes with at least $M$ episodes processed after them, which misses at most $M$ non-skipped episodes. For $s\in\cS_h$, let $c_k(s,a)\triangleq q^\star(s)\pi_k(a\mid s)L_{k,h}/2$, so that $q^\star(s)\pi_k(a\mid s)\Qhat_k(s,a)=c_k(s,a)\what_k(s,a)$. For $k\in\mathcal K_M$, define
  \begin{equation*}
    X_k\triangleq\ind[G_k]\sum_{s,a}\ind_k(s,a)\frac{c_k(s,a)}{p_k(s,a)}\quad\text{and}\quad Y_k\triangleq(1+4\nu H)\eta\,\ind[G_k]\sum_{s,a}\ind_k(s,a)\frac{c_k(s,a)^2}{p_k(s,a)^2},
  \end{equation*}
  where $X_k$ replaces the numbers of attempts by their means, and $Y_k$ is the second-order term of this replacement. As $\ind[G_k]=1$ under $G$, we decompose $\BiasOne$ as
  \begin{equation}\label{eq:bias1-decomp}
    \begin{aligned}
    \BiasOne&=\sum_{k\in\mathcal K_M}\Bigl(X_k-\sum_{s,a}c_k(s,a)\what_k(s,a)\Bigr)+\sum_{k\in\mathcal K_M}\bigl(\E[X_k\mid\cH_{k-1}]-X_k\bigr)\\
    &\quad+\sum_{k\in\mathcal K_M}\Bigl(\sum_{s,a}q^\star(s)\pi_k(a\mid s)Q_k^{\pi_k}(s,a)-\E[X_k\mid\cH_{k-1}]\Bigr)+\sum_{k\notin\mathcal K_M}\sum_{s,a}q^\star(s)\pi_k(a\mid s)\left(Q_k^{\pi_k}-\Qhat_k\right)(s,a).
    \end{aligned}
  \end{equation}

  For the first term, we apply \Cref{lem:attempts}\ref{item:att-lower} to the estimates $(k,s,a)$ with $k\in\mathcal K_M$, with the weight $\ind[G_k]c_k(s,a)$ at every attempt, $\lambda=\eta$, $\delta'=\delta/32$, and
  \begin{equation*}
    \epsilon_k(s,a)\triangleq\frac{(1+4\nu H)\eta\,\ind[G_k]c_k(s,a)}{p_k(s,a)}\overset{(a)}\le\frac{(1+4\nu H)\eta H}\theta\overset{(b)}\le\frac12,
  \end{equation*}
  where (a) used $c_k(s,a)\le H/2$ and $p_k(s,a)\ge\theta/2$ on $G_k$, and (b) used \Cref{eq:tab-conditions}. As these estimates have $\sum_\kappa I_{k,\kappa}(s,a)=1$, \Cref{eq:att-use-lower} and $1/(1+\epsilon)\ge1-\epsilon$ imply that, except with probability $\delta/32$, the first term in \Cref{eq:bias1-decomp} is at most $\sum_kY_k+\iota/\eta$.

  For the second term, by the Cauchy--Schwarz inequality over the at most $H$ estimates that start in episode $k$, $c_k(s,a)\le q^\star(s)\pi_k(a\mid s)H/2$, $p_k\ge q^{\pi_k}/2$, \Cref{eq:tab-start}, and $q^\star(s)\pi_k(a\mid s)\le1$, we have
  \begin{align*}
    \E[X_k^2\mid\cH_{k-1}]&\le H\,\E\left[\ind[G_k]\sum_{s,a}\ind_k(s,a)\frac{c_k(s,a)^2}{p_k(s,a)^2}\middle|\cH_{k-1}\right]\\&\le H^3\ind[G_k]\sum_sq^\star(s)\sum_{a:\,u_k(s,a)=0}\frac{\pi_k(a\mid s)}{q^{\pi_k}(s,a)}\le\frac1{c_1\eta}\sum_sq^\star(s)b_k^{(1)}(s),
  \end{align*}
  where the last step used \Cref{eq:tab-Phi-bonus}. Hence, by Freedman's inequality (\Cref{lem:freedman}\ref{lem:adapted-lower} with $\lambda=\eta$ and $\delta'=\delta/32$), except with probability $\delta/32$, the second term is at most $\sum_{k,s}q^\star(s)b_k^{(1)}(s)/(2c_1)+\iota/\eta$. In the same way,
  \begin{equation*}
    \E[Y_k\mid\cH_{k-1}]\le(1+4\nu H)\eta H^2\ind[G_k]\sum_sq^\star(s)\sum_{a:\,u_k(s,a)=0}\pi_k(a\mid s)/q^{\pi_k}(s,a)\le(5\nu/c_1)\sum_sq^\star(s)b_k^{(1)}(s).
  \end{equation*}

  Moreover, $0\le Y_k\le(1+4\nu H)\eta H^3/\theta^2=\O(1/(\eta\iota))$ by \Cref{eq:tab-params}, as at most $H$ estimates start in episode $k$. Hence, by Freedman's inequality (\Cref{lem:freedman}\ref{lem:adapted-upper} with $\delta'=\delta/32$), except with probability $\delta/32$,
  \begin{equation*}
    \sum_kY_k\le\frac{5(e-1)\nu}{c_1}\sum_{k,s}q^\star(s)b_k^{(1)}(s)+\O\left(\frac1\eta\right)\le\frac{9\nu}{c_1}\sum_{k,s}q^\star(s)b_k^{(1)}(s)+\O\left(\frac1\eta\right).
  \end{equation*}

  For the third term, \Cref{eq:tab-start} and $\E[L_{k,h}\mid\cH_{k-1},(s_{k,h},a_{k,h})=(s,a)]=Q_k^{\pi_k}(s,a)$ give $\E[\ind_k(s,a)L_{k,h}\mid\cH_{k-1}]=(1-u_k(s,a))q^{\pi_k}(s,a)Q_k^{\pi_k}(s,a)$. Hence, every $k\in\mathcal K_M$ satisfies
  \begin{align*}
    &\quad\sum_{s,a}q^\star(s)\pi_k(a\mid s)Q_k^{\pi_k}(s,a)-\E[X_k\mid\cH_{k-1}]\\
    &=\sum_{s,a}q^\star(s)\pi_k(a\mid s)Q_k^{\pi_k}(s,a)\left(1-(1-u_k(s,a))\frac{q^{\pi_k}(s,a)}{2p_k(s,a)}\right)\\
    &\overset{(a)}\le\sum_{s,a}q^\star(s)\pi_k(a\mid s)u_k(s,a)Q_k^{\pi_k}(s,a)+\sum_{s,a}(1-u_k(s,a))q^\star(s)\pi_k(a\mid s)Q_k^{\pi_k}(s,a)\frac{2\gamma}{q^{\pi_k}(s,a)}\\
    &\overset{(b)}\le\sum_sq^\star(s)b_k^{(2)}(s)+\frac2{c_1}\sum_sq^\star(s)b_k^{(1)}(s),
  \end{align*}
  where (a) used $q^{\pi_k}(s,a)\le2p_k(s,a)\le q^{\pi_k}(s,a)+2\gamma$, and (b) used $Q_k^{\pi_k}\le H$, $u_k(s,a)=\ind[\qlow_k(s,a)<\theta]$, $\gamma=\eta H^2$, and \Cref{eq:tab-Phi-bonus}.

  For the last term, every $k\notin\mathcal K_M$ contributes at most $\sum_sq^\star(s)\sum_a\pi_k(a\mid s)Q_k^{\pi_k}(s,a)\le H^2$, and $H^2M\le4/(C^2\eta(\nu+H))+H^2=\O(1/\eta)$.

  Combining the four terms, we get
  \begin{equation*}
    \BiasOne\le\sum_{k=1}^K\sum_sq^\star(s)\left(b_k^{(2)}(s)+\frac{1/2+9\nu+2}{c_1}b_k^{(1)}(s)\right)+\O\left(\frac\iota\eta\right),
  \end{equation*}
  which proves the claim as $c_1\ge56\nu$ by \Cref{eq:tab-conditions}.
\end{proof}

\begin{lemma}[Bounding \BiasTwo]\label{prop:bias2}
  With probability at least $1-\delta/16$, we have $\BiasTwo=\O(\nu\iota/\eta)$.
\end{lemma}

\begin{proof}
  We first replace the numbers of attempts by their means using \Cref{lem:attempts}\ref{item:att-upper}, and then bound the starts by the implicit exploration argument of \citet[Lemma~1]{neu2015explore}. For $s\in\cS_h$, let $c_k^\star(s,a)\triangleq q^\star(s)\pi^\star(a\mid s)L_{k,h}/2$. We apply \Cref{lem:attempts}\ref{item:att-upper} with the weight $c_k^\star(s,a)$ at every attempt, $\lambda=\gamma/(2H(1+\nu H))$, $\delta'=\delta/32$, and
  \begin{equation*}
    \epsilon_k(s,a)\triangleq\frac{(1+\nu H)\lambda c_k^\star(s,a)}{p_k(s,a)}=\frac{\gamma\,q^\star(s)\pi^\star(a\mid s)L_{k,h}}{4Hp_k(s,a)}\le\frac\gamma{4p_k(s,a)}\le\frac14,
  \end{equation*}
  where the last step used $p_k(s,a)\ge\gamma$ by \Cref{eq:pk}. Moreover, $(1-\epsilon_k(s,a))p_k(s,a)\ge p_k(s,a)-\gamma/4\ge(q^{\pi_k}(s,a)+\gamma/2)/2$ by \Cref{eq:pk}. Hence, by \Cref{eq:att-use} and $\sum_\kappa I_{k,\kappa}(s,a)\le1$, except with probability $\delta/32$,
  \begin{equation}\label{eq:bias2-attempts}
    \sum_{k,s,a}q^\star(s)\pi^\star(a\mid s)\Qhat_k(s,a)\le\sum_kX_k+\frac{2H(1+\nu H)\iota}\gamma,\quad\text{where }X_k\triangleq\sum_{s,a}\ind_k(s,a)\frac{q^\star(s)\pi^\star(a\mid s)L_{k,h}}{q^{\pi_k}(s,a)+\gamma/2}.
  \end{equation}

  Next, we bound $\sum_kX_k$. Let $X_{k,h}$ be the part of $X_k$ with $s\in\cS_h$. By $\log(1+v)\ge v/(1+v/2)$ for $v\ge0$, every $y\in(0,1]$ and $z\in[0,1]$ satisfy $z/(y+\gamma/2)\le\gamma^{-1}\log(1+\gamma z/y)$. Hence, we have
  \begin{align*}
    \E\left[\exp\left(\frac\gamma{H^2}X_k\right)\middle|\cH_{k-1}\right]&\overset{(a)}\le\frac1H\sum_h\E\left[\exp\left(\frac\gamma HX_{k,h}\right)\middle|\cH_{k-1}\right]\\
    &\overset{(b)}\le\frac1H\sum_h\E\left[\prod_{s\in\cS_h,a}\left(1+\frac{\gamma\ind_k(s,a)L_{k,h}}{Hq^{\pi_k}(s,a)}\right)^{q^\star(s)\pi^\star(a\mid s)}\middle|\cH_{k-1}\right]\\
    &\overset{(c)}\le1+\frac\gamma{H^2}\sum_h\sum_{s\in\cS_h,a}\frac{q^\star(s)\pi^\star(a\mid s)\E\left[\ind_k(s,a)L_{k,h}\middle|\cH_{k-1}\right]}{q^{\pi_k}(s,a)}\\
    &\overset{(d)}\le\exp\left(\frac\gamma{H^2}\sum_sq^\star(s)\sum_a\pi^\star(a\mid s)Q_k^{\pi_k}(s,a)\right),
  \end{align*}
  where the products only range over the pairs with $q^{\pi_k}(s,a)>0$, as the other pairs are almost surely not visited and contribute zero to $X_k$, (a) used the convexity of $\exp$, (b) used the above inequality with $y=q^{\pi_k}(s,a)$ and $z=L_{k,h}/H$, (c) used that at most one factor differs from $1$ and $(1+v)^w\le1+wv$ for $w\in[0,1]$, and (d) used $\E[\ind_k(s,a)L_{k,h}\mid\cH_{k-1}]\le q^{\pi_k}(s,a)Q_k^{\pi_k}(s,a)$ and $1+x\le e^x$. As $\pi^\star$ is fixed and $q^{\pi_k}$ and $Q_k^{\pi_k}$ are $\cH_{k-1}$-measurable, the tower rule of expectation and Markov's inequality give that, except with probability $\delta/32$,
  \begin{equation}\label{eq:bias2-ix}
    \sum_kX_k\le\sum_{k,s,a}q^\star(s)\pi^\star(a\mid s)Q_k^{\pi_k}(s,a)+\frac{H^2\iota}\gamma.
  \end{equation}

  Combining \Cref{eq:bias2-attempts,eq:bias2-ix}, except with probability $\delta/16$, we have
  \begin{equation*}
    \BiasTwo=\sum_{k,s,a}q^\star(s)\pi^\star(a\mid s)\bigl(\Qhat_k(s,a)-Q_k^{\pi_k}(s,a)\bigr)\le\frac{\bigl(2H(1+\nu H)+H^2\bigr)\iota}\gamma\le\frac{(2\nu+3)\iota}\eta,
  \end{equation*}
  where the last step used $\gamma=\eta H^2$.
\end{proof}

\subsection{Bounding \texorpdfstring{\RegTerm}{Reg-Term}}
\label{app:tab-reg}

$\RegTerm$ is the regret of exponential weights on the estimates $\Qhat_k-B_k$, which it receives late: the pieces of every estimate arrive over several episodes, and, with delays, $-B_k$ arrives when episode $k$ is processed. As in the analysis of delayed feedback \citep{jin2022near,lancewicki2023delay}, we split $\RegTerm$ into the regret on the received losses and the delay costs, which grow with the second moments of the numbers of attempts and with the delays.

\begin{lemma}[Bounding \RegTerm]\label{prop:reg}
  With probability at least $1-\delta/8$, if the good event $G$ holds, then
  \begin{equation*}
    \RegTerm\le\frac{H\log A}\eta+2\eta H^5K+\frac12\sum_{k=1}^K\sum_sq^\star(s)b_k^{(1)}(s)+\O\left(\frac1\eta+H+\eta H^3(\nu+H)^2\mD+\nu H\dbar\right).
  \end{equation*}
\end{lemma}

\begin{proof}
  We first bound $\RegTerm$ on every sample path in terms of the numbers of attempts, and then concentrate them. Fix $s\in\cS_h$. For an estimate $(k,s,a)$, let $\kappa_n$ be the $n$-th episode processed after $k$, so that its $n$-th attempt adds the piece $L_{k,h}/2$ to $g_{t_{\kappa_n}}(s,a)$, while processing episode $k$ adds $-B_k$ to $g_{t_k}$. Hence, $\sum_tg_t(s,a)=\sum_k(\Qhat_k-B_k)(s,a)$, and
  \begin{align*}
    &\quad\sum_k\sum_a\left(\pi_k-\pi^\star\right)(a\mid s)\,(\Qhat_k-B_k)(s,a)\\
    &=\underbrace{\sum_{t=1}^K\sum_a\left(\pi_t-\pi^\star\right)(a\mid s)\,g_t(s,a)}_{\textsc{Exp-Weights}}+\underbrace{\sum_k\sum_a\left(\pi_{t_k}-\pi_k\right)(a\mid s)B_k(s,a)}_{\textsc{Delayed-Bonus}}+\underbrace{\sum_k\sum_a\frac{L_{k,h}}2\sum_{n=1}^{\what_k(s,a)}\left(\pi_k-\pi_{t_{\kappa_n}}\right)(a\mid s)}_{\textsc{Delay-Cost}}.
  \end{align*}
  By \Cref{lem:counting}\ref{item:cnt-delay} and \Cref{eq:tab-step}, every $0\le n\le M$ satisfies
  \begin{equation}\label{eq:tab-delayed-ratio}
    \left|\log\pi_{t_{\kappa_n}}(a\mid s)-\log\pi_k(a\mid s)\right|\le\left(n+\fd_k+\sigma_k\right)\frac{\eta H(\nu+2H)}2\le\frac14,
  \end{equation}
  where $\kappa_0=k$, and the last step used $\fd_k+\sigma_k\le2\dbar$ and \Cref{eq:tab-conditions}.

  For the \textsc{Exp-Weights} term, $\eta|g_t(s,a)|\le1$ by \Cref{lem:stability}, so the standard exponential weights analysis (\Cref{lem:hedge}) applies. Let $U_\kappa(s,a)\le\nu$ be the number of estimates of $(s,a)$ that attempt with $\kappa$, and let $x_\kappa(s,a)\le U_\kappa(s,a)H/2$ be the sum of their pieces. As $g_t(s,a)=\sum_{\kappa:\,t_\kappa=t}(x_\kappa(s,a)-B_\kappa(s,a))$ and $0\le B_\kappa\le H^2$, we have
  \begin{equation*}
    g_t(s,a)^2\le2\Bigl(\sum_{\kappa:\,t_\kappa=t}x_\kappa(s,a)\Bigr)^2+2\mathfrak n_t^2H^4\le\frac{\nu H^2}2\sum_{\kappa:\,t_\kappa=t}U_\kappa(s,a)+\frac{\nu^2H^2}2\mathfrak n_t(\mathfrak n_t-1)+2\mathfrak n_t^2H^4,
  \end{equation*}
  where the second step used $x_\kappa(s,a)^2\le\nu H^2U_\kappa(s,a)/4$ and $x_\kappa(s,a)x_{\kappa'}(s,a)\le\nu^2H^2/4$ for $\kappa\ne\kappa'$. Hence, we have
  \begin{align*}
    \textsc{Exp-Weights}&\overset{(a)}\le\frac{\log A}\eta+\eta\sum_{t,a}\pi_t(a\mid s)g_t(s,a)^2\\
    &\overset{(b)}\le\frac{\log A}\eta+\frac{\eta\nu H^2}2\sum_{\kappa,a}\pi_{t_\kappa}(a\mid s)U_\kappa(s,a)+\eta\sum_t\left(\frac{\nu^2H^2}2\mathfrak n_t(\mathfrak n_t-1)+2\mathfrak n_t^2H^4\right)\\
    &\overset{(c)}\le\frac{\log A}\eta+\eta\nu H^2\sum_{k,a}\pi_k(a\mid s)\what_k(s,a)+2\eta H^4K+\eta\left(\nu^2H^2+4H^4\right)\mD,
  \end{align*}
  where (a) used \Cref{lem:hedge}, (b) used the bound on $g_t(s,a)^2$, and (c) used $\sum_t\mathfrak n_t\le K$, \Cref{lem:counting}\ref{item:cnt-batch}, and $\sum_\kappa\pi_{t_\kappa}(a\mid s)U_\kappa(s,a)=\sum_k\sum_{n=1}^{\what_k(s,a)}\pi_{t_{\kappa_n}}(a\mid s)\le2\sum_k\pi_k(a\mid s)\what_k(s,a)$ by \Cref{eq:tab-delayed-ratio}.

  For the \textsc{Delayed-Bonus} term, \Cref{eq:tab-delayed-ratio} with $n=0$, $e^x-1\le2x$ for $0\le x\le1$, $0\le B_k\le H^2$, and $\sum_k(\fd_k+\sigma_k)\le2\mD$ (\Cref{lem:counting}\ref{item:cnt-delay}) give
  \begin{equation*}
    \textsc{Delayed-Bonus}\le\sum_k\left(\fd_k+\sigma_k\right)\eta H(\nu+2H)\sum_a\pi_k(a\mid s)B_k(s,a)\le2\eta H^3(\nu+2H)\mD.
  \end{equation*}

  For the \textsc{Delay-Cost} term, by $L_{k,h}\le H$, \Cref{eq:tab-delayed-ratio}, and $e^{-x}\ge1-x$, we have
  \begin{align*}
    \textsc{Delay-Cost}&\le\sum_{k,a}\frac H2\sum_{n=1}^{\what_k(s,a)}\left(n+\fd_k+\sigma_k\right)\frac{\eta H(\nu+2H)}2\pi_k(a\mid s)\\
    &=\frac{\eta H^2(\nu+2H)}8\sum_{k,a}\pi_k(a\mid s)\what_k(s,a)\left(\what_k(s,a)+1+2\fd_k+2\sigma_k\right).
  \end{align*}

  Summing the three bounds over $s$ with weights $q^\star(s)$, where $\sum_sq^\star(s)=H$, we get
  \begin{equation}\label{eq:tab-reg-pathwise}
    \begin{aligned}
    \RegTerm&\le\frac{H\log A}\eta+2\eta H^5K+\eta H^2\sum_{k,s,a}q^\star(s)\pi_k(a\mid s)\what_k(s,a)\left(\nu+\frac{\nu+2H}8\left(\what_k(s,a)+1\right)\right)\\
    &\quad+\frac{\eta H^2(\nu+2H)}4\sum_{k,s,a}q^\star(s)\pi_k(a\mid s)\left(\fd_k+\sigma_k\right)\what_k(s,a)+\O\left(\eta H^3(\nu+H)^2\mD\right),
    \end{aligned}
  \end{equation}
  where the last term collects $\eta H(\nu^2H^2+4H^4)\mD$ and $2\eta H^4(\nu+2H)\mD$. It remains to concentrate the two sums in \Cref{eq:tab-reg-pathwise}, where only the estimates that start contribute.

  The first sum is quadratic in $\what_k(s,a)$, so we apply \Cref{lem:attempts}\ref{item:att-upper} with a weight that increases with the attempts. Specifically, at the $n$-th attempt of the estimate $(k,s,a)$, we use the weight
  \begin{equation*}
    c_n\triangleq\ind[G_k]\,q^\star(s)\pi_k(a\mid s)H^2\left(\nu+\frac{\nu+2H}4\left(n+\frac2{p_k(s,a)}\right)\right),
  \end{equation*}
  together with $\epsilon_k(s,a)=1/2$, $\lambda=\theta^2/(24\iota(1+\nu H)H^2(\nu+2H))$, and $\delta'=\delta/32$. As $n\le M\le4\iota/\theta+1$ and $p_k(s,a)\ge\theta/2$ on $G_k$, we have
  \begin{equation*}
    c_n\le H^2\left(\nu+\frac{\nu+2H}4\left(\frac{4\iota}\theta+1+\frac4\theta\right)\right)\le\frac{3\iota H^2(\nu+2H)}\theta,
  \end{equation*}
  so $(1+\nu H)\lambda c_n\le\theta/8\le p_k(s,a)/2$ on $G_k$. Moreover, as only the last attempt can have $I_{k,\kappa}(s,a)=1$, every estimate with $w\triangleq\what_k(s,a)$ attempts satisfies
  \begin{align*}
    \sum_{n=1}^wc_n\left(1-\frac{2I_{k,\kappa_n}(s,a)}{p_k(s,a)}\right)&\ge\sum_{n=1}^wc_n-\frac{2c_w}{p_k(s,a)}\\
    &=\ind[G_k]\,q^\star(s)\pi_k(a\mid s)H^2\left(\nu w+\frac{\nu+2H}8w(w+1)-\frac{2\nu}{p_k(s,a)}-\frac{\nu+2H}{p_k(s,a)^2}\right).
  \end{align*}
  Hence, by \Cref{lem:attempts}\ref{item:att-upper}, $p_k\ge q^{\pi_k}/2$, and $\ind[G_k]=1$, except with probability $\delta/32$, we have
  \begin{align*}
    &\quad\eta H^2\sum_{k,s,a}q^\star(s)\pi_k(a\mid s)\what_k(s,a)\left(\nu+\frac{\nu+2H}8\left(\what_k(s,a)+1\right)\right)\\
    &\le\eta H^2\sum_{k,s,a}\ind_k(s,a)q^\star(s)\pi_k(a\mid s)\left(\frac{2\nu}{p_k(s,a)}+\frac{\nu+2H}{p_k(s,a)^2}\right)+\frac{\eta\iota}\lambda\le\sum_kZ_k+\O\left(\frac1\eta\right),
  \end{align*}
  where $Z_k\triangleq4\eta H^2\ind[G_k]\sum_{s,a}\ind_k(s,a)q^\star(s)\pi_k(a\mid s)(\nu/q^{\pi_k}(s,a)+(\nu+2H)/q^{\pi_k}(s,a)^2)$, and the last step used
  \begin{equation*}
    \frac{\eta\iota}\lambda=\frac{24\eta\iota^2(1+\nu H)H^2(\nu+2H)}{\theta^2}\le\frac{240\eta\iota^2\nu H^3(\nu+H)}{C^4\eta^2H^4(\nu+H)^2\iota^2}=\O\left(\frac1\eta\right).
  \end{equation*}
  We concentrate $\sum_kZ_k$ by \Cref{lem:freedman}\ref{lem:adapted-upper}. By \Cref{eq:tab-start}, $q^{\pi_k}\le1$, and \Cref{eq:tab-Phi-bonus}, we have
  \begin{align*}
    \E[Z_k\mid\cH_{k-1}]&=4\eta H^2\ind[G_k]\sum_{s,a}(1-u_k(s,a))q^\star(s)\pi_k(a\mid s)\left(\nu+\frac{\nu+2H}{q^{\pi_k}(s,a)}\right)\\&\le\frac{8(\nu+H)}{c_1H}\sum_sq^\star(s)b_k^{(1)}(s)\le\frac{16\nu}{c_1}\sum_sq^\star(s)b_k^{(1)}(s).
  \end{align*}
  Moreover, as at most $H$ estimates start in episode $k$ and $q^{\pi_k}\ge\theta$ for them on $G_k$, we have
  \begin{equation*}
    \iota Z_k\le4\iota\eta H^3\left(\frac\nu\theta+\frac{\nu+2H}{\theta^2}\right)\le\frac{4\iota\eta H^3\nu}{C^2\eta H^2(\nu+H)\iota}+\frac{8\iota\eta H^3(\nu+H)}{C^4\eta^2H^4(\nu+H)^2\iota^2}=\O\left(H+\frac1\eta\right).
  \end{equation*}
  Hence, by \Cref{lem:freedman}\ref{lem:adapted-upper} with $\delta'=\delta/32$ and $c_1\ge56\nu$, except with probability $\delta/32$,
  \begin{equation*}
    \sum_kZ_k\le\frac{16(e-1)\nu}{c_1}\sum_{k,s}q^\star(s)b_k^{(1)}(s)+\O\left(H+\frac1\eta\right)\le\frac12\sum_{k,s}q^\star(s)b_k^{(1)}(s)+\O\left(H+\frac1\eta\right).
  \end{equation*}

  The second sum in \Cref{eq:tab-reg-pathwise} is zero without delays, so we assume $\dbar\ge1$. We apply \Cref{lem:attempts}\ref{item:att-upper} with the weight $\ind[G_k]q^\star(s)\pi_k(a\mid s)(\fd_k+\sigma_k)\le2\dbar$ at every attempt of the estimate $(k,s,a)$, $\epsilon_k(s,a)=1/2$, $\lambda=\theta/(8(1+\nu H)\dbar)$, and $\delta'=\delta/32$, for which $(1+\nu H)\lambda\cdot2\dbar=\theta/4\le p_k(s,a)/2$ on $G_k$. We then apply \Cref{lem:freedman}\ref{lem:adapted-upper} with $\delta'=\delta/32$ to the starts. Except with probability $\delta/16$, they give
  \begin{align*}
    &\quad\sum_{k,s,a}q^\star(s)\pi_k(a\mid s)\left(\fd_k+\sigma_k\right)\what_k(s,a)\\
    &\overset{(a)}\le4\sum_k\left(\fd_k+\sigma_k\right)\ind[G_k]\sum_{s,a}\ind_k(s,a)\frac{q^\star(s)\pi_k(a\mid s)}{q^{\pi_k}(s,a)}+\frac{8(1+\nu H)\dbar\iota}\theta\\
    &\overset{(b)}\le4(e-1)H\sum_k\left(\fd_k+\sigma_k\right)+\frac{8H\dbar\iota}\theta+\frac{8(1+\nu H)\dbar\iota}\theta\overset{(c)}=\O\left(H\mD+\frac{\nu H\dbar\iota}\theta\right),
  \end{align*}
  where (a) used \Cref{eq:att-use}, $\sum_\kappa I_{k,\kappa}(s,a)\le1$, and $p_k\ge q^{\pi_k}/2$, (b) used that the summand of episode $k$ is at most $8H\dbar/\theta$ and has mean at most $4H(\fd_k+\sigma_k)$ given $\cH_{k-1}$ by \Cref{eq:tab-start}, and (c) used \Cref{lem:counting}\ref{item:cnt-delay}. As $\theta=C^2\eta H^2(\nu+H)\iota$, the second sum in \Cref{eq:tab-reg-pathwise} thus contributes
  \begin{equation*}
    \frac{\eta H^2(\nu+2H)}4\cdot\O\left(H\mD+\frac{\nu H\dbar\iota}\theta\right)=\O\left(\eta H^3(\nu+H)\mD+\nu H\dbar\right).
  \end{equation*}
  Plugging the two bounds into \Cref{eq:tab-reg-pathwise} proves the claim, where the failure probability is at most $4\delta/32=\delta/8$.
\end{proof}

\subsection{Bounding \texorpdfstring{\Bonus}{Bonus}}
\label{app:tab-bonus}\label{app:proof-bonus}

$\Bonus$ is the regret on the bonus Q-functions $B_k$. We first prove \Cref{lem:bonus} in a per-episode form, which is the non-dilated version of \citet[Lemma~3.1]{luo2021policy}; see also \citet[Lemma~C.5]{lancewicki2023delay}.

\begin{lemma}[Per-Episode Version of \Cref{lem:bonus}]\label{lem:bonus-per-episode}
  Fix an episode $k$, and suppose that $B_k$ is an optimistic Q-function of $\pi_k$ for $b_k$ (as defined in \Cref{eq:app-bonus-q}). Then
  \begin{equation}\label{eq:bonus-per-episode}
    \sum_{s\in\cS}q^\star(s)\sum_{a\in\cA}\left(\pi_k(a\mid s)-\pi^\star(a\mid s)\right)B_k(s,a)\le\sum_{a\in\cA}\pi_k(a\mid s_0)B_k(s_0,a)-\sum_{s\in\cS}q^\star(s)b_k(s),
  \end{equation}
  where $\sum_a\pi_k(a\mid s_0)B_k(s_0,a)=\sum_sq^{\pi_k}(s)b_k(s)$ if \Cref{eq:app-bonus-q} holds with equality.
\end{lemma}

\begin{proof}
  Let $\bar B_k(s)\triangleq\sum_a\pi_k(a\mid s)B_k(s,a)$, so that $\bar B_k=0$ on $\cS_H$. As every $s'\in\cS_{h+1}$ has $\sum_{s\in\cS_h,a}q^\star(s)\pi^\star(a\mid s)\PP(s'\mid s,a)=q^\star(s')$, \Cref{eq:app-bonus-q} implies
  \begin{align*}
    &\quad \sum_{s\in\cS_h}q^\star(s)\sum_a\left(\pi_k-\pi^\star\right)(a\mid s)B_k(s,a)\\
    &\le\sum_{s\in\cS_h}q^\star(s)\bar B_k(s)-\sum_{s\in\cS_h}q^\star(s)\sum_a\pi^\star(a\mid s)\left(b_k(s)+\sum_{s'}\PP(s'\mid s,a)\bar B_k(s')\right)\\
    &=\sum_{s\in\cS_h}q^\star(s)\bar B_k(s)-\sum_{s'\in\cS_{h+1}}q^\star(s')\bar B_k(s')-\sum_{s\in\cS_h}q^\star(s)b_k(s).
  \end{align*}
  Summing over $h$, the terms of adjacent layers telescope, which, together with $q^\star(s_0)=1$, gives \Cref{eq:bonus-per-episode}. With equality in \Cref{eq:app-bonus-q}, $B_k$ is the Q-function of $\pi_k$ for $b_k$, so $\bar B_k(s_0)=\sum_sq^{\pi_k}(s)b_k(s)$.
\end{proof}

\begin{proof}[Proof of \Cref{lem:bonus}]
  We write $Q_k^{\pi_k}=(Q_k^{\pi_k}-B_k)+B_k$ in \Cref{eq:app-pdl}, and apply \Cref{lem:bonus-per-episode} to the part with $B_k$ in every episode.
\end{proof}

Unlike \citet{jin2020learning} and \citet{luo2021policy}, who bound the gap between $\Ptil_k$ and $\PP$ through the widths of $\cP_k$ (namely, $\overline q_k(s,a)-\underline q_k(s,a)$) and suffer an extra $\sqrt S$, we compare the occupancy measures multiplicatively:

\begin{lemma}\label{lem:comparison}
  If $G_k$ holds, then every $P'\in\cP_k$ satisfies
  \begin{equation*}
    \sum_{s,a}\left[q^{P',\pi_k}(s,a)-e^2\qlow_k(s,a)\right]_+=\O\left(SH^3\beta\sum_{s,a}\frac{q^{\pi_k}(s,a)}{\max\{1,N_k(s,a)\}}\right).
  \end{equation*}
\end{lemma}

\begin{proof}
  We first bound $|P''(s'\mid\widetilde s,b)-\PP(s'\mid\widetilde s,b)|$ for every $P''\in\cP_k$, then split the transitions by whether they are well observed, and finally decompose the visits to $(s,a)$ by their first transition that is not well observed. For $H=1$, every occupancy measure equals $\pi_k(\cdot\mid s_0)$, so we assume $H\ge2$.

  Fix $(\widetilde s,b,s')$, let $n\triangleq\max\{1,N_k(\widetilde s,b)\}$, and abbreviate $p\triangleq\PP(s'\mid\widetilde s,b)$, $\widehat p\triangleq\Phat_n(s'\mid\widetilde s,b)$, and $p''\triangleq P''(s'\mid\widetilde s,b)$ for any $P''\in\cP_k$. If $N_k(\widetilde s,b)\ge1$, then $G_k$ and \Cref{eq:app-conf-set} give
  \begin{equation*}
    \widehat p\le p+2\sqrt{\frac{\widehat p}2\cdot\frac{2\beta}n}+\frac{5\beta}n\le p+\frac{\widehat p}2+\frac{2\beta}n+\frac{5\beta}n,
  \end{equation*}
  and rearranging yields $\widehat p\le2p+14\beta/n$. As both $P''$ and $\PP$ satisfy the constraints in \Cref{eq:app-conf-set}, we then have
  \begin{equation*}
    |p''-p|\le|p''-\widehat p|+|\widehat p-p|\le4\sqrt{\frac{\widehat p\beta}n}+\frac{10\beta}n\le4\sqrt{\frac{2p\beta}n}+\frac{4\sqrt{14}\beta}n+\frac{10\beta}n\le6\sqrt{\frac{p\beta}n}+\frac{25\beta}n,
  \end{equation*}
  which also holds if $N_k(\widetilde s,b)=0$, as then $n=1$ and $\beta\ge1$.

  We call $(\widetilde s,b,s')$ \emph{well observed} if
  \begin{equation}\label{eq:large}
    \PP(s'\mid\widetilde s,b)\max\{1,N_k(\widetilde s,b)\}\ge144H^2\beta.
  \end{equation}
  If \Cref{eq:large} holds, that is, $pn\ge144H^2\beta$, then $\sqrt{p\beta/n}\le p/(12H)$ and $\beta/n\le p/(144H^2)$, so every $P''\in\cP_k$ satisfies
  \begin{equation*}
    |p''-p|\le\frac{6p}{12H}+\frac{25p}{144H^2}\le\frac pH.
  \end{equation*}
  Otherwise, we have $p<144H^2\beta/n$, so every $P''\in\cP_k$ satisfies
  \begin{equation*}
    p''\le p+6\sqrt{\frac{p\beta}n}+\frac{25\beta}n\le\frac{(144H^2+72H+25)\beta}n\le\frac{241H^2\beta}n.
  \end{equation*}

  Now fix $(s,a)$, and let $\mathcal W$ be the set of well-observed transitions. Let $\mathcal E(s,a)$ be the event that a trajectory reaches $(s,a)$ after some transition outside $\mathcal W$, and $\mathcal E'(\widetilde s,b)$ the event that it reaches $(\widetilde s,b)$ using only transitions in $\mathcal W$. On a partial trajectory that ends at $(s,a)$ and only uses transitions in $\mathcal W$, the probabilities under $P'$ and under any $P''\in\cP_k$ have a ratio of at most $((1+1/H)/(1-1/H))^{H-1}\le e^2$. Summing over these partial trajectories and minimizing over $P''$, we get
  \begin{equation*}
    \left[q^{P',\pi_k}(s,a)-e^2\qlow_k(s,a)\right]_+\le\Pr_{P'}\left(\mathcal E(s,a)\right).
  \end{equation*}
  Moreover, as $P'$ is within a factor $1+1/H$ of $\PP$ on $\mathcal W$, we have
  \begin{equation*}
    \Pr_{P'}\left(\mathcal E'(\widetilde s,b)\right)\le\left(1+\frac1H\right)^{H-1}q^{\pi_k}(\widetilde s,b)\le e\,q^{\pi_k}(\widetilde s,b).
  \end{equation*}
  A trajectory visits at most $H$ pairs, and on $\mathcal E(s,a)$, it takes a first transition $(\widetilde s,b,s')\notin\mathcal W$ on $\mathcal E'(\widetilde s,b)$, so
  \begin{align*}
    \sum_{s,a}\left[q^{P',\pi_k}(s,a)-e^2\qlow_k(s,a)\right]_+&\le H\sum_{(\widetilde s,b,s')\notin\mathcal W}\Pr_{P'}\left(\mathcal E'(\widetilde s,b)\right)P'(s'\mid\widetilde s,b)\\
    &\le eH\sum_{\widetilde s,b}q^{\pi_k}(\widetilde s,b)\frac{241SH^2\beta}{\max\{1,N_k(\widetilde s,b)\}},
  \end{align*}
  which finishes the proof.
\end{proof}

The next standard lemma bounds the inverse visit counts in \Cref{lem:comparison}, summed over the episodes.

\begin{lemma}\label{lem:harmonic}
  With probability at least $1-\delta/32$, we have
  \begin{equation*}
    \sum_{k}\sum_{s,a}\frac{q^{\pi_k}(s,a)}{\max\{1,N_k(s,a)\}}\le2SA(2+\log K)+2H\iota,
  \end{equation*}
  where $k$ ranges over the non-skipped episodes.
\end{lemma}

\begin{proof}
  Let $X_k\triangleq\sum_{s,a}\ind[\text{episode $k$ visits }(s,a)]/\max\{1,N_k(s,a)\}\in[0,H]$, so that $\E[X_k\mid\cH_{k-1}]$ is the summand of episode $k$ in the claim, and $\E[X_k^2\mid\cH_{k-1}]\le H\,\E[X_k\mid\cH_{k-1}]$. The $n$-th visit to $(s,a)$ among the non-skipped episodes, in the processing order, has $N_k(s,a)\ge n-1$. Hence, by \Cref{lem:freedman}\ref{lem:adapted-lower} with $\lambda=1/H$ and $\delta'=\delta/32$, except with probability $\delta/32$, we have
  \begin{equation*}
    \frac12\sum_k\E[X_k\mid\cH_{k-1}]\le\sum_kX_k+H\iota\le SA\left(1+\sum_{n=1}^K\frac1n\right)+H\iota\le SA(2+\log K)+H\iota,
  \end{equation*}
  which finishes the proof.
\end{proof}

\begin{lemma}[Bounding \Bonus]\label{prop:bonus}
  If the good event $G$ and the bound of \Cref{lem:harmonic} hold, then
  \begin{equation*}
    \sum_{k=1}^K\sum_{s\in\cS}q^\star(s)b_k(s)+\Bonus=\O\left(\eta H^3(\nu+H)\iota SAK+H^4S^2A\beta\iota\right).
  \end{equation*}
\end{lemma}

\begin{proof}
  On $G$, every non-skipped $k$ has $\PP\in\cP_k$, so the maximum in \Cref{eq:app-tab-B} is at least its value at $\PP$, that is, $B_k$ is an optimistic Q-function of $\pi_k$ for $b_k$. Moreover, as the constraints of $\cP_k$ are separate for every pair, $\Ptil_k$ attains the maximum in \Cref{eq:app-tab-B} at every pair simultaneously, so $B_k$ is the Q-function of $\pi_k$ for $b_k$ under $\Ptil_k$. Hence, we have
  \begin{align*}
    &\quad\sum_sq^\star(s)b_k(s)+\sum_{s,a}q^\star(s)\left(\pi_k-\pi^\star\right)(a\mid s)B_k(s,a)\\
    &\overset{(a)}\le\sum_a\pi_k(a\mid s_0)B_k(s_0,a)=\sum_sq^{\Ptil_k,\pi_k}(s)b_k(s)\\
    &\overset{(b)}\le e^2SA\left(c_1\eta H^3+H\theta\right)+H\sum_{s,a}\left[q^{\Ptil_k,\pi_k}(s,a)-e^2\qlow_k(s,a)\right]_+\\
    &\overset{(c)}=\O\left(\eta H^3(\nu+H)\iota SA+SH^4\beta\sum_{s,a}\frac{q^{\pi_k}(s,a)}{\max\{1,N_k(s,a)\}}\right),
  \end{align*}
  where (a) used \Cref{lem:bonus-per-episode}, (b) used $q^{\Ptil_k,\pi_k}\le e^2\qlow_k+[q^{\Ptil_k,\pi_k}-e^2\qlow_k]_+$, that $\qlow_k(s,a)$ cancels the denominator in \Cref{eq:app-tab-bonus}, and that both terms in \Cref{eq:app-tab-bonus} are at most $H$, and (c) used \Cref{eq:tab-conditions}, the definition of $\theta$, and \Cref{lem:comparison}. Summing over $k$ and applying \Cref{lem:harmonic}, with $H\le S$ and $\log K\le\iota$, we get
  \begin{align*}
    \sum_{k=1}^K\sum_sq^\star(s)b_k(s)+\Bonus&=\O\left(\eta H^3(\nu+H)\iota SAK+SH^4\beta\left(SA(2+\log K)+H\iota\right)\right)\\
    &=\O\left(\eta H^3(\nu+H)\iota SAK+H^4S^2A\beta\iota\right),
  \end{align*}
  which finishes the proof.
\end{proof}

\section{Proofs for (Delayed) Linear AMDPs}
\label{app:linear}

In this appendix, we prove \Cref{thm:delayed-linear}, which contains \Cref{thm:linear} as the special case without delays. The delayed algorithm, \Cref{alg:delayed-linear} below, processes the episodes in the same way as \Cref{alg:delayed-tabular} (\Cref{app:delay-alg}), so we reuse the processing order, the numbers $\mathfrak n_t$, $N(t_1,t_2)$, and $\sigma_k$, and \Cref{lem:counting}, whose proof only uses that every non-skipped episode $k$ has $\fd_k\le\dbar$.

We again restate several notations and formulae.
Recall that, when processing a non-skipped episode $k$, we run \Cref{alg:regions} for every layer $h$ on the layer-$h$ features of the last $\sqrt K$ episodes processed before $k$. With $\rho\triangleq2d\log(1+1/\lambda)$, \Cref{alg:regions} repeatedly fits an ellipsoid $\{\phi:\lVert\phi\rVert_{\Lambda_j^{-1}}^2\le\rho\}$ to the remaining features and takes it as the next region $\cC_j$ (minus the previous regions), until at most $\sqrt K\theta$ features remain. For every region $\cC_j$, it returns the fraction $\phat_j$ of the $\sqrt K$ features that are contained in $\cC_j$, and the empirical covariance matrix $\Sigmahat_j$ of these features, with its small eigenvalues truncated to zero.
For an estimate started at layer $h$ in a region $\cC_j$ when processing episode $k$, its $n$-th attempt, with an episode $\kappa$, updates the product $Z_0=I$ as (where recall that $\Sigmahat_j^{-1/2}\triangleq (\Sigmahat_j^\dagger)^{1/2}$ and $\Sigmahat_j^\dagger$ is the Moore--Penrose pseudo-inverse of $\Sigmahat_j$)
\begin{equation}\label{eq:app-matrix-update}
  Z_n=\left(I-\frac{\ind[\text{episode $\kappa$ is accepted and }\phi_{\kappa,h}\in\cC_j]}{4\rho}\Sigmahat_j^{-1/2}\phi_{\kappa,h}\phi_{\kappa,h}^\top\Sigmahat_j^{-1/2}\right)Z_{n-1}.
\end{equation}
The bonus $b_k$ is defined in \Cref{eq:app-lin-bonus} of \Cref{alg:delayed-linear}, and the bonus Q-function $B_k$ is $0$ on $\cS_H$ and satisfies
\begin{equation}\label{eq:app-linear-B}
  B_k(s,a)\triangleq b_k(s)+(1-u_k(s,a))\sum_{s'}\PP(s'\mid s,a)\sum_{a'}\pi_k(a'\mid s')B_k(s',a')+4H(H-h-1)\,u_k(s,a),
\end{equation}
for all $s\in \cS_h$, $a\in \cA$, and $0\le h<H$. The cumulative loss-minus-bonus of episode $k$ from layer $h$ is
\begin{equation}\label{eq:app-response}
  Y_{k,h}\triangleq L_{k,h}-\sum_{h'=h+1}^{H-1}\left(\prod_{h''=h+1}^{h'-1}\bigl(1-u_k(s_{k,h''},a_{k,h''})\bigr)\right)\Bigl(b_k(s_{k,h'})+4H(H-h'-1)\,u_k(s_{k,h'},a_{k,h'})\Bigr).
\end{equation}

\Cref{alg:delayed-linear} processes episode $\kappa$ at the end of episode $t_\kappa=\kappa+\min\{\fd_\kappa,\dbar\}$, and skips it unless \textit{i)} $\fd_\kappa\le\dbar$, \textit{ii)} at least $\sqrt K$ episodes are processed before $\kappa$, and \textit{iii)} $\kappa+\fd_\kappa\le K-\sqrt K-\dbar$. Without delays, \textit{ii)} and \textit{iii)} read $\sqrt K<\kappa\le K-\sqrt K$, and \Cref{alg:delayed-linear} is exactly \Cref{alg:linear}. As in \Cref{alg:delayed-tabular}, a skipped episode starts no estimate, but still makes attempts for the unfinished estimates when it is processed.

\begin{algorithm}[htb]
\caption{Policy Optimization with MGRR for Linear AMDPs with Delayed Feedback}
\label{alg:delayed-linear}
\begin{algorithmic}[1]
\Require Cap $\dbar$; parameters $\lambda,m_0,\nu,\eta,\theta$ (specified in \Cref{eq:lin-params}).
\For{$t=1,2,\ldots,K$}
  \State For every visited state $s$, let $\pi_t(a\mid s)\propto\exp\bigl(-\eta\sum_{t'<t}g_{t'}(s,a)\bigr)$.
  \State Execute $\pi_t$, and observe the trajectory $(s_{t,h},a_{t,h})_{h=0}^{H-1}$; let $\phi_{t,h}\triangleq\phi(s_{t,h},a_{t,h})$; its losses $(\ell_t(s_{t,h},a_{t,h}))_{h=0}^{H-1}$ arrive at the end of episode $t+\fd_t$.
  \For{each episode $\kappa$ with $\kappa+\min\{\fd_\kappa,\dbar\}=t$, in increasing order of $\kappa$}
  \If{$\kappa$ satisfies \textit{i)}--\textit{iii)} (i.e., $\kappa$ is not skipped)}
    \State For every $h$, construct regions $(\cC_j,\phat_j,\Sigmahat_j)_j$ via \Cref{alg:regions} with the features $\phi_{k,h}$ of the last $\sqrt K$ episodes $k$ processed before $\kappa$.
    \State Let $u_\kappa(s,a)\gets0$ if $\phi(s,a)$ lies in a region and its layer has fewer than $\nu$ unfinished estimates, and $u_\kappa(s,a)\gets1$ otherwise; define the bonus as
    \algeq[eq:app-lin-bonus]{$\displaystyle b_\kappa(s)\triangleq H\theta\sum_{a:\,u_\kappa(s,a)=0}\frac{\pi_\kappa(a\mid s)}{\phat_\kappa(s,a)}+4H\sum_{a:\,u_\kappa(s,a)=1}\pi_\kappa(a\mid s),\;\forall s\in\cS.$}
    \State Initialize $\Qhat_\kappa(s,a)\gets-b_\kappa(s)-4H(H-h-1)u_\kappa(s,a)-K^{-3}$ for all $s\in\cS_h$ and $a\in\cA$.
  \EndIf
  \For{each unfinished estimate, started in an episode $k$ processed before $\kappa$ at layer $h$ in region $\cC_j$}
    \State Let $n$ be the number of attempts this estimate has made, including this one.
    \State Accept episode $\kappa$ with probability $\min\{1,\alpha_{k,\kappa}\}$ (\Cref{eq:app-alpha}), and update $Z_n$ via \Cref{eq:app-matrix-update}.
    \State For every $(s,a)\in\cS_h\times\cA$ with $\phi(s,a)\in\cC_j$, add to $\Qhat_k(s,a)$ the piece
    \algeq[eq:app-increment]{$\displaystyle v_n(s,a)\triangleq\frac{\phi(s,a)^\top\Sigmahat_j^{-1/2}Z_n\Sigmahat_j^{-1/2}\phi_{k,h}Y_{k,h}}{16d\log(1+1/\lambda)}.$}
    \State Finish the estimate if $n=M_j-1$.
  \EndFor
  \If{$\kappa$ satisfies \textit{i)}--\textit{iii)} (i.e., $\kappa$ is not skipped)}
    \For{each $h$ such that $u_\kappa(s_{\kappa,h},a_{\kappa,h})=0$}
      \State Let $\cC_j$ be the region of layer $h$ that contains $\phi_{\kappa,h}$.
      \State Compute the cumulative loss-minus-bonus $Y_{\kappa,h}$ via \Cref{eq:app-response}.
      \State Start an MGRR estimate in $\cC_j$ with $Z_0=I$ and $M_j=\lceil m_0/\phat_j\rceil$.
      \State Add $v_0(s,a)$, that is, \Cref{eq:app-increment} with $n=0$ and $k=\kappa$, to $\Qhat_\kappa(s,a)$ for every $(s,a)\in\cS_h\times\cA$ with $\phi(s,a)\in\cC_j$.
    \EndFor
  \EndIf
  \EndFor
  \State Let $g_t(s,a)$ be the total change of $\sum_k(\Qhat_k-\eta\Qhat_k^2)(s,a)$ during this episode.
\EndFor
\end{algorithmic}
\end{algorithm}

We use the $\sigma$-algebras $\cH_{\kappa-1}$ of \Cref{app:delay-alg}, now generated by the trajectories of the episodes processed before $\kappa$ and the acceptance tests used when processing them. \Cref{lem:processing} and its proof apply verbatim to \Cref{alg:delayed-linear}, with the regions and $u_\kappa$ in place of $\cP_\kappa$ and $\qlow_\kappa$ in \Cref{lem:processing}\ref{item:proc-measurable}.

We also use the following notations. Let $\mathcal K$ be the set of non-skipped episodes, the only ones that start estimates, which is $\{k:\sqrt K<k\le K-\sqrt K\}$ without delays; sums over $k$ range over $\mathcal K$. When processing episode $k$, the regions of layer $h$ are $\cC_{k,h,j}$, with empirical probabilities $\phat_{k,h,j}$ and matrices $\Sigmahat_{k,h,j}$, and we drop $k$ and $h$ when they are clear from the context. For a pair $(s,a)\in\cS_h\times\cA$ whose feature lies in a region $\cC_{k,h,j}$, we write $\phat_k(s,a)\triangleq\phat_{k,h,j}$. We write $\Qhat_k$ for the final value of the running estimate and $\Qtil_k\triangleq\Qhat_k-\eta\Qhat_k^2$, so that the received loss $g_t(s,a)$ collects the changes of $\Qtil_k(s,a)$ made when processing the episodes $\kappa$ with $t_\kappa=t$. As in \Cref{app:tab-key}, $\E_\pi$ and $\Pr_\pi$ refer to a trajectory of a policy $\pi$ under $\PP$ that is independent of everything else, whose layer-$h$ feature we write as $\phi_h\triangleq\phi(s_h,a_h)$, so that $\Sigma_{h,j}^\pi=\E_\pi[\ind[\phi_h\in\cC_j]\phi_h\phi_h^\top]$.

\subsection{Proof of Main Theorem}
\label{app:lin-main}

This subsection proves the main theorem, using lemmas proved in the remaining subsections. Let $C_1$, $C$, and $C_0$ be universal constants, each large enough given the previous ones. Let $\iota\triangleq\log(16KdHA/\delta)$, $J_0\triangleq\lceil\log_2(2\sqrt K)\rceil$, which bounds the number of regions (\Cref{lem:lin-regions-det}), and $d_{\mathrm{VC}}\triangleq\lceil2(d+1)(d+2)(J_0+1)\log_2(J_0+2)\rceil$, which bounds the VC dimension of the regions (\Cref{lem:lin-vc}). With $\dbar$ the cap on the delays, we pick the following parameters:
\begin{equation}\label{eq:lin-params}
  \begin{aligned}
  &\lambda=(KdH)^{-C},\; m_0=\lceil C_0\rho\iota\rceil=\Thetatil(d),\;\nu=\lceil C_0m_0\iota\rceil=\Thetatil(d),\\
  &\eta=\frac1{200H^3(\sqrt K+2\dbar)(\nu+4)}=\Thetatil\left(\frac1{dH^3(\sqrt K+\dbar)}\right),\;\theta=\frac{C_0}{\sqrt K}\left(\rho(d_{\mathrm{VC}}\iota+\iota^2)+m_0^2\right)=\Thetatil\left(\frac{d^3}{\sqrt K}\right),
  \end{aligned}
\end{equation}
where the orders follow from $\rho=\Thetatil(d)$ and $d_{\mathrm{VC}}=\Thetatil(d^2)$. Here, $\lambda$ is the regularization of \Cref{alg:regions}, $m_0$ sets the truncation levels $M_j=\lceil m_0/\phat_j\rceil$, $\nu$ caps the number of unfinished estimates of a layer, and $\theta$ is the threshold of \Cref{alg:regions}. If $\theta\le1/2$, the proofs only use the below bounds (which follow trivially from \Cref{eq:lin-params}):
\begin{equation}\label{eq:lin-conditions}
  \begin{aligned}
  &5H^3(\sqrt K+2\dbar)\,\eta(\nu+4)\le\frac1{40},\;\theta\ge\frac{4m_0}{\sqrt K},\; C_1\eta(\nu+4)H^3m_0^2\le\theta,\;\sqrt K\ge200\iota,\\
  &\sqrt K\theta\ge C_1\rho\left(d_{\mathrm{VC}}\iota+\iota^2\right),\; C_1H^2\sqrt d\left(\rho^{3/2}\sqrt\lambda+\sqrt\rho\,e^{-m_0/(32\rho)}\right)\le K^{-3}.
  \end{aligned}
\end{equation}
Indeed, the first one holds with equality. The second and third ones follow from $\theta\ge C_0m_0^2/\sqrt K$ and $\eta(\nu+4)H^3\le1/(200\sqrt K)$, as $C_0$ is large given $C_1$. The fourth and fifth ones follow from $\theta\le1/2$ and $\sqrt K\theta\ge C_0\rho(d_{\mathrm{VC}}\iota+\iota^2)$, which give $\sqrt K\ge2\sqrt K\theta\ge2C_0\rho\iota^2\ge200\iota$. For the last one, $\theta\le1/2$ implies $KdH\ge2$, so $\rho=2d\log(1+(KdH)^C)$ only grows linearly in $C$, while $\sqrt\lambda=(KdH)^{-C/2}$ decays exponentially in $C$; moreover, $m_0\ge C_0\rho\iota$ gives $e^{-m_0/(32\rho)}\le e^{-C_0\iota/32}$. Hence, the last one holds as $C$ is large given $C_1$ and $C_0$ is large given $C$.

As in \Cref{app:tab-main}, we write
\begin{equation}\label{eq:lin-bonus-est}
  b_k^{(1)}(s)\triangleq H\theta\sum_{a:\,u_k(s,a)=0}\frac{\pi_k(a\mid s)}{\phat_k(s,a)}\le b_k(s)
\end{equation}
for the part of the bonus in \Cref{eq:app-lin-bonus} that covers the estimated pairs. Half of $b_k$ covers the second moments of the estimates and the unestimated pairs in $\BiasOne$ (\Cref{lem:lin-bias}), and half of $b_k^{(1)}$ covers the delay cost of the pieces in $\RegTerm$ (\Cref{lem:lin-reg}).

\begin{proof}[Proof of \Cref{thm:linear,thm:delayed-linear}]
  If $\mD\ge K^2$, then $\mR_K\le HK\le H\sqrt\mD$. If $\theta>1/2$, then $\sqrt K<2\sqrt K\theta=\Otil(d^3)$ and $\mR_K\le HK=\Otil(d^3H\sqrt K)$, and if $K<2\sqrt K$, then $\mR_K\le3H$. Hence, we assume $\mD<K^2$, $\theta\le1/2$, and $K\ge2\sqrt K$. Let $\mathcal K_{\mathrm{skip}}$ be the set of skipped episodes. At most $\mD/(\dbar+1)$ episodes violate \textit{i)}, at most $\sqrt K$ episodes violate \textit{ii)}, and every episode that satisfies \textit{i)} but violates \textit{iii)} has $k>K-\sqrt K-2\dbar$, so
  \begin{equation*}
    |\mathcal K_{\mathrm{skip}}|\le\frac\mD{\dbar+1}+2\sqrt K+2\dbar.
  \end{equation*}
  For every $k\in\mathcal K$, we write $Q_k^{\pi_k}=(Q_k^{\pi_k}-B_k-\Qtil_k)+\Qtil_k+B_k$ in \Cref{eq:app-pdl}, and we bound every skipped episode by $H$, which gives
  \begin{align}
    \mR_K&\le\underbrace{\sum_{k\in\mathcal K}\sum_{h,s}q^\star(s)\sum_a\pi_k(a\mid s)\left(Q_k^{\pi_k}-B_k-\Qtil_k\right)(s,a)}_{\text{\BiasOne}}+\underbrace{\sum_{k\in\mathcal K}\sum_{h,s}q^\star(s)\sum_a\pi^\star(a\mid s)\left(\Qtil_k-Q_k^{\pi_k}+B_k\right)(s,a)}_{\text{\BiasTwo}}\notag\\
    &\quad+\underbrace{\sum_{k\in\mathcal K}\sum_{h,s}q^\star(s)\sum_a\left(\pi_k-\pi^\star\right)(a\mid s)\,\Qtil_k(s,a)}_{\text{\RegTerm}}+\underbrace{\sum_{k\in\mathcal K}\sum_{h,s}q^\star(s)\sum_a\left(\pi_k-\pi^\star\right)(a\mid s)\,B_k(s,a)}_{\text{\Bonus}}+H|\mathcal K_{\mathrm{skip}}|.\label{eq:lin-decomp}
  \end{align}
  By a union bound, the events of \Cref{lem:lin-good,lem:lin-capacity,lem:lin-bias,lem:lin-reg} hold simultaneously with probability at least $1-\delta/8-\delta/8-\delta/4-\delta/4\ge1-\delta$. In the rest of the proof, we assume that they hold. Then \Cref{lem:lin-bias} gives $\BiasTwo\le H\nu\iota/\eta$, and \Cref{lem:lin-bias,lem:lin-reg} and $b_k^{(1)}\le b_k$ give
  \begin{align*}
    \BiasOne+\RegTerm&\le\sum_k\sum_{h,s}q^\star(s)b_k(s)+\frac{H\nu\iota+H\log A}\eta\\
    &\quad+\O\left(\eta H^5\nu^2(K+\mD)+\eta\nu H^5m_0\mD+H^2\iota(\sqrt K+\dbar)\right).
  \end{align*}
  Moreover, \Cref{lem:lin-bonus} gives $\Bonus+\sum_k\sum_{h,s}q^\star(s)b_k(s)=\O(H^3\theta K\log K)$. Plugging these bounds into \Cref{eq:lin-decomp}, and using $\log A\le\iota$, we get
  \begin{equation*}
    \mR_K=\O\left(\frac{H\nu\iota}\eta+\eta H^5\nu^2(K+\mD)+\eta\nu H^5m_0\mD+H^3\theta K\log K+H^2\iota(\sqrt K+\dbar)+\frac{H\mD}{\dbar+1}\right).
  \end{equation*}
  By the orders in \Cref{eq:lin-params}, we have $H\nu\iota/\eta=\Otil(d^2H^4(\sqrt K+\dbar))$, $H^3\theta K\log K=\Otil(d^3H^3\sqrt K)$, and
  \begin{equation*}
    \eta H^5\nu^2(K+\mD)+\eta\nu H^5m_0\mD=\Otil\left(\frac{dH^2(K+\mD)}{\sqrt K+\dbar}\right)=\Otil\left(dH^2\sqrt K+\frac{dH^2\mD}{\max\{1,\dbar\}}\right).
  \end{equation*}
  Hence, without delays, $\mR_K=\Otil(d^2H^4\sqrt K+d^3H^3\sqrt K)=\Otil(d^3H^4\sqrt K)$. With delays, $\dbar=\lceil\sqrt{\mD/(dH^2)}\rceil\ge1$ gives $d^2H^4\dbar\le d^{3/2}H^3\sqrt\mD+d^2H^4$ and $dH^2\mD/\dbar+H\mD/(\dbar+1)\le2d^{3/2}H^3\sqrt\mD$, that is, $\mR_K=\Otil(d^3H^4\sqrt K+d^{3/2}H^3\sqrt\mD)$. Finally, \Cref{alg:delayed-linear} is computationally efficient, as $\pi_k(\cdot\mid s)$ takes $\poly(K,d,H,A)$ time to compute by replaying the updates at $s$.
\end{proof}

\subsection{Properties of \Cref{alg:regions}}
\label{app:lin-regions}

The first lemma shows that \Cref{alg:regions} splits the features into few regions, each of which contains a lot of them, and that every feature in a region has a bounded norm, which keeps the factors in \Cref{eq:app-matrix-update} between $I/2$ and $I$. Recall that, in \Cref{alg:regions}, $m_j$ is the number of features at the start of the $j$-th outer iteration, and $R_j$ is the set of the features that it keeps to form region $\cC_j$.

\begin{lemma}\label{lem:lin-regions-det}
  If $\lVert\phi_i\rVert_2\le1$ for all $i$ and $\theta\ge1/\sqrt K$, then \Cref{alg:regions} returns $J\le J_0$ regions, which contain exactly the features in $R_1,\ldots,R_J$ and leave at most $\sqrt K\theta$ features outside. Moreover, $\sqrt K\phat_j=|R_j|>m_j/2$, $\phat_1\ge\cdots\ge\phat_J>\theta/2$, $\Sigmahat_j\preceq I$, and every $\phi\in\cC_j$ satisfies $\phi^\top\Sigmahat_j^\dagger\phi\le2\rho$ and $\lVert(I-\Sigmahat_j\Sigmahat_j^\dagger)\phi\rVert_2\le2\sqrt{\rho\lambda}$.
\end{lemma}

\begin{proof}
  We follow the elliptical potential argument of \citet[Lemma~11]{abbasi2011improved}. Let $\overline R_j$ be the set of features removed in the $j$-th outer iteration. As $\lambda I\preceq\Lambda_j\preceq(1+\lambda)I$ throughout this iteration, the matrix determinant lemma implies
  \begin{equation*}
    |\overline R_j|\cdot\frac\rho{m_j}\le\sum_{i\in\overline R_j}-\log\left(1-\frac{\phi_i^\top\Lambda_j^{-1}\phi_i}{m_j}\right)\le\log\det\left((1+\lambda)I\right)-\log\det(\lambda I)=d\log\left(1+\frac1\lambda\right)=\frac\rho2,
  \end{equation*}
  where $\Lambda_j$ inside the sum is the matrix right before the removal of $\phi_i$, for which $\phi_i^\top\Lambda_j^{-1}\phi_i>\rho$. Hence, fewer than $m_j/2$ features are removed (the first inequality is strict unless $\overline R_j=\varnothing$). As $\phi^\top\Lambda_j^{-1}\phi$ only increases during the removals, the features in $\cC_j$ are exactly those in $R_j$. Hence, $\sqrt K\phat_{j+1}\le m_{j+1}=m_j-|R_j|<|R_j|=\sqrt K\phat_j$ and $m_{j+1}<m_j/2$. As $m_1=\sqrt K$ and the loop requires $m_J>\sqrt K\theta\ge1$, this gives
  \begin{equation*}
    J\le J_0\quad \text{and}\quad\phat_J>\frac{m_J}{2\sqrt K}>\frac \theta2.
  \end{equation*}

  Next, let $(\mu_i,e_i)$ be the eigenpairs of the empirical covariance matrix $\overline\Sigma_j$ of the features in $R_j$, and let $\lambda_j\triangleq\lambda m_j/|R_j|\in[\lambda,2\lambda)$, so that $\Lambda_j=(|R_j|/m_j)(\overline\Sigma_j+\lambda_jI)$. By construction, $\Sigmahat_j$ keeps exactly the eigenvalues $\mu_i\ge\lambda_j$, for which $\mu_i\ge(\mu_i+\lambda_j)/2$, while the others have $\mu_i+\lambda_j<4\lambda$. Therefore, every $\phi\in\cC_j$ satisfies
  \begin{equation*}
    \phi^\top\Sigmahat_j^\dagger\phi=\sum_{i:\,\mu_i\ge\lambda_j}\frac{(e_i^\top\phi)^2}{\mu_i}\le2\sum_i\frac{(e_i^\top\phi)^2}{\mu_i+\lambda_j}=\frac{2|R_j|}{m_j}\phi^\top\Lambda_j^{-1}\phi\le2\rho,
  \end{equation*}
  and, in the same way,
  \begin{equation*}
    \left\lVert(I-\Sigmahat_j\Sigmahat_j^\dagger)\phi\right\rVert_2^2=\sum_{i:\,\mu_i<\lambda_j}(e_i^\top\phi)^2\le4\lambda\sum_i\frac{(e_i^\top\phi)^2}{\mu_i+\lambda_j}=\frac{4\lambda|R_j|}{m_j}\phi^\top\Lambda_j^{-1}\phi\le4\lambda\rho,
  \end{equation*}
  which finishes the proof.
\end{proof}

When processing episode $k$, the regions and their matrices $\Sigmahat_j$ are computed from the features of the last $\sqrt K$ episodes processed before $k$, but MGRR wants them to cover the current policy $\pi_k$. Fortunately, by \Cref{lem:lin-stability} in \Cref{app:lin-key}, the policies of these episodes stay close to $\pi_k$, so their features are almost i.i.d.\ samples from the policy of the first of them. Hence, uniform convergence over the class of regions transfers the empirical quantities of every region to $\pi_k$. We formalize this by the following definition and lemma.

\begin{definition}\label{def:lin-good}
  For $k\in\mathcal K$ and a layer $h$, we call $(k,h)$ \emph{good} if $\Pr_{\pi_k}(\phi_h\notin\bigcup_j\cC_{k,h,j})\le4\theta$ and every region $\cC_j=\cC_{k,h,j}$ satisfies $\Pr_{\pi_k}(\phi_h\in\cC_j)\le4\phat_j$ and $(\phat_j/4)\Sigmahat_j\preceq\Sigma_{h,j}^{\pi_k}\preceq4\phat_j\Sigmahat_j$ on $\operatorname{range}(\Sigmahat_j)$.
\end{definition}

\begin{lemma}\label{lem:lin-good}
  For every $k\in\mathcal K$ and layer $h$, the event that $(k,h)$ is good (\Cref{def:lin-good}) is $\cH_{k-1}$-measurable. Furthermore, with probability at least $1-\delta/8$, every $(k,h)$ is good.
\end{lemma}

\begin{proof}
  The regions and $\pi_k$ are $\cH_{k-1}$-measurable by \Cref{lem:processing}, so the event that $(k,h)$ is good is $\cH_{k-1}$-measurable. Fix $k\in\mathcal K$ and $h$, and let $\kappa_1,\ldots,\kappa_{\sqrt K}$ be as in \Cref{lem:lin-window} (stated and proved after this lemma), whose layer-$h$ features define the regions. The additive term of \Cref{lem:lin-window} is small, as
  \begin{equation*}
    4\rho\bar\epsilon=\frac{36\rho}{\sqrt K}\left(d_{\mathrm{VC}}\log(2e\sqrt K)+\log\frac{512KH}\delta\right)\le\frac{180\rho}{\sqrt K}\left(d_{\mathrm{VC}}\iota+\iota\right)\le\frac{\min\{\phat_j,\theta\}}{100},
  \end{equation*}
  where the first step used $\log(2e\sqrt K)\le\iota$ and $\log(512KH/\delta)\le5\iota$, and the second step used \Cref{eq:lin-conditions} and $\phat_j>\theta/2$. Fix a region $\cC_j$ and $x\in\operatorname{range}(\Sigmahat_j)$ with $x^\top\Sigmahat_jx=1$, and let $f(\phi)\triangleq\ind[\phi\in\cC_j](x^\top\phi)^2/(2\rho)$, whose level sets are in the class of \Cref{lem:lin-vc} (stated and proved after this lemma). As $(x^\top\phi)^2\le(x^\top\Sigmahat_jx)(\phi^\top\Sigmahat_j^\dagger\phi)\le2\rho$ for every $\phi\in\cC_j$ by the Cauchy--Schwarz inequality and \Cref{lem:lin-regions-det}, $f$ takes values in $[0,1]$.

  By \Cref{lem:lin-regions-det}, the empirical frequency of $\cC_j$ is $\phat_j$, and that of the features outside all regions is at most $\theta$, so \Cref{eq:lin-window} implies $\Pr_{\pi_{\kappa_1}}(\phi_h\in\cC_j)\le3\phat_j+\bar\epsilon\le3.01\phat_j$ and $\Pr_{\pi_{\kappa_1}}(\phi_h\notin\bigcup_j\cC_j)\le3\theta+\bar\epsilon\le3.01\theta$. Moreover, as $x$ is orthogonal to the eigenvectors of the eigenvalues that $\Sigmahat_j$ truncates to zero, the empirical mean of $2\rho f$ is $\phat_jx^\top\Sigmahat_jx=\phat_j$, so applying \Cref{eq:lin-window} to $f$ gives
  \begin{equation*}
    0.33\phat_j\le\frac{\phat_j-2\rho\bar\epsilon}3\le\E_{\pi_{\kappa_1}}\left[\ind[\phi_h\in\cC_j](x^\top\phi_h)^2\right]\le3\phat_j+2\rho\bar\epsilon\le3.01\phat_j.
  \end{equation*}
  As $\sqrt K$ episodes are processed from $\kappa_1$ to $k$, we have $N(\kappa_1,k)\le\sqrt K+2\dbar$ by \Cref{lem:counting}\ref{item:cnt-between}, so, by \Cref{lem:lin-stability}, the laws of the partial trajectories of $\pi_{\kappa_1}$ and $\pi_k$ are within a factor of $e^{\pm1/40}$ of each other. Hence, $(k,h)$ is good by $3.01e^{1/40}\le4$ and $0.33e^{-1/40}\ge1/4$, unless the event of \Cref{lem:lin-window} fails. Finally, $\sqrt K\ge200\iota$ by \Cref{eq:lin-conditions}, so $2e^{-\sqrt K/100}\le2e^{-2\iota}\le2\delta/(64KH)$, and a union bound over $k$ and $h$ finishes the proof.
\end{proof}

The next lemma shows that all the sets that we need belong to a class that is simple enough for uniform convergence.

\begin{lemma}\label{lem:lin-vc}
  The class of intersections of at most $J_0+1$ sets of the form $\{\phi:\phi^\top W\phi\le c\}$ or $\{\phi:\phi^\top W\phi>c\}$, where $W$ is symmetric and $c\in\R$, has VC dimension at most $d_{\mathrm{VC}}$. Moreover, for every region $\cC_j=\cC_{k,h,j}$, $x\in\R^d$, and $t\in\R$, this class contains $\cC_j$, $\R^d\setminus\bigcup_j\cC_{k,h,j}$, and $\{\phi\in\cC_j:(x^\top\phi)^2>t\}$.
\end{lemma}

\begin{proof}
  For the first claim, every set $\{\phi:\phi^\top W\phi\le c\}$, and its complement, is a halfspace in the $d(d+1)/2$ coordinates $\{\phi_i\phi_{i'}\}_{i\le i'}$, so these sets have VC dimension at most $(d+1)(d+2)/2$ \citep[Theorem~9.3]{shalev2014understanding}. Hence, \citet[Lemma~3.2.3]{blumer1989learnability} bounds the VC dimension of their $(J_0+1)$-fold intersections by
  \begin{equation*}
    2\cdot\frac{(d+1)(d+2)}2(J_0+1)\log_2\left(3(J_0+1)\right)\le2(d+1)(d+2)(J_0+1)\log_2(J_0+2)\le d_{\mathrm{VC}}.
  \end{equation*}

  For the second claim, let $\Lambda_j$ be the final matrix of the $j$-th iteration of \Cref{alg:regions}. Then $\cC_j$ is the set $\{\phi:\phi^\top\Lambda_j^{-1}\phi\le\rho\}$ minus those sets for $j'<j$, and the features outside all regions are those outside all these sets. As there are $J\le J_0$ regions by \Cref{lem:lin-regions-det}, both sets are in the class, and so is
  \begin{equation*}
    \{\phi\in\cC_j:(x^\top\phi)^2>t\}=\cC_j\cap\{\phi:\phi^\top(xx^\top)\phi>t\}.\qedhere
  \end{equation*}
\end{proof}

\begin{lemma}\label{lem:lin-window}
  Fix $k\in\mathcal K$ and $h$, and let $\kappa_1,\ldots,\kappa_{\sqrt K}$ be the last $\sqrt K$ episodes processed before $k$, in the processing order. Then, with probability at least $1-3\delta/(64KH)-2e^{-\sqrt K/100}$, every set $T$ in the class of \Cref{lem:lin-vc} satisfies
  \begin{equation}\label{eq:lin-window}
    \frac{\Pr_{\pi_{\kappa_1}}(\phi_h\in T)-\bar\epsilon}3\le\frac1{\sqrt K}\sum_{i=1}^{\sqrt K}\ind[\phi_{\kappa_i,h}\in T]\le3\Pr_{\pi_{\kappa_1}}(\phi_h\in T)+\bar\epsilon,
  \end{equation}
  where $\bar\epsilon\triangleq(9/\sqrt K)(d_{\mathrm{VC}}\log(2e\sqrt K)+\log(512KH/\delta))$, and so does every $f\colon\R^d\to[0,1]$ whose level sets are in this class.
\end{lemma}

\begin{proof}
  We couple most of the features with i.i.d.\ samples from $\pi_{\kappa_1}$. Recall that $\Pr_{\pi_{\kappa_i}}$ is the law of a trajectory $\tau$ of $\pi_{\kappa_i}$ under $\PP$, and let $\phi_h(\tau)$ be its layer-$h$ feature. Fewer than $\sqrt K$ episodes are processed from $\kappa_1$ to $\kappa_i$, so $N(\kappa_1,\kappa_i)\le\sqrt K+2\dbar$ by \Cref{lem:counting}\ref{item:cnt-between}. Hence, as the density of $\Pr_{\pi_{\kappa_i}}$ with respect to $\Pr_{\pi_{\kappa_1}}$ is the product of the policy ratios along the trajectory (as in \Cref{eq:rejection-change}), \Cref{lem:lin-stability} gives $e^{-1/40}\Pr_{\pi_{\kappa_1}}\le\Pr_{\pi_{\kappa_i}}\le e^{1/40}\Pr_{\pi_{\kappa_1}}$ on every sample path.

Given $\cH_{\kappa_1-1}$, the policy $\pi_{\kappa_1}$, and thus $\Pr_{\pi_{\kappa_1}}$, is fixed. Let $\tau_1,\ldots,\tau_{\sqrt K}$ be i.i.d.\ trajectories from $\Pr_{\pi_{\kappa_1}}$, and let $U_1,\ldots,U_{\sqrt K}\sim\mathrm{Bernoulli}(e^{-1/40})$ be independent. Further let $\tau_1',\tau_2',\ldots$ be an infinite sequence of i.i.d.\ trajectories from $\Pr_{\pi_{\kappa_1}}$. We generate the trajectories of the episodes $\kappa_1,\kappa_2,\ldots$ in order. If $U_i=1$, the trajectory of $\kappa_i$ is $\tau_i$. Otherwise, we go through the trajectories $\tau'_r$ that have not been examined yet, in increasing order of $r$; for every such $\tau'_r$, we draw a new indicator $U'_r\in\{0,1\}$ that, conditionally on everything generated so far, equals $1$ with probability
\begin{equation*}
    \frac{\frac{\mathrm d\Pr_{\pi_{\kappa_i}}}{\mathrm d\Pr_{\pi_{\kappa_1}}}(\tau'_r)-e^{-1/40}}{e^{1/40}-e^{-1/40}}\in[0,1],
  \end{equation*}
  and the trajectory of $\kappa_i$ is the first examined $\tau'_r$ with $U'_r=1$. After the last episode $\kappa_{\sqrt K}$, we examine the remaining $\tau'_r$ with $r\le\sqrt K$ in the same way with $\pi_{\kappa_{\sqrt K}}$ in place of $\pi_{\kappa_i}$, so that $U'_1,\ldots,U'_{\sqrt K}$ are all defined. As an unexamined $\tau'_r$ is independent of everything generated before it is examined, and the density $\mathrm d\Pr_{\pi_{\kappa_i}}/\mathrm d\Pr_{\pi_{\kappa_1}}$ has mean $1$ under $\Pr_{\pi_{\kappa_1}}$, every $U'_r$ equals $1$ with probability $(1-e^{-1/40})/(e^{1/40}-e^{-1/40})=1/(1+e^{1/40})>0.49$ given everything generated before $\tau'_r$ is examined, and, given $U'_r=1$, the trajectory $\tau'_r$ has law $(\Pr_{\pi_{\kappa_i}}-e^{-1/40}\Pr_{\pi_{\kappa_1}})/(1-e^{-1/40})$. Hence, given $\cH_{\kappa_i-1}$, the trajectory of $\kappa_i$ has law
  \begin{equation*}
    e^{-1/40}\Pr_{\pi_{\kappa_1}}+\left(1-e^{-1/40}\right)\frac{\Pr_{\pi_{\kappa_i}}-e^{-1/40}\Pr_{\pi_{\kappa_1}}}{1-e^{-1/40}}=\Pr_{\pi_{\kappa_i}},
  \end{equation*}
  so this coupling does not change the joint distribution of the data, nor the probability of \Cref{eq:lin-window}.

  Next, by $\Pr(U_i=0)=1-e^{-1/40}<0.025$ and Hoeffding's inequality \citep[Eq.~(5.8)]{lattimore2020bandit}, we have
  \begin{equation*}
    \Pr\left(\sum_{i=1}^{\sqrt K}\ind[U_i=0]\ge\frac{\sqrt K}{10}\right)\le e^{-2\sqrt K(0.1-0.025)^2}\le e^{-\sqrt K/100}.
  \end{equation*}
  Similarly, $U'_1,U'_2,\ldots$ are i.i.d.\ with mean $1/(1+e^{1/40})>0.49$, so the same inequality shows that $\sum_{r=1}^{\sqrt K}U'_r<\sqrt K/10$ with probability at most $e^{-\sqrt K/100}$. Outside these two events, fewer than $\sqrt K/10$ episodes have $U_i=0$, and they use the first trajectories $\tau'_r$ with $U'_r=1$, which all have $r\le\sqrt K$. Hence, every episode $\kappa_i$ uses a different trajectory among $\tau_1,\ldots,\tau_{\sqrt K},\tau'_1,\ldots,\tau'_{\sqrt K}$.

  Finally, let $\cI$ be the first $\lceil\sqrt K/2\rceil$ indices $i$ with $U_i=1$, completed by the first indices with $U_i=0$ if there are fewer, so that, given $(U_i)_i$, $\{\tau_i\}_{i\in\cI}$ is an i.i.d.\ sample of size $\lceil\sqrt K/2\rceil$. As the class of \Cref{lem:lin-vc} has VC dimension at most $d_{\mathrm{VC}}$, \Cref{lem:vc} (used given $\cH_{\kappa_1-1}$ with $V=d_{\mathrm{VC}}$ and failure probability $\delta/(64KH)$) applies to the i.i.d.\ samples $\{\tau_i\}_{i\le\sqrt K}$, $\{\tau'_r\}_{r\le\sqrt K}$, and $\{\tau_i\}_{i\in\cI}$ with additive terms $\bar\epsilon/2$, $\bar\epsilon/2$, and $\bar\epsilon$, respectively. As every episode uses a different trajectory, we get
  \begin{equation*}
    \frac1{\sqrt K}\sum_{i=1}^{\sqrt K}\ind[\phi_{\kappa_i,h}\in T]\le\frac1{\sqrt K}\sum_{i=1}^{\sqrt K}\ind[\phi_h(\tau_i)\in T]+\frac1{\sqrt K}\sum_{r=1}^{\sqrt K}\ind[\phi_h(\tau'_r)\in T]\le3\Pr_{\pi_{\kappa_1}}(\phi_h\in T)+\bar\epsilon.
  \end{equation*}
  Moreover, outside the first event above, more than $9\sqrt K/10\ge\lceil\sqrt K/2\rceil$ indices have $U_i=1$, so the samples in $\cI$ are the trajectories of the episodes $\kappa_i$ with $i\in\cI$, and we have
  \begin{equation*}
    \Pr_{\pi_{\kappa_1}}(\phi_h\in T)\le\frac3{2|\cI|}\sum_{i\in\cI}\ind[\phi_h(\tau_i)\in T]+\bar\epsilon\le\frac3{\sqrt K}\sum_{i=1}^{\sqrt K}\ind[\phi_{\kappa_i,h}\in T]+\bar\epsilon.
  \end{equation*}
  The same bounds hold for $f$ by the version of \Cref{lem:vc} for functions.
\end{proof}

\subsection{Properties of MGRR}
\label{app:lin-key}

The following lemma shows that $Y_{k,h}$ has a conditional mean that is linear in the features.

\begin{lemma}\label{lem:lin-response}
  For every $k\in\mathcal K$ and layer $h$, there is an $\cH_{k-1}$-measurable vector $\xi_{k,h}\in\R^d$ with $\lVert\xi_{k,h}\rVert_2\le5H^2\sqrt d$ such that every $(s,a)\in\cS_h\times\cA$ satisfies
  \begin{equation}\label{eq:lin-xi}
    Q_k^{\pi_k}(s,a)-\sum_{s'}\PP(s'\mid s,a)\sum_{a'}\pi_k(a'\mid s')B_k(s',a')=\langle\phi(s,a),\xi_{k,h}\rangle=\E\left[Y_{k,h}\middle|\cH_{k-1},(s_{k,h},a_{k,h})=(s,a)\right].
  \end{equation}
\end{lemma}

\begin{proof}
  Let $\xi_{k,h}\triangleq\zeta_{k,h}+\sum_{s'\in\cS_{h+1}}\psi(s')(V_k^{\pi_k}(s')-\sum_{a'}\pi_k(a'\mid s')B_k(s',a'))$, without the sum if $h=H-1$. Then the first identity follows from \Cref{def:linear,eq:Q-function}, and $\lVert\xi_{k,h}\rVert_2\le5H^2\sqrt d$ follows from $0\le V_k^{\pi_k}\le H$ and $0\le B_k\le4H^2$ (\Cref{lem:lin-bonus-bounds}, stated and proved immediately after this lemma). For the second identity, let $W_{h'}$ be the bonus that \Cref{eq:app-response} subtracts from layer $h'$ onward, so that $L_{k,h}-Y_{k,h}=W_{h+1}$. By \Cref{eq:app-response,eq:app-linear-B}, we have
  \begin{equation*}
    W_{h'}=b_k(s_{k,h'})+4H(H-h'-1)u_k(s_{k,h'},a_{k,h'})+\left(1-u_k(s_{k,h'},a_{k,h'})\right)W_{h'+1}.
  \end{equation*}
  As episode $k$ follows $\pi_k$ given $\cH_{k-1}$ (\Cref{lem:processing}), backward induction implies $\E[W_{h'}\mid\cH_{k-1},(s_{k,h'},a_{k,h'})]=B_k(s_{k,h'},a_{k,h'})$. Hence, by $\E[L_{k,h}\mid\cH_{k-1},(s_{k,h},a_{k,h})=(s,a)]=Q_k^{\pi_k}(s,a)$, we have
  \begin{equation*}
    \E\left[Y_{k,h}\middle|\cH_{k-1},(s_{k,h},a_{k,h})=(s,a)\right]=Q_k^{\pi_k}(s,a)-\sum_{s'}\PP(s'\mid s,a)\sum_{a'}\pi_k(a'\mid s')B_k(s',a'),
  \end{equation*}
  which finishes the proof.
\end{proof}

Next, we bound the bonuses and the responses $Y_{k,h}$ in \Cref{eq:app-response}, which MGRR averages.

\begin{lemma}\label{lem:lin-bonus-bounds}
  For every $k\in\mathcal K$, $u_k$, $b_k$, and $B_k$ are $\cH_{k-1}$-measurable, $B_k$ is an optimistic Q-function of $\pi_k$ for $b_k$, and every $(s,a)\in\cS_h\times\cA$ satisfies $0\le b_k(s)\le4H$, $0\le B_k(s,a)\le4H(H-h)$, and $|Y_{k,h}|\le5H^2$.
\end{lemma}

\begin{proof}
  The regions and the numbers of unfinished estimates only depend on the episodes processed before $k$, so $u_k$, $b_k$, and $B_k$ are $\cH_{k-1}$-measurable by \Cref{lem:processing}. By \Cref{eq:app-lin-bonus} and $\phat_k(s,a)>\theta/2$ (\Cref{lem:lin-regions-det}), we have
  \begin{equation*}
    b_k(s)\le\frac{H\theta}{\theta/2}\sum_{a:\,u_k(s,a)=0}\pi_k(a\mid s)+4H\sum_{a:\,u_k(s,a)=1}\pi_k(a\mid s)\le4H.
  \end{equation*}
  Hence, by \Cref{eq:app-linear-B} and induction over the layers, we have
  \begin{equation*}
    B_k(s,a)\le b_k(s)+\max\left\{\max_{s',a'}B_k(s',a'),4H(H-h-1)\right\}\le4H+4H(H-h-1)=4H(H-h).
  \end{equation*}
  Moreover, \Cref{eq:app-linear-B} differs from the right-hand side of \Cref{eq:app-bonus-q} only when $u_k(s,a)=1$, where it replaces the expected future bonus by $4H(H-h-1)$, which is at least as large by the bound above, so $B_k$ is an optimistic Q-function of $\pi_k$ for $b_k$. Similarly, the bonus subtracted in \Cref{eq:app-response} is at most $4H(H-h-1)$, so $-4H^2\le Y_{k,h}\le H$.
\end{proof}

We now bound the sizes of MGRR pieces, which controls how much one processed episode changes the received losses.

\begin{lemma}\label{lem:lin-pieces}
  On every sample path, we have $M_j\le\sqrt K$, $|v_n(s,a)|\le5H^2/4$, and $\eta|\Qhat_k(s,a)|\le1/8$. Moreover, every piece changes $\Qtil_k(s,a)$ by at most $5H^2/2$, every initial value of $\Qtil_k(s,a)$ is at most $10H^2$ in absolute value, and every processed episode changes every received loss by at most $5(\nu+4)H^2/2$.
\end{lemma}

\begin{proof}
  By \Cref{lem:lin-regions-det} and \Cref{eq:lin-conditions}, $M_j=\lceil m_0/\phat_j\rceil\le\lceil2m_0/\theta\rceil\le\sqrt K$. By \Cref{lem:lin-regions-det}, every factor in \Cref{eq:app-matrix-update} lies between $I/2$ and $I$, so $\lVert Z_n\rVert_2\le1$. Hence, by \Cref{eq:app-increment} and \Cref{lem:lin-bonus-bounds}, we have
  \begin{equation*}
    |v_n(s,a)|\le\frac{\lVert\Sigmahat_j^{-1/2}\phi(s,a)\rVert_2\lVert Z_n\rVert_2\lVert\Sigmahat_j^{-1/2}\phi_{k,h}\rVert_2|Y_{k,h}|}{8\rho}\le\frac{2\rho\cdot5H^2}{8\rho}=\frac{5H^2}4.
  \end{equation*}
  As the initial value $-b_k(s)-4H(H-h-1)u_k(s,a)-K^{-3}$ lies in $[-5H^2,0]$, \Cref{eq:lin-conditions} implies
  \begin{equation*}
    \eta|\Qhat_k(s,a)|\le\eta\left(5H^2+\sqrt K\cdot\frac{5H^2}4\right)\le5\eta H^2(\sqrt K+1)\le\frac18.
  \end{equation*}
  Meanwhile, any two running values $y$ and $y'$ satisfy $|(y-\eta y^2)-(y'-\eta y'^2)|=|y-y'||1-\eta(y+y')|\le2|y-y'|$, which gives the claims on $\Qtil_k$. A pair at layer $h$ receives at most one initial value and the pieces of at most $\nu$ estimates when an episode is processed, so its received loss changes by at most
  \begin{equation*}
    2\cdot5H^2+\nu\cdot2\cdot\frac{5H^2}4=\frac{5(\nu+4)H^2}2,
  \end{equation*}
  which finishes the proof.
\end{proof}

As in \Cref{lem:stability}, \Cref{lem:lin-pieces} keeps the policies close during the lifetime of an estimate, so that every acceptance probability is at most $1$.

\begin{lemma}\label{lem:lin-stability}
  Every episode $t$, state $s$, and action $a$ satisfy $\eta|g_t(s,a)|\le1$. Moreover, if $N(t_1,t_2)\le\sqrt K+2\dbar$, then every partial trajectory $(s_0,a_0,\ldots,s_h,a_h)$ satisfies
  \begin{equation*}
    e^{-1/40}\le\prod_{h'=0}^h\frac{\pi_{t_1}(a_{h'}\mid s_{h'})}{\pi_{t_2}(a_{h'}\mid s_{h'})}\le e^{1/40},
  \end{equation*}
  and every state $s$ and action $a$ satisfy
  \begin{equation*}
    \left|\pi_{t_2}(a\mid s)-\pi_{t_1}(a\mid s)\right|\le10N(t_1,t_2)\,\eta(\nu+4)H^2\pi_{t_1}(a\mid s).
  \end{equation*}
\end{lemma}

\begin{proof}
  By \Cref{lem:lin-pieces}, $g_t(s,\cdot)$ takes values in an interval of length $5\mathfrak n_t(\nu+4)H^2$. Hence, by \Cref{lem:hedge-step}, every episode $t$, state $s$, and action $a$ satisfy
  \begin{equation}\label{eq:lin-step}
    \left|\log\pi_{t+1}(a\mid s)-\log\pi_t(a\mid s)\right|\le5\mathfrak n_t\eta(\nu+4)H^2.
  \end{equation}
  Moreover, $\mathfrak n_t\le\dbar+1$ by \Cref{lem:counting}\ref{item:cnt-batch}, so \Cref{eq:lin-conditions} gives $\eta|g_t(s,a)|\le5(\dbar+1)\eta(\nu+4)H^2/2\le1$. Summing \Cref{eq:lin-step} over the at most $H$ factors and over the updates at the ends of episodes $\min\{t_1,t_2\},\ldots,\max\{t_1,t_2\}-1$, whose $\mathfrak n_t$ add up to $N(t_1,t_2)\le\sqrt K+2\dbar$, \Cref{eq:lin-conditions} implies
  \begin{equation*}
    \left|\log\prod_{h'=0}^h\frac{\pi_{t_1}(a_{h'}\mid s_{h'})}{\pi_{t_2}(a_{h'}\mid s_{h'})}\right|\le5HN(t_1,t_2)\,\eta(\nu+4)H^2\le\frac1{40}.
  \end{equation*}
  Moreover, by $e^x-1\le2x$ for $x\le1$, we have
  \begin{equation*}
    \left|\pi_{t_2}(a\mid s)-\pi_{t_1}(a\mid s)\right|\le\left(e^{5N(t_1,t_2)\eta(\nu+4)H^2}-1\right)\pi_{t_1}(a\mid s)\le10N(t_1,t_2)\,\eta(\nu+4)H^2\pi_{t_1}(a\mid s),
  \end{equation*}
  which finishes the proof.
\end{proof}

In particular, if an estimate started when processing $k$ attempts with $\kappa$, then $\kappa$ is the $n$-th episode processed after $k$ for some $n\le M_j-1<\sqrt K$, so $N(k,\kappa)\le\sqrt K+2\dbar$ by \Cref{lem:counting}\ref{item:cnt-between}, and \Cref{lem:lin-stability} gives $\alpha_{k,\kappa}\le e^{1/40}/2<1$.

We now show that, when $(k,h)$ is good (\Cref{def:lin-good}), MGRR estimates $Q_k^{\pi_k}-B_k$ with a negligible bias at the estimated pairs, and we can also bound its second moment.

\begin{lemma}\label{lem:lin-mean}
  Let $(k,h)$ be good (\Cref{def:lin-good}), and let $(s,a)\in\cS_h\times\cA$. If $u_k(s,a)=1$, then $\Qhat_k(s,a)=-B_k(s,a)-K^{-3}$. If $u_k(s,a)=0$, then $\E[\Qhat_k(s,a)^2\mid\cH_{k-1}]\le100H^4m_0^2/\phat_k(s,a)$. In both cases,
  \begin{equation*}
    0\le Q_k^{\pi_k}(s,a)-B_k(s,a)-\E[\Qhat_k(s,a)\mid\cH_{k-1}]\le Hu_k(s,a)+2K^{-3}.
  \end{equation*}
\end{lemma}

\begin{proof}
  If $u_k(s,a)=1$, then $(s,a)$ receives no piece and keeps its initial value $-B_k(s,a)-K^{-3}$, which proves the claims as $0\le Q_k^{\pi_k}\le H$. Otherwise, we write $\Qhat_k(s,a)$ as a sum of pieces, compute the mean of every piece by the tower rule of expectation, and compare their total with $(Q_k^{\pi_k}-B_k)(s,a)$; we then bound the second moment.

  Let $\cC_j\ni\phi(s,a)$ and $M\triangleq M_j$, so that an estimate starts in $\cC_j$ if and only if $\phi_{k,h}\in\cC_j$, and define $Z_n$ by \Cref{eq:app-matrix-update} with the $n$-th episode processed after $k$ for all $n\le M-1$, even if no estimate starts. Let
  \begin{equation*}
    X\triangleq\frac{\Sigmahat_j^{-1/2}\Sigma_{h,j}^{\pi_k}\Sigmahat_j^{-1/2}}{8\rho},\;\Pi\triangleq\Sigmahat_j\Sigmahat_j^\dagger,\; r\triangleq\Sigmahat_j^{-1/2}\E_{\pi_k}\left[\ind[\phi_h\in\cC_j]\phi_h\left\langle(I-\Pi)\phi_h,\xi_{k,h}\right\rangle\right],
  \end{equation*}
  where $\phat_jI/(32\rho)\preceq X\preceq I/2$ and $\lVert(I-X)^n\rVert_2\le(1-\phat_j/(32\rho))^n$ on $\operatorname{range}(\Sigmahat_j)$, as $(k,h)$ is good and by \Cref{lem:lin-regions-det}.

  For the mean, by \Cref{eq:app-increment} and \Cref{lem:lin-lifetime} (stated and proved immediately after this lemma), which ensures that all $M-1$ attempts are made, $\Qhat_k(s,a)=-b_k(s)-K^{-3}+\sum_{n=0}^{M-1}v_n(s,a)$, where
  \begin{equation*}
    v_n(s,a)=\frac1{8\rho}\phi(s,a)^\top\Sigmahat_j^{-1/2}Z_n\,\ind[\phi_{k,h}\in\cC_j]\Sigmahat_j^{-1/2}\phi_{k,h}Y_{k,h}.
  \end{equation*}
  Let $\kappa$ be the $n$-th episode processed after $k$. By \Cref{lem:lin-stability,lem:rejection-general} with $f=\ind[\phi_h\in\cC_j]\phi_h\phi_h^\top$, every factor in \Cref{eq:app-matrix-update} has conditional mean $I-X$ given $\cH_{\kappa-1}$, so the tower rule of expectation implies
  \begin{equation*}
    \E\left[Z_n\ind[\phi_{k,h}\in\cC_j]\Sigmahat_j^{-1/2}\phi_{k,h}Y_{k,h}\middle|\cH_{k-1}\right]=(I-X)^n\,\E\left[\ind[\phi_{k,h}\in\cC_j]\Sigmahat_j^{-1/2}\phi_{k,h}Y_{k,h}\middle|\cH_{k-1}\right],
  \end{equation*}
  where, by \Cref{lem:lin-response} and $\Pi=\Sigmahat_j^{1/2}\Sigmahat_j^{-1/2}$,
  \begin{equation*}
    \E\left[\ind[\phi_{k,h}\in\cC_j]\Sigmahat_j^{-1/2}\phi_{k,h}Y_{k,h}\middle|\cH_{k-1}\right]=8\rho X\Sigmahat_j^{1/2}\xi_{k,h}+r.
  \end{equation*}
  As $\sum_{n=0}^{M-1}(I-X)^nX=I-(I-X)^M$ on $\operatorname{range}(\Sigmahat_j)$, we get
  \begin{equation*}
    \E\left[\sum_{n=0}^{M-1}v_n(s,a)\middle|\cH_{k-1}\right]=\phi(s,a)^\top\Sigmahat_j^{-1/2}\left(I-(I-X)^M\right)\Sigmahat_j^{1/2}\xi_{k,h}+\frac1{8\rho}\phi(s,a)^\top\Sigmahat_j^{-1/2}\sum_{n=0}^{M-1}(I-X)^nr.
  \end{equation*}
  Meanwhile, by $u_k(s,a)=0$ and \Cref{eq:app-linear-B,eq:lin-xi}, we have
  \begin{equation*}
    Q_k^{\pi_k}(s,a)-B_k(s,a)=\phi(s,a)^\top\Sigmahat_j^{-1/2}\Sigmahat_j^{1/2}\xi_{k,h}+\langle(I-\Pi)\phi(s,a),\xi_{k,h}\rangle-b_k(s).
  \end{equation*}
  Comparing the two displays, we get
  \begin{align*}
    &\quad\left|Q_k^{\pi_k}(s,a)-B_k(s,a)-\E[\Qhat_k(s,a)\mid\cH_{k-1}]-K^{-3}\right|\\
    &\le\left|\langle(I-\Pi)\phi(s,a),\xi_{k,h}\rangle\right|+\left|\phi(s,a)^\top\Sigmahat_j^{-1/2}(I-X)^M\Sigmahat_j^{1/2}\xi_{k,h}\right|+\frac1{8\rho}\left|\phi(s,a)^\top\Sigmahat_j^{-1/2}\sum_{n=0}^{M-1}(I-X)^nr\right|\\
    &\overset{(a)}\le2\sqrt{\rho\lambda}\cdot5H^2\sqrt d+\sqrt{2\rho}\,e^{-\frac{M\phat_j}{32\rho}}\cdot5H^2\sqrt d+\frac{\sqrt{2\rho}}{8\rho}\cdot\frac{32\rho}{\phat_j}\cdot4\phat_j\sqrt{2\rho}\cdot2\sqrt{\rho\lambda}\cdot5H^2\sqrt d\\
    &\overset{(b)}\le C_1H^2\sqrt d\left(\rho^{3/2}\sqrt\lambda+\sqrt\rho\,e^{-m_0/(32\rho)}\right)\overset{(c)}\le K^{-3},
  \end{align*}
  where (a) used \Cref{lem:lin-response,lem:lin-regions-det}, $\lVert\Sigmahat_j^{1/2}\rVert_2\le1$, $\Pr_{\pi_k}(\phi_h\in\cC_j)\le4\phat_j$, and $\sum_n\lVert(I-X)^n\rVert_2\le32\rho/\phat_j$, (b) used $M\phat_j\ge m_0$ and $\rho\ge1$, and (c) used \Cref{eq:lin-conditions}.

  For the second moment, the at most $M\le2m_0/\phat_j$ pieces, each at most $5H^2/4$ by \Cref{lem:lin-pieces}, are nonzero only if $\phi_{k,h}\in\cC_j$, which has probability at most $4\phat_j$ given $\cH_{k-1}$. Hence, as the initial value is at most $5H^2$ in absolute value, we have
  \begin{equation*}
    \E[\Qhat_k(s,a)^2\mid\cH_{k-1}]\le2(5H^2)^2+2\cdot4\phat_j\left(\frac{2m_0}{\phat_j}\cdot\frac{5H^2}4\right)^2=50H^4+\frac{50H^4m_0^2}{\phat_j}\le\frac{100H^4m_0^2}{\phat_j},
  \end{equation*}
  which finishes the proof.
\end{proof}

Condition \textit{iii)} ensures that every estimate makes all its attempts before the game ends.

\begin{lemma}\label{lem:lin-lifetime}
  Every estimate started in a region $\cC_j$ makes its $M_j-1$ attempts by the end of episode $K-1$.
\end{lemma}

\begin{proof}
  By \Cref{lem:lin-pieces}, $M_j\le\sqrt K$. A non-skipped episode $k$ is processed at the end of episode $k+\fd_k$. For $1\le r<\sqrt K$, episode $k+\fd_k+r$ is processed after $k$, at the end of an episode at most
  \begin{equation*}
    k+\fd_k+r+\dbar\le k+\fd_k+\sqrt K-1+\dbar\le K-1
  \end{equation*}
  by condition \textit{iii)}. Hence, at least $\sqrt K-1\ge M_j-1$ episodes are processed after $k$ by the end of episode $K-1$.
\end{proof}

Finally, as in \Cref{lem:unfinished}, no layer accumulates $\nu$ unfinished estimates, so every pair whose feature lies in a region is estimated.

\begin{lemma}\label{lem:lin-capacity}
  With probability at least $1-\delta/8$, if every $(k,h)$ is good (\Cref{lem:lin-good}), then every layer has fewer than $\nu$ unfinished estimates before processing every episode $k\in\mathcal K$.
\end{lemma}

\begin{proof}
  Fix $k\in\mathcal K$ and a layer $h$. For $t\ge1$, let $\kappa$ be the $t$-th episode processed before $k$, counting backwards, which is fixed as the delays are fixed in advance. An estimate started in a region $\cC_{\kappa,h,j}$ when processing $\kappa$ is still unfinished before processing $k$ if and only if $M_j>t$, where $M_j\le\sqrt K$. Let $X_{\kappa,t}\in\{0,1\}$ indicate that $\kappa\in\mathcal K$, $(\kappa,h)$ is good, and episode $\kappa$ starts an estimate at layer $h$ in a region $\cC_{\kappa,h,j}$ with $M_j>t$. If every $(\kappa,h)$ is good, then the number of unfinished estimates at layer $h$ before processing $k$ is $\sum_{t\le\sqrt K}X_{\kappa,t}$.

  We now bound the conditional mean of $X_{\kappa,t}$. As $\phat_{\kappa,h,1}\ge\phat_{\kappa,h,2}\ge\cdots$, for every $x>0$, the regions with $\phat_{\kappa,h,j'}\le x$ are formed from the $m_j<2\sqrt K\phat_{\kappa,h,j}$ features left for the first such region $\cC_{\kappa,h,j}$, so
  \begin{equation}\label{eq:lin-small-regions}
    \sum_{j':\,\phat_{\kappa,h,j'}\le x}\phat_{\kappa,h,j'}\le\frac{m_j}{\sqrt K}<2x.
  \end{equation}
  If $(\kappa,h)$ is good, every region is visited with probability at most $4\phat_{\kappa,h,j}$ given $\cH_{\kappa-1}$, and $M_j>t\ge2$ implies $\phat_{\kappa,h,j}<m_0/(t-1)$, so \Cref{eq:lin-small-regions} gives
  \begin{equation*}
    \E\left[X_{\kappa,t}\middle|\cH_{\kappa-1}\right]\le\sum_{j:\,\phat_{\kappa,h,j}<m_0/(t-1)}4\phat_{\kappa,h,j}<\frac{8m_0}{t-1}.
  \end{equation*}
  As whether $(\kappa,h)$ is good is $\cH_{\kappa-1}$-measurable, this bound holds on every sample path. Therefore, by \Cref{lem:freedman}\ref{lem:adapted-upper} with $R=1$ and $\delta'=\delta/(8KH)$, except with probability $\delta/(8KH)$, we have
  \begin{equation*}
    \sum_{t\le\sqrt K}X_{\kappa,t}\le(e-1)\left(1+\sum_{t=2}^{\sqrt K}\min\left\{1,\frac{8m_0}{t-1}\right\}\right)+\log\frac{8KH}\delta\le2\left(2+8m_0(1+\log\sqrt K)\right)+\iota<\nu,
  \end{equation*}
  where we used $e-1\le2$, $\sum_{t=2}^{\sqrt K}1/(t-1)\le1+\log\sqrt K$, $\log(8KH/\delta)\le\iota$, and $\nu\ge C_0m_0\iota$ with $\log K\le\iota$. A union bound over $k$ and $h$ finishes the proof.
\end{proof}

Hence, on the events of \Cref{lem:lin-good,lem:lin-capacity}, $u_k(s,a)=1$ if and only if $\phi(s,a)$ lies outside all regions.

\subsection{Bounding \texorpdfstring{\BiasOne}{Bias-1} and \texorpdfstring{\BiasTwo}{Bias-2}}
\label{app:lin-bias}

Unlike in \Cref{app:tabular}, the estimates of all pairs in a region share the same attempts, so \Cref{lem:attempts} does not apply. Instead, we derive \Cref{lem:lin-exp} below to control the exponential moment of $\Qtil_k(s,a)$ for every pair, and \Cref{lem:lin-slots} to combine these moments over estimates that overlap in time.

\begin{lemma}\label{lem:lin-exp}
  Let $\cH'$ be a $\sigma$-algebra, $X$ a random variable with $|X|\le R$ and $\eta R\le1/4$, and $m$ an $\cH'$-measurable upper bound on $\E[X\mid\cH']$. Then $\E[e^{\eta(X-\eta X^2-m)}\mid\cH']\le1$ and
  \begin{equation*}
    \E\left[\exp\left(\eta\left(\E[X-\eta X^2\mid\cH']-X+\eta X^2\right)-2\eta^2\E[X^2\mid\cH']\right)\middle|\cH'\right]\le1.
  \end{equation*}
\end{lemma}

\begin{proof}
  Both inequalities follow the argument of \citet{bartlett2008high} and \citet{zimmert2022return}. For the first one, by $e^{v-v^2}\le1+v$ for $v\ge-1/2$ with $v=\eta X$, and by $1+x\le e^x$, we have
  \begin{equation*}
    \E\left[e^{\eta(X-\eta X^2)}\middle|\cH'\right]\le1+\eta\E[X\mid\cH']\le1+\eta m\le e^{\eta m}.
  \end{equation*}
  For the second one, let $Z\triangleq\eta(\E[X-\eta X^2\mid\cH']-X+\eta X^2)$. As $X-\eta X^2$ is increasing in $X\in[-R,R]$, it takes values in an interval of length $2R$, so $|Z|\le2\eta R\le1/2$. Hence, by $\E[Z\mid\cH']=0$, $e^v\le1+v+v^2$ for $|v|\le1/2$, and $|X-\eta X^2|\le(5/4)|X|$, we have
  \begin{equation*}
    \E\left[e^Z\middle|\cH'\right]\le1+\E[Z^2\mid\cH']\le1+\eta^2\E\left[(X-\eta X^2)^2\middle|\cH'\right]\le\exp\left(2\eta^2\E[X^2\mid\cH']\right),
  \end{equation*}
  which finishes the proof.
\end{proof}

The estimates of different episodes overlap in time, so the variables in \Cref{lem:lin-exp} are not adapted to a single filtration. Adopting the reduction idea of \citet{joulani2013online}, which gives every round to a copy of a base algorithm whose earlier feedback has all arrived, we also put every variable into a slot when it starts and only reuse a slot after the previous variable in it is determined.

\begin{lemma}\label{lem:lin-slots}
  Let $\cH'_1,\ldots,\cH'_{n_0}$ be $\sigma$-algebras, and let $X_1,\ldots,X_{n_0}$ be random variables in slots $i_1,\ldots,i_{n_0}\in[N]$, where $i_n$ is $\cH'_n$-measurable. Suppose that $\E[e^{\eta X_n}\mid\cH'_n]\le1$ for every $n$, and that $X_n\ind[i_n=i_{n'}]$ is $\cH'_{n'}$-measurable for all $n<n'$. Then, with probability at least $1-\delta'$,
  \begin{equation*}
    \sum_{n=1}^{n_0}X_n\le\frac{N\log(1/\delta')}\eta.
  \end{equation*}
\end{lemma}

\begin{proof}
  Fix a slot $i$, and let $Y_n\triangleq X_n\ind[i_n=i]$. For every $n'$, $Y_{n'}=0$ unless $i_{n'}=i$, and, on the event $\{i_{n'}=i\}$, the sum $\sum_{n<n'}Y_n=\sum_{n<n'}X_n\ind[i_n=i_{n'}]$ is $\cH'_{n'}$-measurable. Hence, we have
  \begin{equation*}
    \E\left[e^{\eta\sum_{n\le n'}Y_n}\right]=\E\left[e^{\eta\sum_{n<n'}Y_n}\left(\ind[i_{n'}\ne i]+\ind[i_{n'}=i]\,\E\left[e^{\eta X_{n'}}\middle|\cH'_{n'}\right]\right)\right]\le\E\left[e^{\eta\sum_{n<n'}Y_n}\right],
  \end{equation*}
  and induction over $n'$ gives $\E[\exp(\eta\sum_nX_n\ind[i_n=i])]\le1$. Therefore, by H\"older's inequality,
  \begin{equation*}
    \E\left[\exp\left(\frac\eta N\sum_nX_n\right)\right]=\E\left[\prod_{i=1}^N\exp\left(\frac\eta N\sum_nX_n\ind[i_n=i]\right)\right]\le\prod_{i=1}^N\left(\E\left[\exp\left(\eta\sum_nX_n\ind[i_n=i]\right)\right]\right)^{1/N}\le1,
  \end{equation*}
  and Markov's inequality finishes the proof.
\end{proof}

We are now ready to bound \BiasOne and \BiasTwo.

\begin{lemma}[Bounding \BiasOne and \BiasTwo]\label{lem:lin-bias}
  With probability at least $1-\delta/4$, if every $(k,h)$ is good (\Cref{lem:lin-good}), then $\BiasTwo\le H\nu\iota/\eta$ and
  \begin{equation*}
    \BiasOne\le\frac12\sum_k\sum_{h,s}q^\star(s)b_k(s)+\frac{H\nu\iota}\eta+2HK^{-2}.
  \end{equation*}
\end{lemma}

\begin{proof}
  For $k\in\mathcal K$ and a layer $h$, let
  \begin{align*}
    X_{k,h}^{(1)}&\triangleq\ind[(k,h)\text{ is good}]\sum_{s\in\cS_h}q^\star(s)\sum_a\pi_k(a\mid s)\left(\E[\Qtil_k\mid\cH_{k-1}]-\Qtil_k-2\eta\E[\Qhat_k^2\mid\cH_{k-1}]\right)(s,a),\\
    X_{k,h}^{(2)}&\triangleq\ind[(k,h)\text{ is good}]\sum_{s\in\cS_h}q^\star(s)\sum_a\pi^\star(a\mid s)\left(\Qtil_k-Q_k^{\pi_k}+B_k\right)(s,a).
  \end{align*}
  If $(k,h)$ is good, \Cref{lem:lin-mean} shows that $m=(Q_k^{\pi_k}-B_k)(s,a)$ is an $\cH_{k-1}$-measurable upper bound on $\E[\Qhat_k(s,a)\mid\cH_{k-1}]$, and \Cref{lem:lin-pieces} gives $\eta|\Qhat_k(s,a)|\le1/8$. Hence, as $\sum_{s\in\cS_h}q^\star(s)=1$ and the indicator is $\cH_{k-1}$-measurable by \Cref{lem:lin-good}, Jensen's inequality and \Cref{lem:lin-exp} with $X=\Qhat_k(s,a)$ give (for $i=1,2$)
  \begin{equation}\label{eq:lin-exp-moment}
    \E\left[e^{\eta X_{k,h}^{(i)}}\middle|\cH_{k-1}\right]\le1.
  \end{equation}

  We now assign the slots for \Cref{lem:lin-slots}. When processing $k\in\mathcal K$, before its attempts, we put the pairs $(k,h)$ whose layer $h$ has fewer than $\nu$ unfinished estimates into slots, in increasing order of $h$. Every such $(k,h)$ goes into the smallest slot in $[H\nu]$ that no \emph{active} pair occupies, where a pair $(k',h')$ in a slot is active until the estimate started at $(k',h')$ makes its last attempt, or, if no estimate starts at $(k',h')$, until the processing of $k'$ is complete. Such a slot exists, as an active pair $(k',h')$ with $k'\ne k$ has an unfinished estimate at layer $h'$, and every layer has at most $\nu$ unfinished estimates at any time, since an estimate only starts at a layer with fewer than $\nu$ of them. Hence, when $(k,h)$ is put into a slot, every layer $h''\ne h$ has at most $\nu$ active pairs, namely at most $\nu$ unfinished estimates if $(k,h'')$ is not in a slot, and at most $\nu-1$ unfinished estimates and $(k,h'')$ otherwise, while layer $h$ has at most $\nu-1$ active pairs, which is fewer than $H\nu$ in total.

  We order the pairs in slots by the processing order of $k$ and then by $h$, and use $\cH'=\cH_{k-1}$ for $(k,h)$. Whether an estimate is unfinished before processing $k$ is $\cH_{k-1}$-measurable by \Cref{lem:processing}, so the slots of $k$ are $\cH_{k-1}$-measurable. If $(k',h')$ and $(k,h)$ share a slot and $(k',h')$ comes first, then $k'\ne k$ and $(k',h')$ is inactive before processing $k$, so $\Qhat_{k'}$ is final on $\cS_{h'}$ and $X_{k',h'}^{(i)}$ is $\cH_{k-1}$-measurable. Together with \Cref{eq:lin-exp-moment}, the conditions of \Cref{lem:lin-slots} hold. If $(k,h)$ is not in a slot, every $(s,a)\in\cS_h\times\cA$ has $u_k(s,a)=1$, so, by \Cref{lem:lin-mean}, $\Qhat_k(s,a)=-B_k(s,a)-K^{-3}$ is $\cH_{k-1}$-measurable, $\Qtil_k(s,a)\le\Qhat_k(s,a)$, and $X_{k,h}^{(1)},X_{k,h}^{(2)}\le0$. Therefore, by \Cref{lem:lin-slots} with $N=H\nu$ and $\delta'=\delta/8$, applied once for each $i=1,2$, except with probability $\delta/4$, we have (for $i=1,2$)
  \begin{equation}\label{eq:lin-bias-mart}
    \sum_{k,h}X_{k,h}^{(i)}\le\sum_{(k,h)\text{ in a slot}}X_{k,h}^{(i)}\le\frac{H\nu\iota}\eta.
  \end{equation}

  In the rest of the proof, we assume that every $(k,h)$ is good and \Cref{eq:lin-bias-mart} holds. Then $\BiasTwo=\sum_{k,h}X_{k,h}^{(2)}\le H\nu\iota/\eta$. For $\BiasOne$, by $\E[\Qtil_k\mid\cH_{k-1}]=\E[\Qhat_k\mid\cH_{k-1}]-\eta\E[\Qhat_k^2\mid\cH_{k-1}]$ and \Cref{lem:lin-mean}, every pair satisfies
  \begin{align*}
    Q_k^{\pi_k}-B_k-\Qtil_k&=\left(Q_k^{\pi_k}-B_k-\E[\Qhat_k\mid\cH_{k-1}]\right)+\eta\E[\Qhat_k^2\mid\cH_{k-1}]+\left(\E[\Qtil_k\mid\cH_{k-1}]-\Qtil_k\right)\\
    &\le Hu_k+2K^{-3}+3\eta\E[\Qhat_k^2\mid\cH_{k-1}]+\left(\E[\Qtil_k\mid\cH_{k-1}]-\Qtil_k-2\eta\E[\Qhat_k^2\mid\cH_{k-1}]\right).
  \end{align*}
  We sum it with the weights $q^\star(s)\pi_k(a\mid s)$. The last term adds up to $\sum_{k,h}X_{k,h}^{(1)}\le H\nu\iota/\eta$, and $2K^{-3}$ adds up to at most $2HK^{-2}$. For the other two terms, by \Cref{lem:lin-mean}, $\E[\Qhat_k(s,a)^2\mid\cH_{k-1}]$ is at most $(5H^2)^2$ if $u_k(s,a)=1$ and at most $100H^4m_0^2/\phat_k(s,a)$ if $u_k(s,a)=0$. Moreover, \Cref{eq:lin-conditions} gives $75\eta H^3\le1$ and $600\eta H^3m_0^2\le\theta$. Hence, every state $s$ satisfies
  \begin{align*}
    \sum_a\pi_k(a\mid s)\left(Hu_k(s,a)+3\eta\E[\Qhat_k(s,a)^2\mid\cH_{k-1}]\right)&\le2H\sum_{a:\,u_k(s,a)=1}\pi_k(a\mid s)+\frac{H\theta}2\sum_{a:\,u_k(s,a)=0}\frac{\pi_k(a\mid s)}{\phat_k(s,a)}\\
    &=\frac{b_k(s)}2,
  \end{align*}
  where the last step used \Cref{eq:app-lin-bonus}. Combining the three parts finishes the proof.
\end{proof}

\subsection{Bounding \texorpdfstring{\RegTerm}{Reg-Term}}
\label{app:lin-reg}

As in \Cref{app:tab-reg}, we split $\RegTerm$ into the regret of exponential weights on the received losses and the delay cost of the pieces. Unlike in \Cref{prop:reg}, every estimate makes exactly $M_j-1$ attempts, all by the end of episode $K-1$ (\Cref{lem:lin-lifetime}).

\begin{lemma}[Bounding \RegTerm]\label{lem:lin-reg}
  With probability at least $1-\delta/4$, if every $(k,h)$ is good (\Cref{lem:lin-good}), then
  \begin{equation*}
    \RegTerm\le\frac{H\log A}\eta+\frac12\sum_k\sum_{h,s}q^\star(s)b_k^{(1)}(s)+\O\left(\eta H^5\nu^2(K+\mD)+\eta\nu H^5m_0\mD+H^2\iota\left(\sqrt K+\dbar\right)\right).
  \end{equation*}
\end{lemma}

\begin{proof}
  Fix $k\in\mathcal K$ and $(s,a)\in\cS_h\times\cA$. For $n\ge0$, let $\kappa_n$ be the $n$-th episode processed after $k$, where $\kappa_0\triangleq k$, and let $\Delta_{k,n}(s,a)$ be the change of $\Qtil_k(s,a)$ when $\kappa_n$ is processed. This change is collected by the received loss $g_{t_{\kappa_n}}(s,a)$, which the policy $\pi_{t_{\kappa_n}}$ faces. By \Cref{lem:lin-pieces}, $|\Delta_{k,0}(s,a)|\le10H^2+5H^2/2\le13H^2$, as processing $k$ adds the initial value and one piece, and $|\Delta_{k,n}(s,a)|\le5H^2/2$ for $1\le n\le M_j-1$, while $\Delta_{k,n}(s,a)=0$ for $n\ge M_j$. As every estimate makes all its attempts by the end of episode $K-1$, we have $\sum_n\Delta_{k,n}(s,a)=\Qtil_k(s,a)$ and $\sum_tg_t(s,a)=\sum_k\Qtil_k(s,a)$. Hence, writing $\pi_k=\pi_{t_{\kappa_n}}+(\pi_k-\pi_{t_{\kappa_n}})$, we get
  \begin{equation*}
    \RegTerm=\underbrace{\sum_{t=1}^K\sum_{h,s}q^\star(s)\sum_a\left(\pi_t-\pi^\star\right)(a\mid s)\,g_t(s,a)}_{\textsc{Exp-Weights}}+\underbrace{\sum_k\sum_{h,s}q^\star(s)\sum_a\sum_{n\ge0}\left(\pi_k-\pi_{t_{\kappa_n}}\right)(a\mid s)\,\Delta_{k,n}(s,a)}_{\textsc{Delay-Cost}}.
  \end{equation*}

  For the \textsc{Exp-Weights} term, by \Cref{lem:lin-pieces}, every processed episode changes every received loss by at most $5(\nu+4)H^2/2$, so $|g_t(s,a)|\le5\mathfrak n_t(\nu+4)H^2/2$. Moreover, $\eta|g_t(s,a)|\le1$ by \Cref{lem:lin-stability}. Hence, by \Cref{lem:hedge} at every state, $\sum_{h,s}q^\star(s)=H$, and $\sum_t\mathfrak n_t^2\le K+2\mD$ (\Cref{lem:counting}\ref{item:cnt-batch}), we have
  \begin{equation*}
    \textsc{Exp-Weights}\le\sum_{h,s}q^\star(s)\left(\frac{\log A}\eta+\eta\sum_{t,a}\pi_t(a\mid s)g_t(s,a)^2\right)\le\frac{H\log A}\eta+\eta H(K+2\mD)\left(\frac{5(\nu+4)H^2}2\right)^2.
  \end{equation*}

  For the \textsc{Delay-Cost} term, let $n\le M_j-1$. By \Cref{lem:counting}\ref{item:cnt-delay}, $N(k,t_{\kappa_n})\le n+\fd_k+\sigma_k\le\sqrt K+2\dbar$, as $M_j\le\sqrt K$ (\Cref{lem:lin-pieces}) and $\fd_k,\sigma_k\le\dbar$. Hence, \Cref{lem:lin-stability} gives
  \begin{equation*}
    \left|\pi_{t_{\kappa_n}}(a\mid s)-\pi_k(a\mid s)\right|\le10\left(n+\fd_k+\sigma_k\right)\eta(\nu+4)H^2\pi_k(a\mid s).
  \end{equation*}
  As $|\Delta_{k,0}(s,a)|\le13H^2$ and $\sum_k(\fd_k+\sigma_k)\le2\mD$ (\Cref{lem:counting}\ref{item:cnt-delay}), the terms with $n=0$ add up to at most
  \begin{equation*}
    \sum_k\sum_{h,s}q^\star(s)\sum_a10\left(\fd_k+\sigma_k\right)\eta(\nu+4)H^2\pi_k(a\mid s)\cdot13H^2\le260\eta(\nu+4)H^5\mD.
  \end{equation*}
  For $n\ge1$, an estimate started in a region $\cC_{k,h,j}$ only changes $\Qtil_k(s,a)$ if $\phi(s,a)\in\cC_{k,h,j}$, and episode $k$ starts it if and only if $(1-u_k(s_{k,h},a_{k,h}))\ind[\phi_{k,h}\in\cC_{k,h,j}]=1$. Let
  \begin{equation*}
    \ind_{k,h,j}\triangleq\ind[(k,h)\text{ is good}]\bigl(1-u_k(s_{k,h},a_{k,h})\bigr)\ind[\phi_{k,h}\in\cC_{k,h,j}]\quad\text{and}\quad w_{k,h,j}\triangleq\sum_{s\in\cS_h}q^\star(s)\sum_{a:\,\phi(s,a)\in\cC_{k,h,j}}\pi_k(a\mid s).
  \end{equation*}
  As $\sum_{1\le n<M_j}10(n+\fd_k+\sigma_k)\cdot5/2\le25M_j^2+25M_j(\fd_k+\sigma_k)$, if every $(k,h)$ is good, the terms with $n\ge1$ add up to at most $25\eta(\nu+4)H^4\sum_k(\Gamma_k+\Gamma_k')$, where
  \begin{equation*}
    \Gamma_k\triangleq\sum_{h,j}\ind_{k,h,j}M_j^2w_{k,h,j}\quad\text{and}\quad\Gamma_k'\triangleq\left(\fd_k+\sigma_k\right)\sum_{h,j}\ind_{k,h,j}M_jw_{k,h,j}.
  \end{equation*}
  As at most one estimate starts per layer, $M_j\le\sqrt K$, and $\sum_jw_{k,h,j}\le1$, we have $0\le\Gamma_k\le HK$ and $0\le\Gamma_k'\le2\dbar H\sqrt K$. If $(k,h)$ is good, an estimate starts in $\cC_{k,h,j}$ with probability at most $4\phat_{k,h,j}$ given $\cH_{k-1}$. Moreover, an estimate only starts at a layer with fewer than $\nu$ unfinished estimates, where every pair whose feature lies in a region has $u_k(s,a)=0$. Hence, by $M_j\le2m_0/\phat_{k,h,j}$ and the definition of $b_k^{(1)}$ in \Cref{eq:lin-bonus-est}, we have
  \begin{equation*}
    \E\left[\Gamma_k\middle|\cH_{k-1}\right]\le\sum_{h,j}4\phat_{k,h,j}\left(\frac{2m_0}{\phat_{k,h,j}}\right)^2\sum_{s\in\cS_h}q^\star(s)\sum_{\substack{a:\,\phi(s,a)\in\cC_{k,h,j},\\u_k(s,a)=0}}\pi_k(a\mid s)\le\frac{16m_0^2}{H\theta}\sum_{h,s}q^\star(s)b_k^{(1)}(s),
  \end{equation*}
  and, in the same way, $\E[\Gamma_k'\mid\cH_{k-1}]\le8m_0H(\fd_k+\sigma_k)$. Thus, by \Cref{lem:freedman}\ref{lem:adapted-upper} with $R=HK$ and $R=2\dbar H\sqrt K$, respectively, and $\delta'=\delta/8$ for both (if $\dbar=0$, then $\Gamma_k'=0$ and the second application is not needed), except with probability $\delta/4$,
  \begin{align*}
    25\eta(\nu+4)H^4\sum_k(\Gamma_k+\Gamma_k')&\le25\eta(\nu+4)H^4\Bigl(\frac{16(e-1)m_0^2}{H\theta}\sum_k\sum_{h,s}q^\star(s)b_k^{(1)}(s)+HK\iota\\
    &\qquad+32m_0H\mD+2\dbar H\sqrt K\iota\Bigr)\\
    &\le\frac12\sum_k\sum_{h,s}q^\star(s)b_k^{(1)}(s)+\frac{H^2\sqrt K\iota}8+800\eta(\nu+4)H^5m_0\mD+\frac{H^2\dbar\iota}4,
  \end{align*}
  where we used $(e-1)\sum_k8m_0H(\fd_k+\sigma_k)\le32m_0H\mD$, $800(e-1)\eta(\nu+4)H^3m_0^2\le\theta$, and $\eta(\nu+4)H^3\le1/(200(\sqrt K+2\dbar))$ by \Cref{eq:lin-conditions}. Combining the bounds and using $\nu+4\le5\nu$ proves the claim.
\end{proof}

\subsection{Bounding \texorpdfstring{\Bonus}{Bonus}}
\label{app:lin-bonus}

Finally, as in \Cref{app:tab-bonus}, we bound $\Bonus$ together with the bonuses used in \Cref{lem:lin-bias,lem:lin-reg}.

\begin{lemma}[Bounding \Bonus]\label{lem:lin-bonus}
  If every $(k,h)$ is good (\Cref{lem:lin-good}) and every layer has fewer than $\nu$ unfinished estimates before processing every $k\in\mathcal K$ (\Cref{lem:lin-capacity}), then every $k\in\mathcal K$ satisfies
  \begin{equation*}
    \sum_{h,s}q^\star(s)\sum_a\left(\pi_k-\pi^\star\right)(a\mid s)B_k(s,a)+\sum_{h,s}q^\star(s)b_k(s)\le\sum_a\pi_k(a\mid s_0)B_k(s_0,a)=\O(H^3\theta\log K).
  \end{equation*}
\end{lemma}

\begin{proof}
  As $B_k$ is an optimistic Q-function of $\pi_k$ for $b_k$ (\Cref{lem:lin-bonus-bounds}), the inequality is \Cref{lem:bonus-per-episode}. For the bound, as $B_k\ge0$, \Cref{eq:app-linear-B} gives
  \begin{equation*}
    B_k(s,a)\le b_k(s)+4H(H-h-1)u_k(s,a)+\sum_{s'}\PP(s'\mid s,a)\sum_{a'}\pi_k(a'\mid s')B_k(s',a'),\quad\forall s\in\cS_h,a\in\cA.
  \end{equation*}
  Unrolling it along a trajectory of $\pi_k$ and using \Cref{eq:app-lin-bonus} with $a_h\sim\pi_k(\cdot\mid s_h)$, we get
  \begin{align*}
    \sum_a\pi_k(a\mid s_0)B_k(s_0,a)&\le\sum_{h=0}^{H-1}\E_{\pi_k}\left[b_k(s_h)+4H(H-h-1)u_k(s_h,a_h)\right]\\
    &=\sum_{h=0}^{H-1}\E_{\pi_k}\left[\frac{H\theta\ind[u_k(s_h,a_h)=0]}{\phat_k(s_h,a_h)}+4H(H-h)u_k(s_h,a_h)\right].
  \end{align*}
  As every layer has fewer than $\nu$ unfinished estimates before processing $k$, $u_k(s,a)=0$ if and only if $\phi(s,a)$ lies in a region. As $(k,h)$ is good, $\phi_h$ lies in a region $\cC_{k,h,j}$ with probability at most $4\phat_{k,h,j}$ under $\pi_k$, and outside all regions with probability at most $4\theta$. Hence, as every layer has at most $J_0$ regions (\Cref{lem:lin-regions-det}), we have
  \begin{align*}
    \sum_a\pi_k(a\mid s_0)B_k(s_0,a)&\le\sum_{h=0}^{H-1}\left(\sum_jH\theta\cdot\frac{4\phat_{k,h,j}}{\phat_{k,h,j}}+4H(H-h)\cdot4\theta\right)\\
    &\le4H^2\theta J_0+8H^2(H+1)\theta=\O(H^3\theta\log K),
  \end{align*}
  which finishes the proof.
\end{proof}

\section{Auxiliary Lemmas}
\label{app:tools}\label{app:prelim}

\begin{lemma}[Freedman's inequality]\label{lem:freedman}
  Let $\cH'_1\subseteq\cH'_2\subseteq\cdots\subseteq\cH'_{N+1}$ be $\sigma$-algebras, let $X_n\ge0$ be $\cH'_{n+1}$-measurable for every $n\in[N]$, and let $\delta'\in(0,1)$. In our applications, the $n$-th term belongs to the $n$-th considered episode $\kappa_n$ in the processing order, $\cH'_n=\cH_{\kappa_n-1}$ is generated by the data processed before $\kappa_n$ (\Cref{lem:processing}), and $\cH'_{N+1}$ is generated by all data.
  \begin{enumerate}[label=\textup{(\roman*)}]
    \item\label{lem:adapted-upper} If $X_n\le R$, and $m_n\ge\E[X_n\mid\cH'_n]$ is $\cH'_n$-measurable for every $n$, then with probability at least $1-\delta'$,
    \begin{equation*}
      \sum_{n=1}^NX_n\le(e-1)\sum_{n=1}^Nm_n+R\log\frac1{\delta'}.
    \end{equation*}
    \item\label{lem:adapted-lower} If every $X_n$ is square-integrable, then for every $\lambda>0$, with probability at least $1-\delta'$,
    \begin{equation*}
      \sum_{n=1}^NX_n\ge\sum_{n=1}^N\E[X_n\mid\cH'_n]-\frac\lambda2\sum_{n=1}^N\E\left[X_n^2\middle|\cH'_n\right]-\frac{\log(1/\delta')}\lambda.
    \end{equation*}
  \end{enumerate}
\end{lemma}

\begin{proof}
  For \ref{lem:adapted-upper}, by $e^y\le1+(e-1)y$ for $y\in[0,1]$ and $1+x\le e^x$, we have
  \begin{equation*}
    \E\left[\exp\left(\frac{X_n}R\right)\middle|\cH'_n\right]\le1+\frac{e-1}R\E[X_n\mid\cH'_n]\le\exp\left(\frac{(e-1)m_n}R\right),
  \end{equation*}
  so $\exp(\sum_{n'\le n}(X_{n'}-(e-1)m_{n'})/R)$ is a supermartingale with respect to $(\cH'_{n+1})_n$. For \ref{lem:adapted-lower}, by $e^{-y}\le1-y+y^2/2$ for $y\ge0$, we have
  \begin{equation*}
    \E\left[\exp\left(-\lambda X_n\right)\middle|\cH'_n\right]\le1-\lambda\E[X_n\mid\cH'_n]+\frac{\lambda^2}2\E\left[X_n^2\middle|\cH'_n\right]\le\exp\left(-\lambda\E[X_n\mid\cH'_n]+\frac{\lambda^2}2\E\left[X_n^2\middle|\cH'_n\right]\right),
  \end{equation*}
  so $\exp(\sum_{n'\le n}(\lambda\E[X_{n'}\mid\cH'_{n'}]-\lambda X_{n'}-(\lambda^2/2)\E[X_{n'}^2\mid\cH'_{n'}]))$ is a supermartingale with respect to $(\cH'_{n+1})_n$. Both claims then follow from Markov's inequality.
\end{proof}

\begin{lemma}[Stability of Exponential Weights]\label{lem:hedge-step}
  Let $\cA'$ be a finite set, $\pi\in\Delta(\cA')$ with $\pi>0$, $g\colon\cA'\to\R$, $\eta>0$, and $\pi'(a)\propto\pi(a)e^{-\eta g(a)}$. Then every $a\in\cA'$ satisfies
  \begin{equation*}
    \left|\log\frac{\pi'(a)}{\pi(a)}\right|\le\eta\left(\max_{a'}g(a')-\min_{a'}g(a')\right).
  \end{equation*}
\end{lemma}

\begin{proof}
  By definition,
  \begin{equation*}
    \frac{e^{-\eta\max_{a'}g(a')}}{e^{-\eta\min_{a'}g(a')}}\le\frac{\pi'(a)}{\pi(a)}=\frac{e^{-\eta g(a)}}{\sum_{a'}\pi(a')e^{-\eta g(a')}}\le\frac{e^{-\eta\min_{a'}g(a')}}{e^{-\eta\max_{a'}g(a')}},
  \end{equation*}
  which finishes the proof.
\end{proof}

\begin{lemma}[Exponential Weights; {\citealp[Lemma~A.4]{luo2021policy}}]\label{lem:hedge}
  Let $\cA'$ be a finite set, $\eta>0$, and $g_1,\ldots,g_K\colon\cA'\to\R$ with $\eta|g_k(a)|\le1$ for all $k$ and $a$. Then $\pi_k(a)\propto\exp(-\eta\sum_{j<k}g_j(a))$ ensures, for every $p\in\Delta(\cA')$,
  \begin{equation*}
    \sum_{k=1}^K\sum_{a\in\cA'}\left(\pi_k(a)-p(a)\right)g_k(a)\le\frac{\log|\cA'|}\eta+\eta\sum_{k=1}^K\sum_{a\in\cA'}\pi_k(a)g_k(a)^2.
  \end{equation*}
\end{lemma}

For i.i.d.\ samples $X_1,\ldots,X_n\sim\mu$ with empirical distribution $\hat\mu$, let $\mu f\triangleq\int f\,d\mu$ and $\hat\mu f\triangleq(1/n)\sum_if(X_i)$.

\begin{lemma}[Uniform Convergence with Relative Errors]\label{lem:vc}
  Let $\cT$ be a class of measurable subsets of $\R^d$ with VC dimension at most $V\ge1$, $\delta''\in(0,1)$, and
  \begin{equation*}
    \epsilon\triangleq\frac9{2n}\left(V\log(2en)+\log\frac8{\delta''}\right).
  \end{equation*}
  Then, with probability at least $1-\delta''$, every $T\in\cT$ satisfies
  \begin{equation*}
    \frac23\left(\mu(T)-\epsilon\right)\le\hat\mu(T)\le\frac32\mu(T)+\epsilon,
  \end{equation*}
  and so do $\mu f$ and $\hat\mu f$ for every $f\colon\R^d\to[0,1]$ with level sets in $\cT$.
\end{lemma}

\begin{proof}
  Let $\epsilon'\triangleq(V\log(2en)+\log(8/\delta''))/n$, so that $\epsilon=9\epsilon'/2$. By \citet[Theorem~5.1]{boucheron2005theory} and Sauer's lemma \citep[Lemma~6.10]{shalev2014understanding}, with probability at least $1-\delta''$, every $T\in\cT$ satisfies $\mu(T)-\hat\mu(T)\le2\sqrt{\mu(T)\epsilon'}$ and $\hat\mu(T)-\mu(T)\le2\sqrt{\hat\mu(T)\epsilon'}$. Hence, we have
  \begin{equation*}
    \mu(T)-\hat\mu(T)\le2\sqrt{\frac{\mu(T)}3\cdot3\epsilon'}\le\frac{\mu(T)}3+3\epsilon',
  \end{equation*}
  which rearranges to $\hat\mu(T)\ge(2/3)\mu(T)-3\epsilon'=(2/3)(\mu(T)-\epsilon)$. In the same way, we have
  \begin{equation*}
    \hat\mu(T)-\mu(T)\le2\sqrt{\frac{\hat\mu(T)}3\cdot3\epsilon'}\le\frac{\hat\mu(T)}3+3\epsilon',
  \end{equation*}
  which rearranges to $\hat\mu(T)\le(3/2)\mu(T)+(9/2)\epsilon'=(3/2)\mu(T)+\epsilon$. This proves the claim for sets, and integrating it over the level sets, $\mu f=\int_0^1\mu(\{f>t\})\mathrm dt$ and $\hat\mu f=\int_0^1\hat\mu(\{f>t\})\mathrm dt$, gives the claim for $f$.
\end{proof}

\begin{lemma}[Confidence Sets]\label{lem:good-event}
  With probability at least $1-\delta/4$, $\PP$ satisfies every constraint of \Cref{eq:app-conf-set} that defines $\cP_k$, for every non-skipped $k\in[K]$, so that the good event $G$ of \Cref{app:tab-est} holds.
\end{lemma}
\begin{proof}
  The proof mostly follows \citet[Lemma~2]{jin2020learning}. Fix a pair $(s,a)$. By \Cref{lem:processing}\ref{item:proc-fresh}, given the data processed before it, the trajectory of every processed episode follows $\PP$. Hence, the next states observed after the visits to $(s,a)$, taken in the processing order, can be viewed as the first terms of an i.i.d.\ sequence with law $\PP(\cdot\mid s,a)$, and every $\cP_k$ only uses a prefix of this sequence, as its counts come from the episodes processed before $k$.

  Now fix $s'$ and $n\ge2$. The indicators that the first $n$ next states equal $s'$ have mean $\Phat_n(s'\mid s,a)$ and sample variance at most $\frac n{n-1}\Phat_n(s'\mid s,a)$. Hence, applying the empirical Bernstein bound \citep[Theorem~4]{maurer2009empirical} to these indicators and to their complements, we have, except with probability $\delta/(4S^2AK)$,
  \begin{equation*}
    \left|\Phat_n(s'\mid s,a)-\PP(s'\mid s,a)\right|\le\sqrt{\frac{2\Phat_n(s'\mid s,a)\beta}{n-1}}+\frac{7\beta}{3(n-1)}\le2\sqrt{\frac{\Phat_n(s'\mid s,a)\beta}n}+\frac{5\beta}n,
  \end{equation*}
  where $\beta=\log(16S^2AK/\delta)$ is as in \Cref{eq:app-conf-set}, and the last step used $n\le2(n-1)$. For $n=1$, the same bound holds as its right-hand side is at least $5\beta\ge1$.

  Finally, as $\qlow_k$, $B_k$, and the occupancy measures do not depend on the transitions into $\cS_H$, we can merge the terminal states into a single state. Then there are at most $S^2A$ triples $(s,a,s')$, and a union bound over them and the $K$ values of $n$ finishes the proof.
\end{proof}

\end{document}